\documentclass[10pt,twocolumn,letterpaper]{article}

\usepackage[pagenumbers]{cvpr} 

\usepackage{graphicx}
\usepackage{amsmath}
\usepackage{amssymb}
\usepackage{enumitem}
\usepackage{booktabs}
\usepackage{bm}
\usepackage{multirow}
\usepackage[accsupp]{axessibility}
\usepackage{booktabs}
\usepackage{bm}
\usepackage{algorithm}
\usepackage{algpseudocode}
\usepackage{multirow}
\usepackage[accsupp]{axessibility}

\usepackage[pagebackref=false,breaklinks,colorlinks]{hyperref}

\usepackage[capitalize]{cleveref}
\crefname{section}{Sec.}{Secs.}
\Crefname{section}{Section}{Sections}
\Crefname{table}{Table}{Tables}
\crefname{table}{Tab.}{Tabs.}

\def\confName{CVPR}
\def\confYear{2023}
\usepackage{xcolor}
\definecolor{ao}{rgb}{0.0, 0.5, 0.0}
\definecolor{bluepigment}{rgb}{0.2, 0.2, 0.6}
\definecolor{coolblack}{rgb}{0.0, 0.18, 0.39}
\definecolor{codebrown}{rgb}{0.37,0.0,0.0}
\newcommand{\formattedparagraph}[1]{\noindent \textbf{#1}}

\pdfoutput=1

\begin{document}

\title{Single Image Depth Prediction Made Better: A Multivariate Gaussian Take}

\author{Ce Liu$^{1}$~~~~~~Suryansh Kumar$^{1}\thanks{Corresponding Author (k.sur46@gmail.com)}$~~~~~~Shuhang Gu$^{2}$~~~~~~Radu Timofte$^{1,3}$~~~~~~Luc Van Gool$^{1, 4}$ \\ 
$^1$CVL ETH Z\"urich~~~$^2$UESTC China ~~~$^3$University of W\"urzburg~~~$^4$KU Leuven\\
{\tt{\small \{ce.liu, sukumar, vangool\}@vision.ee.ethz.ch}}\\ 
{\tt{\small shuhanggu@uestc.edu.cn, radu.timofte@uni-wuerzburg.de}}
}

\maketitle

\begin{abstract}
Neural-network-based single image depth prediction (SIDP) is a challenging task where the goal is to predict the scene's per-pixel depth at test time. Since the problem, by definition, is ill-posed, the fundamental goal is to come up with an approach that can reliably model the scene depth from a set of training examples. In the pursuit of perfect depth estimation, most existing state-of-the-art learning techniques predict a single scalar depth value per-pixel. Yet, it is well-known that the trained model has accuracy limits and can predict imprecise depth. Therefore, an SIDP approach must be mindful of the expected depth variations in the model's prediction at test time. Accordingly, we introduce an approach that performs continuous modeling of per-pixel depth, where we can predict and reason about the per-pixel depth and its distribution. To this end, we model per-pixel scene depth using  a multivariate Gaussian distribution.
Moreover, contrary to the existing uncertainty modeling methods---in the same spirit, where per-pixel depth is assumed to be independent, we introduce per-pixel covariance modeling that encodes its depth dependency \wrt all the scene points. Unfortunately, per-pixel depth covariance modeling leads to a computationally expensive continuous loss function, which we solve efficiently using the learned low-rank approximation of the overall covariance matrix. Notably, when tested on benchmark datasets such as KITTI, NYU, and SUN-RGB-D, the SIDP model obtained by optimizing our loss function shows state-of-the-art results. Our method's accuracy (named MG) is among the top on the KITTI depth-prediction benchmark leaderboard\footnote{\url{http://www.cvlibs.net/datasets/kitti/eval\_depth.php?benchmark=depth\_prediction}}.
\end{abstract}

\section{Introduction}
Recovering the depth of a scene using images is critical to several applications in computer vision \cite{agarwal2011building, furukawa2015multi, kumar2017monocular, kumar2019superpixel, kaya2022uncertainty}. It is well founded that precise estimation of scene depth from images is likely only under multi-view settings \cite{ullman1979interpretation}---which is indeed a correct statement and hard to contend\footnote{As many 3D scene configurations can have the same image projection.}. But what if we could effectively learn scene depth using images and their ground-truth depth values, and be able to predict the scene depth using just a single image at test time? With the current advancements in deep learning techniques, this seems quite possible empirically and has also led to excellent results for the single image depth prediction (SIDP) task \cite{liuva, ranftl2021vision}. Despite critical geometric arguments against SIDP, practitioners still pursue this problem not only for a scientific thrill but mainly because there are several real-world applications in which SIDP can be extremely beneficial. For instance, in medical \cite{liu2019dense}, augmented and virtual reality \cite{hoiem2005automatic,richter2022enhancing}, gaming \cite{haji2018playing}, novel view synthesis \cite{riegler2021stable,roessle2022dense}, robotics \cite{tateno2017cnn}, and related vision applications \cite{ranftl2021vision, Jain_2022_BMVC}.

\begin{figure}[t]
\begin{center}
\begin{subfigure}[b]{0.15\textwidth}
\begin{center}
\includegraphics[width=1.0\textwidth]{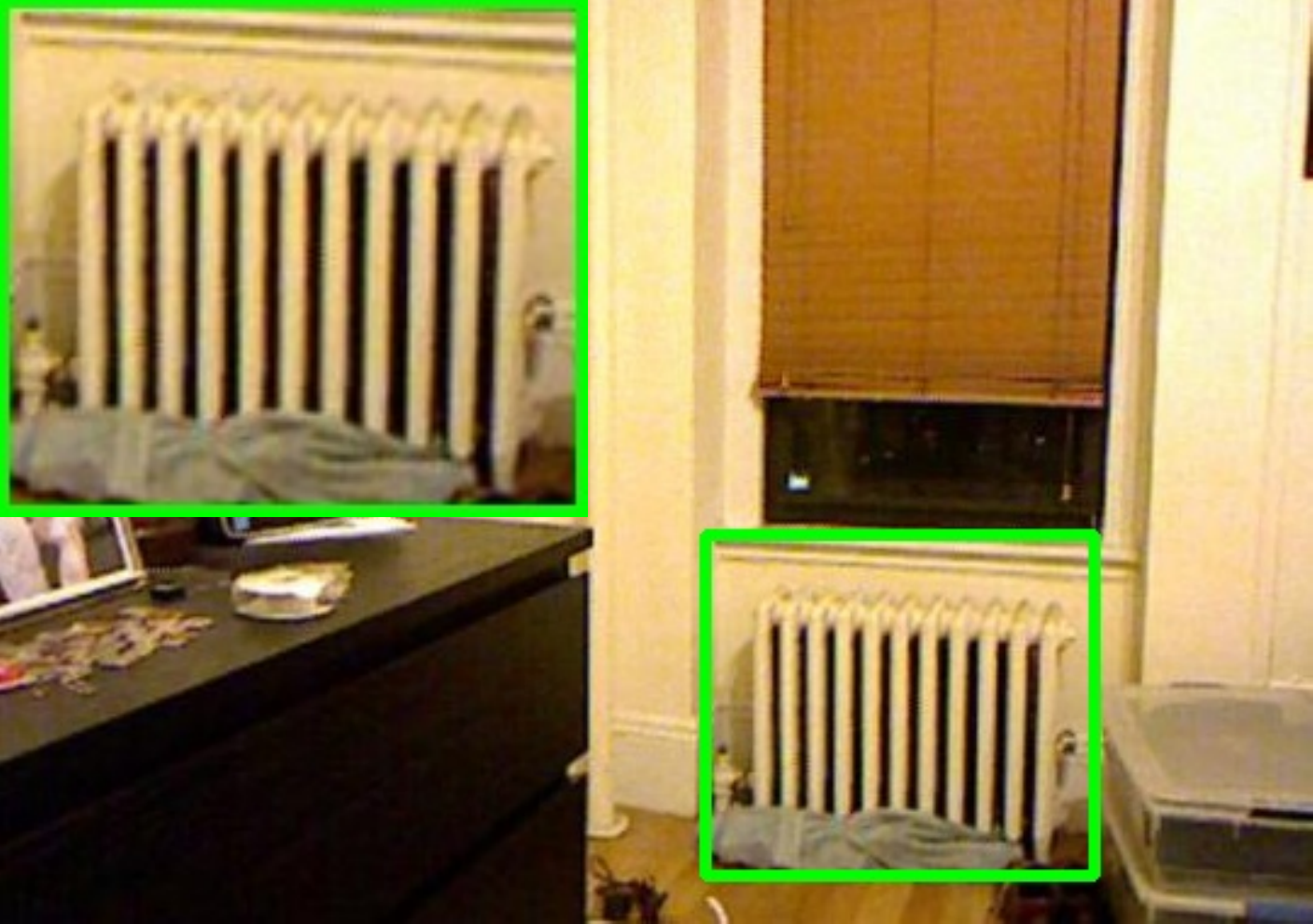}
\caption{Test Image}
\end{center}
\end{subfigure}
\begin{subfigure}[b]{0.15\textwidth}
\begin{center}
\includegraphics[width=1.0\textwidth]{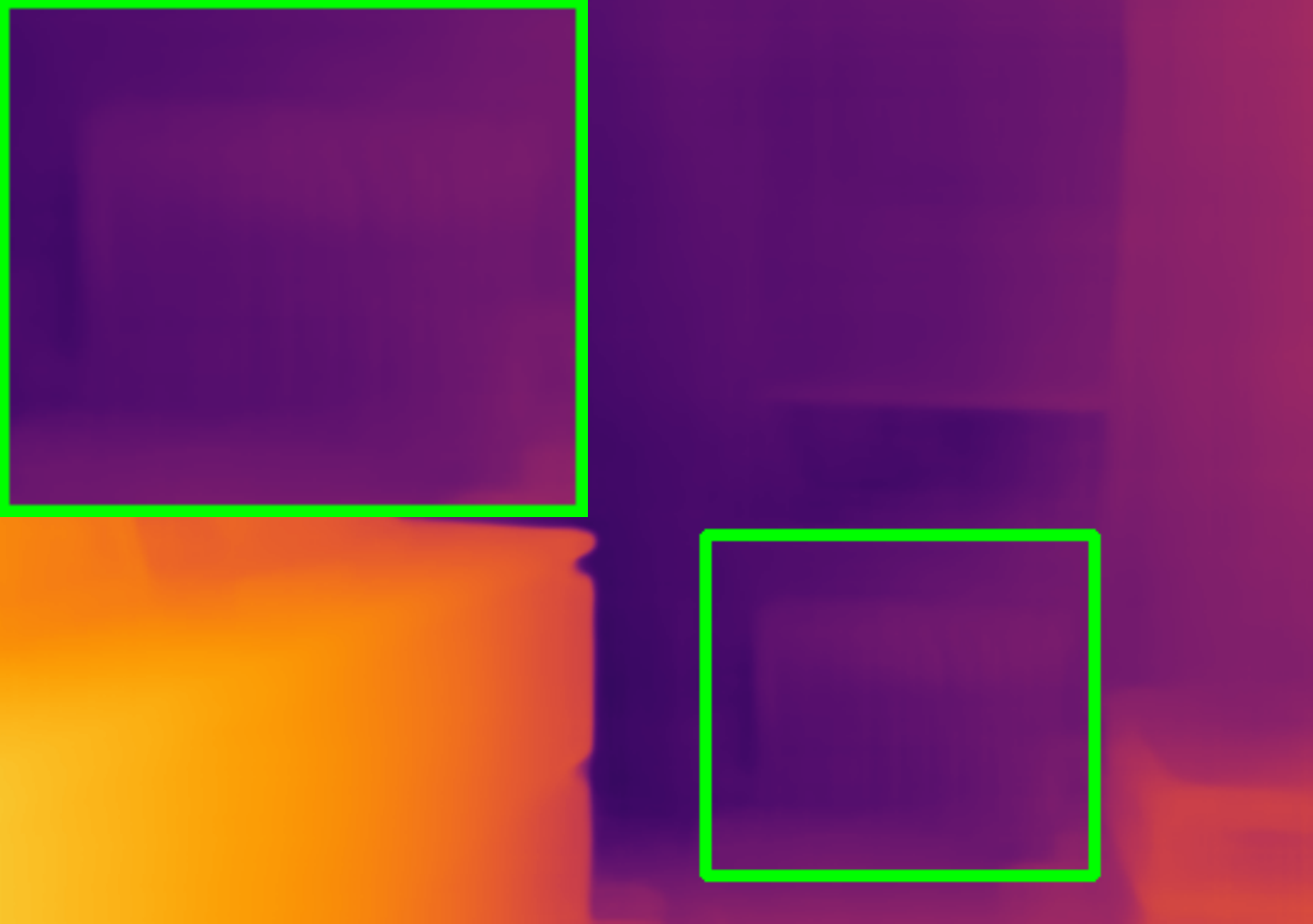}
\caption{DPT \cite{ranftl2021vision}}
\end{center}
\end{subfigure} 
\begin{subfigure}[b]{0.15\textwidth}
\begin{center}
\includegraphics[width=1.0\textwidth]{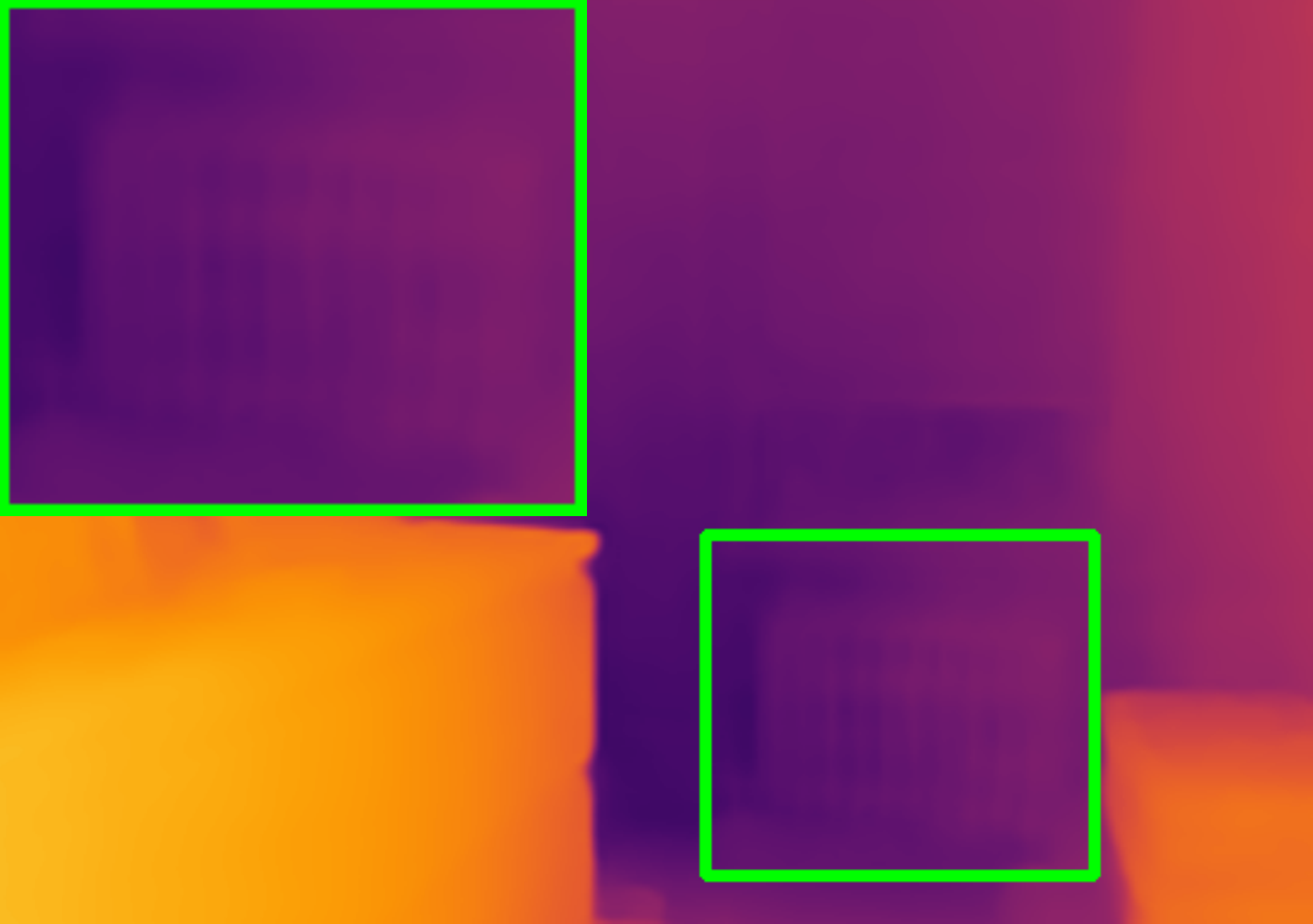}
\caption{AdaBins \cite{bhat2021adabins}}
\end{center}
\end{subfigure} 
\begin{subfigure}[b]{0.15\textwidth}
\begin{center}
\includegraphics[width=1.0\textwidth]{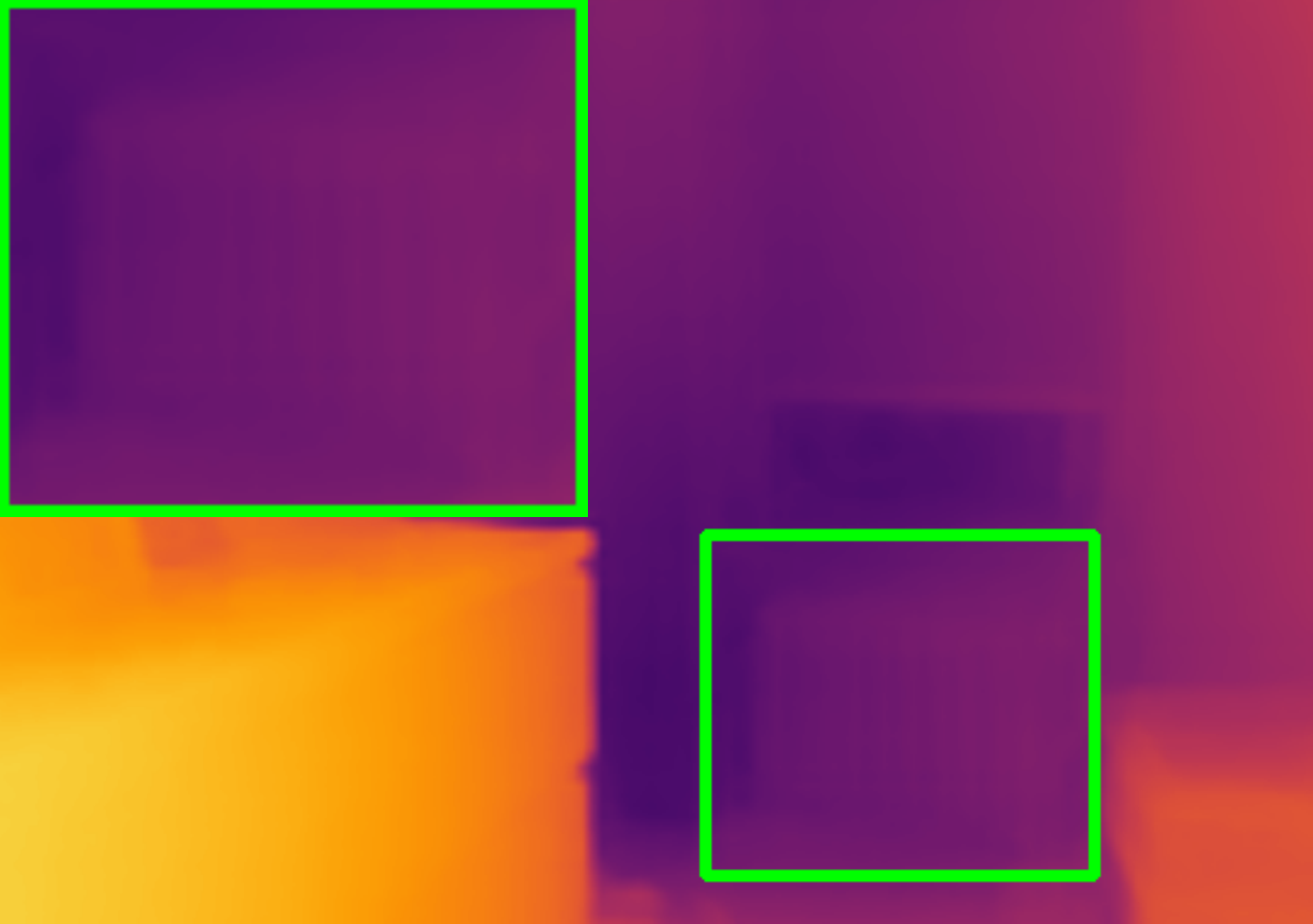}
\caption{NeWCRFs \cite{yuan2022new}}
\end{center}
\end{subfigure} 
\begin{subfigure}[b]{0.15\textwidth}
\begin{center}
\includegraphics[width=1.0\textwidth]{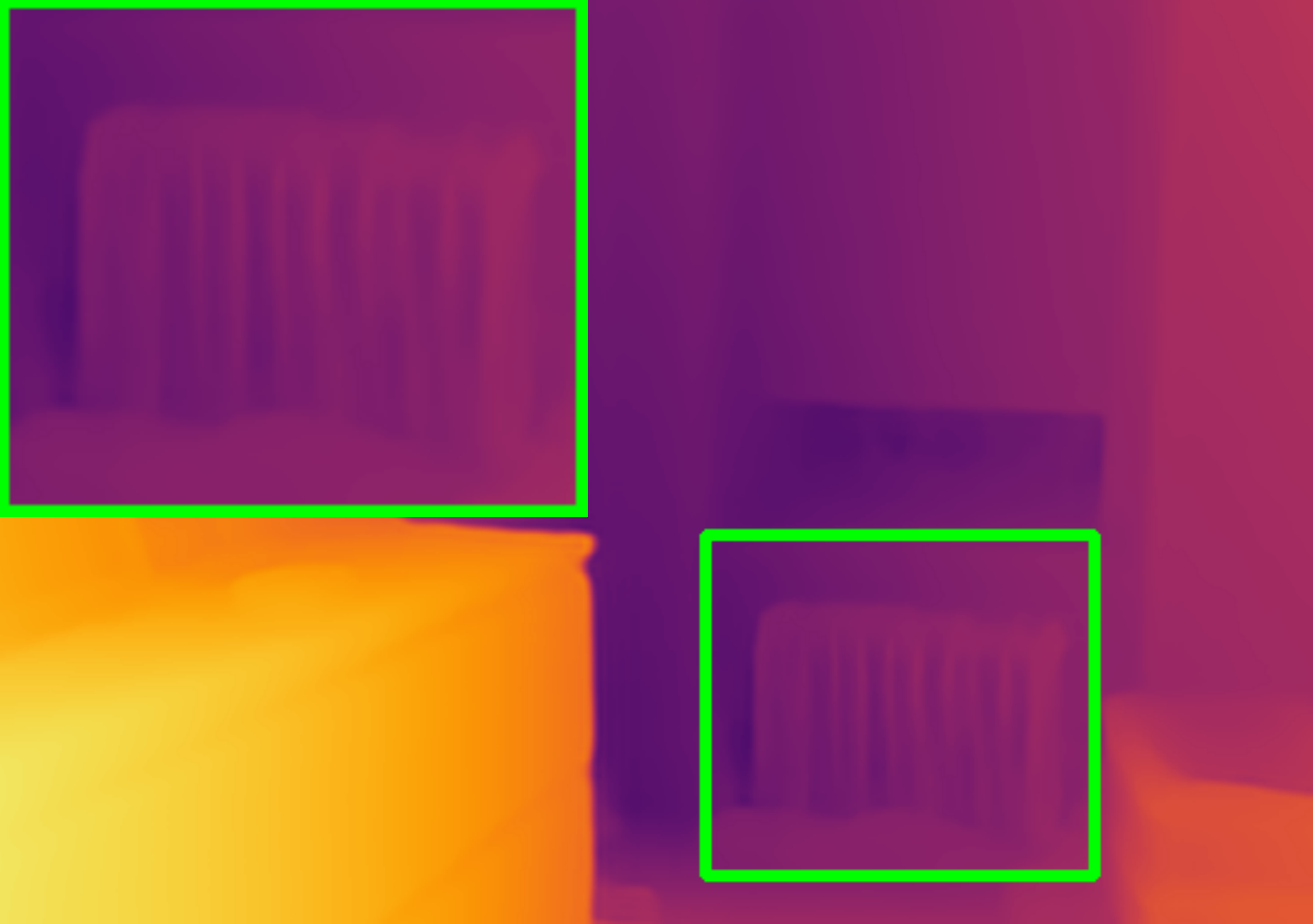}
\caption{\textbf{Ours}}
\end{center}
\end{subfigure} 
\begin{subfigure}[b]{0.15\textwidth}
\begin{center}
\includegraphics[width=1.0\textwidth]{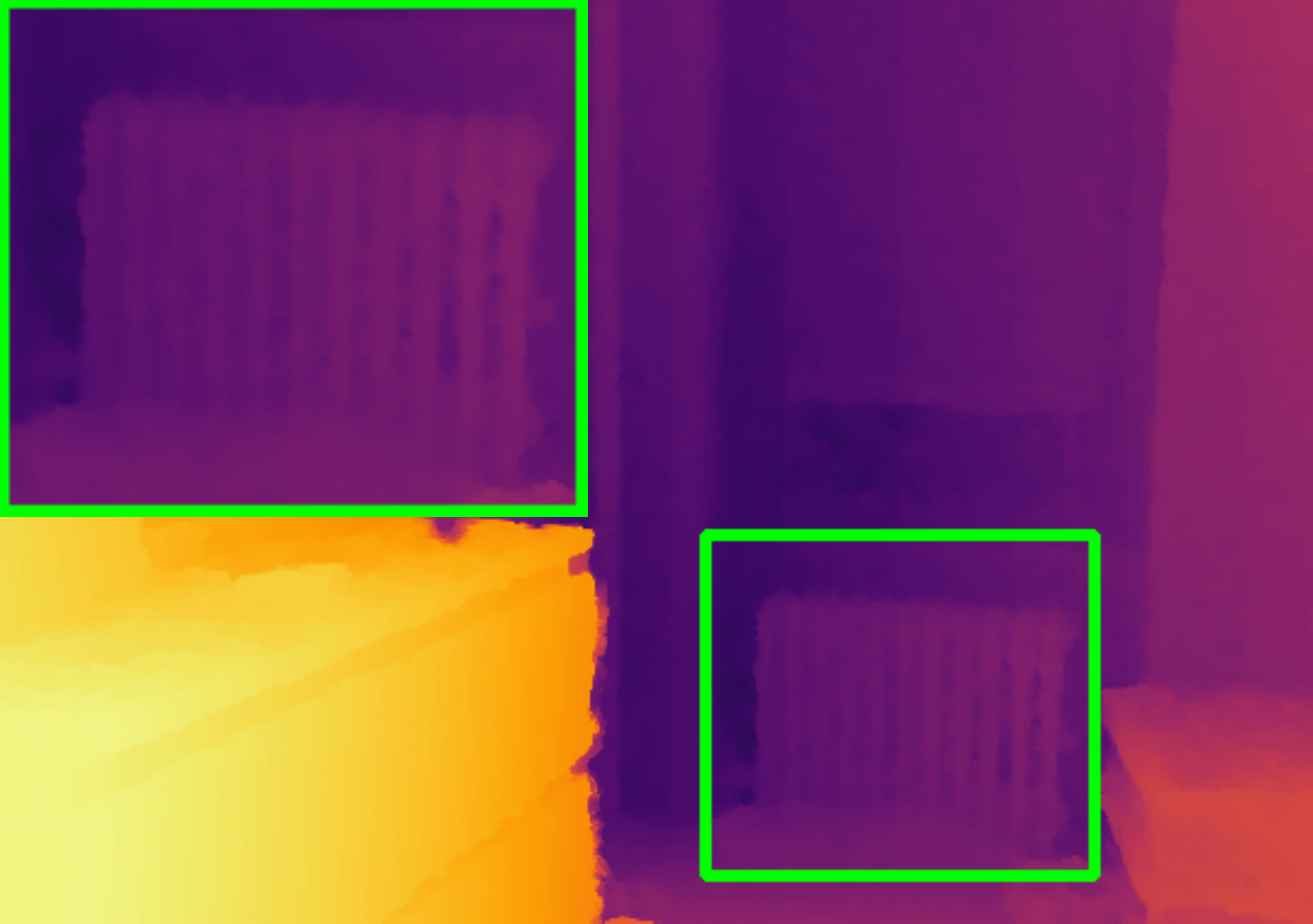}
\caption{Ground Truth}
\end{center}
\end{subfigure} 
\end{center}
\caption{\textbf{Qualitative Comparison}. By modeling scene depth as multivariate Gaussian and enforcing the parametric low-rank covariance constraints in the loss function, we observe that our model can reliably predict depth for both high-frequency and low-frequency scene details. In the above example, we can notice better qualitative results than the state-of-the-art methods.}\label{fig:first_page_teaser}
\end{figure}

\begin{figure}[t]
\begin{center}
\begin{subfigure}[b]{0.23\textwidth}
\begin{center}
\includegraphics[width=0.9\textwidth]{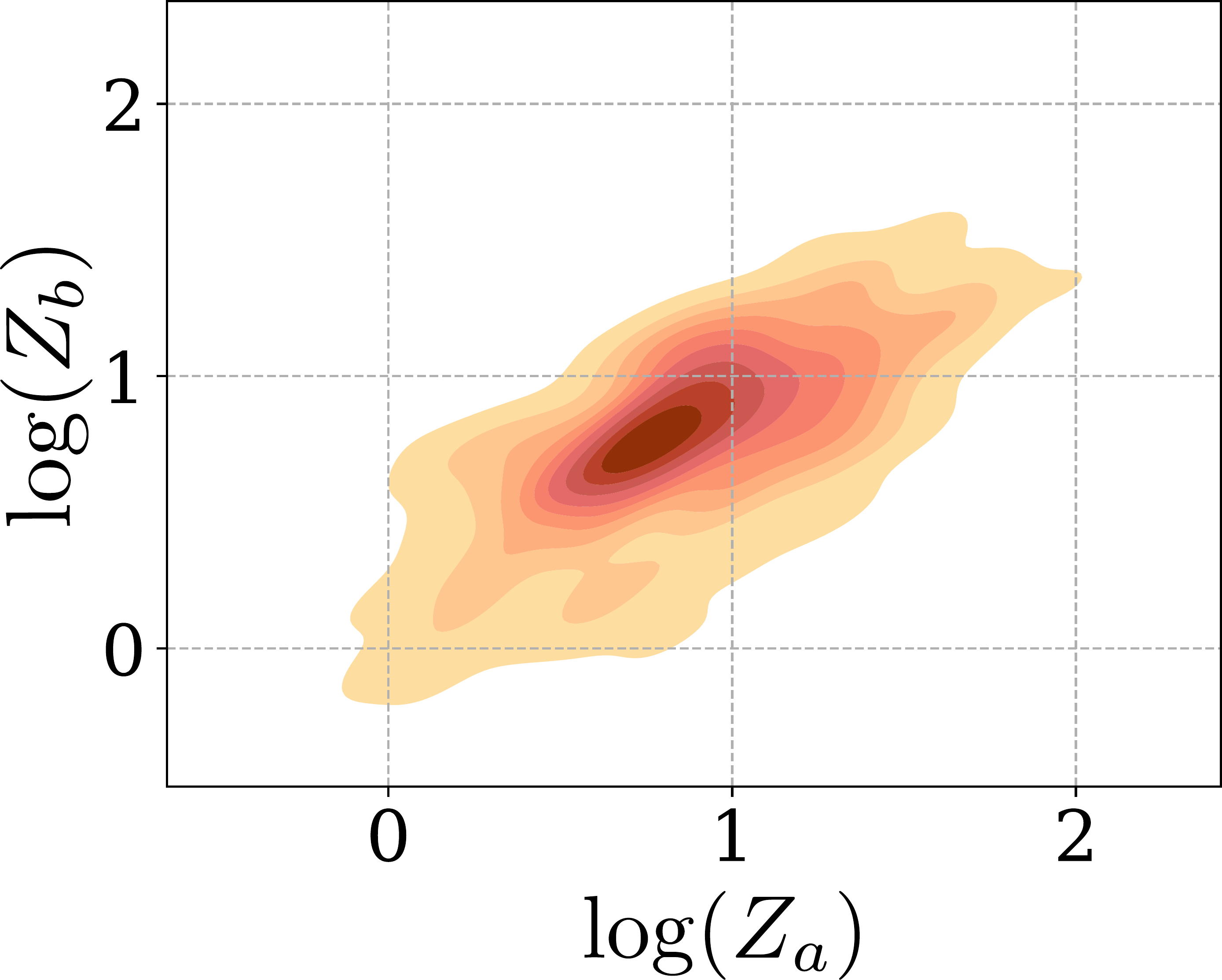}
\caption{First scene}
\end{center}
\end{subfigure}
\hspace{1.5pt}
\begin{subfigure}[b]{0.23\textwidth}
\begin{center}
\includegraphics[width=0.9\textwidth]{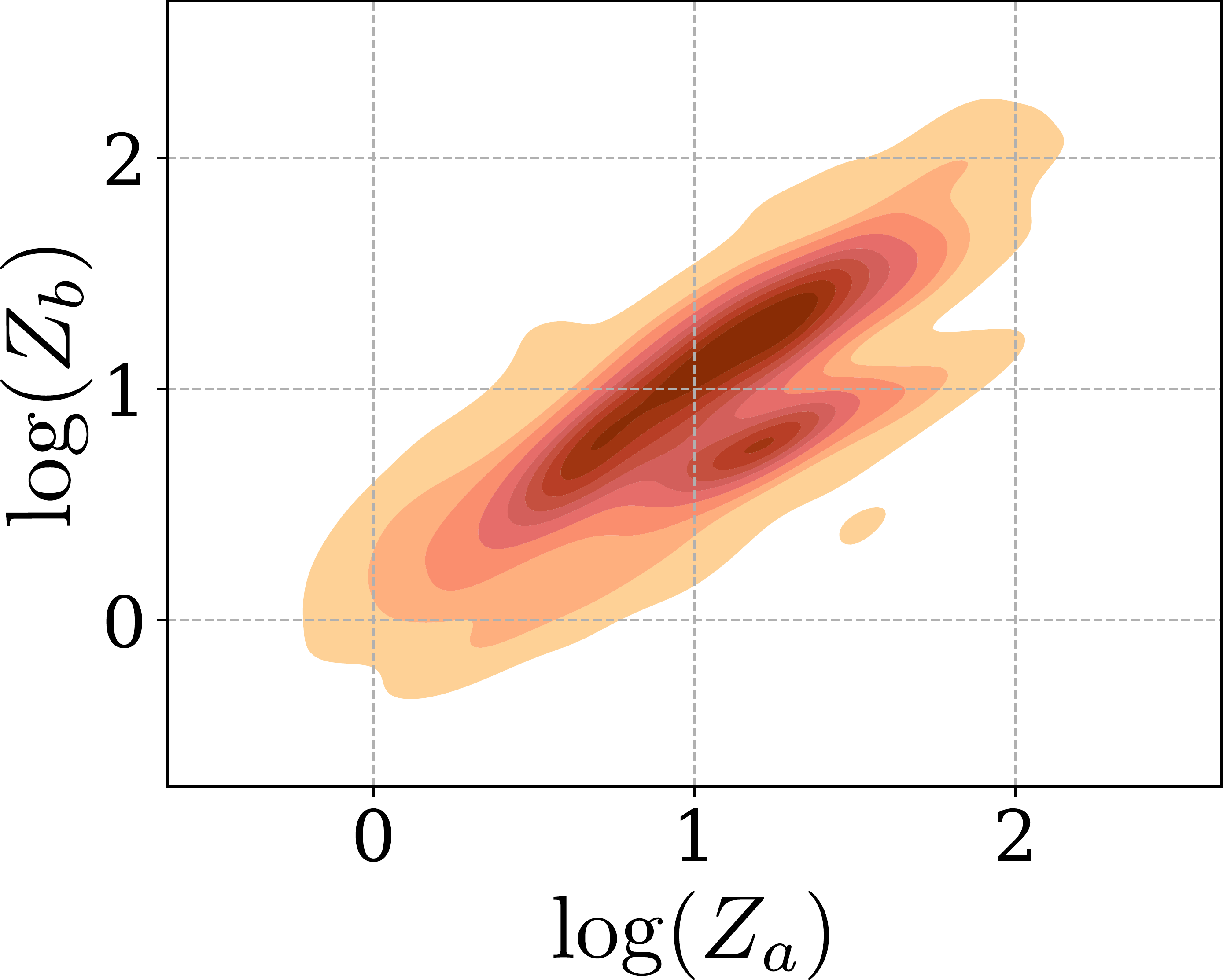}
\caption{Second scene}
\end{center}
\end{subfigure} 
\end{center}
\caption{The marginal ground-truth depth distribution for a pixel pair $Z_a$, $Z_b$ for two scenes. The depth values for the pixel pair are taken from the fixed image location in the dataset, but the selected images are visually similar for the suitability of the feature and its corresponding depth values. The statistics show that the Gaussian distribution assumption with covariance modeling is a sensible choice for SIDP problem and not an unorthodox belief arranged or staged  for an intricate  formulation.}\label{fig:kde_example}
\end{figure}

Regardless of remarkable progress in SIDP \cite{li2022binsformer,yuan2022new,li2022depthformer, qiao2021vip, agarwal2022attention, liu2021deep, liuva}, the recent state-of-the-art deep-learning methods, for the time being, just predict a single depth value per pixel at test time \cite{li2022binsformer}. Yet, as is known, trained models have accuracy limits. As a result, for broader adoption of SIDP in applications, such as robot vision and control, it is essential to have information about the reliability of predicted depth. Consequently, we model the SIDP task using a continuous distribution function. Unfortunately, it is challenging, if not impossible, to precisely model the continuous 3D scene. In this regard, existing methods generally resort to increasing the size and quality of the dataset for better scene modeling and improve SIDP accuracy. On the contrary, little progress is made in finding novel mathematical modeling strategies and exploiting the prevalent scene priors. To this end, we propose a multivariate Gaussian distribution to model scene depth. In practice, our assumption of the Gaussian modeling of data is in consonance with real-world depth data (see Fig. \ref{fig:kde_example}) and generalizes well across different scenes. Furthermore, many computer and robot vision problems have successfully used it and benefited from Gaussian distribution modeling in the past \cite{chen2021gaussian, zhu2022unified, stroupe2001merging, hewing2019cautious, yamashita20193d, rasmussen2003gaussian, murphy2012machine}.

Let's clarify this out way upfront that this is not for the first time an approach with a motivation of continuous modeling for SIDP is proposed \cite{amini2020deep,  lakshminarayanan2017simple, kendall2017uncertainties, klodt2018supervising, hornauer2022gradient, poggi2020uncertainty}. Yet, existing methods in this direction model depth per pixel independently. It is clearly unreasonable, in SIDP modeling, to assume absolute democracy among each pixel, especially for very closeby scene points. Therefore, it is natural to think of modeling this problem in a way where depth at a particular pixel can help infer, refine, and constrain the distribution of depth value for other image pixels. Nevertheless, it has yet to be known a priori the neighboring relation among pixels in the scene space to define the depth covariance among them. We do that here by defining a very general covariance matrix of dimension $\texttt{number of pixels} \times \texttt{number of pixels}$, \ie, depth prediction at a given pixel is assumed to be dependent on all other pixels' depth.

Overall, we aim to advocate multivariate Gaussian modeling with a notion of depth dependency among pixels as a useful scene prior. Now, add a fully dependent covariance modeling proposal to it---as suitable relations among pixels are not known. This makes the overall loss function computationally expensive. To efficiently optimize the proposed formulation, we parameterize the covariance matrix, assuming that it lies in a rather low-dimensional manifold so that it can be learned using a simple neural network. For training our deep network, we utilize the negative log likelihood as the loss
function (cf. Sec. \ref{ssec:mgm}). 
The trained model when tested on standard benchmark datasets gives state-of-the-art results for SIDP task (see Fig. \ref{fig:first_page_teaser} for qualitative comparison).

\smallskip
\noindent
\textbf{Contributions.} To summarize, our key contributions are:

\begin{itemize}[leftmargin=*,topsep=0pt, noitemsep]
    \item A novel formulation to perform multivariate Gaussian covariance modeling for solving the SIDP task in a deep neural network framework is introduced. 
    \item The introduced multivariate Gaussian covariance modeling for SIDP is computationally expensive.  To solve it efficiently, the paper proposes to learn the low-rank covariance matrix approximation by deep neural networks.
    \item Contrary to the popular SIDP methods, the proposed approach provides better depth as well as a measure of the suitability of the predicted depth value at test time.
\end{itemize}

\begin{figure*}[t]
\begin{center}
\begin{subfigure}[b]{0.24\textwidth}
\begin{center}
\includegraphics[width=1.0\textwidth]{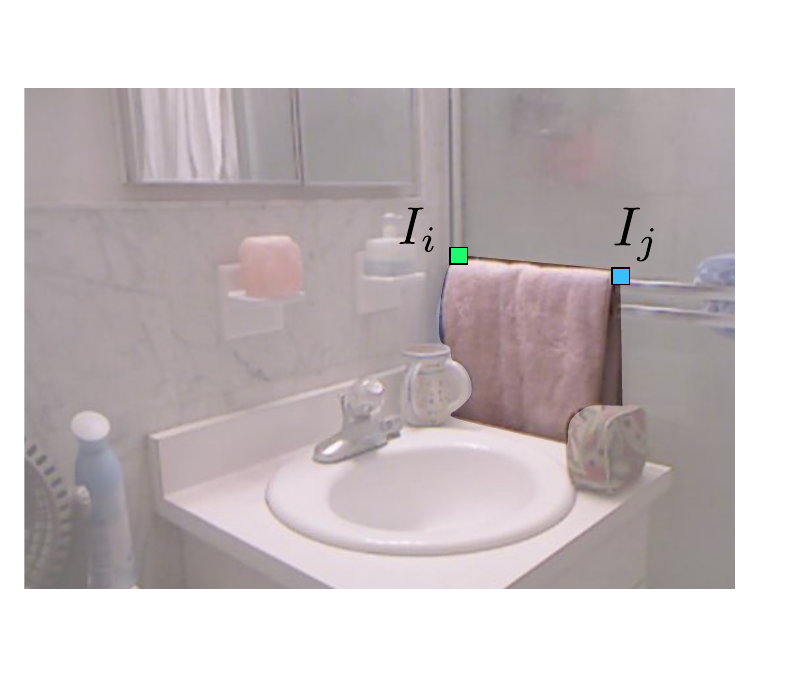}
\caption{Test Image}\label{fig:test_a}
\end{center}
\end{subfigure}
\begin{subfigure}[b]{0.24\textwidth}
\begin{center}
\includegraphics[width=1.0\textwidth]{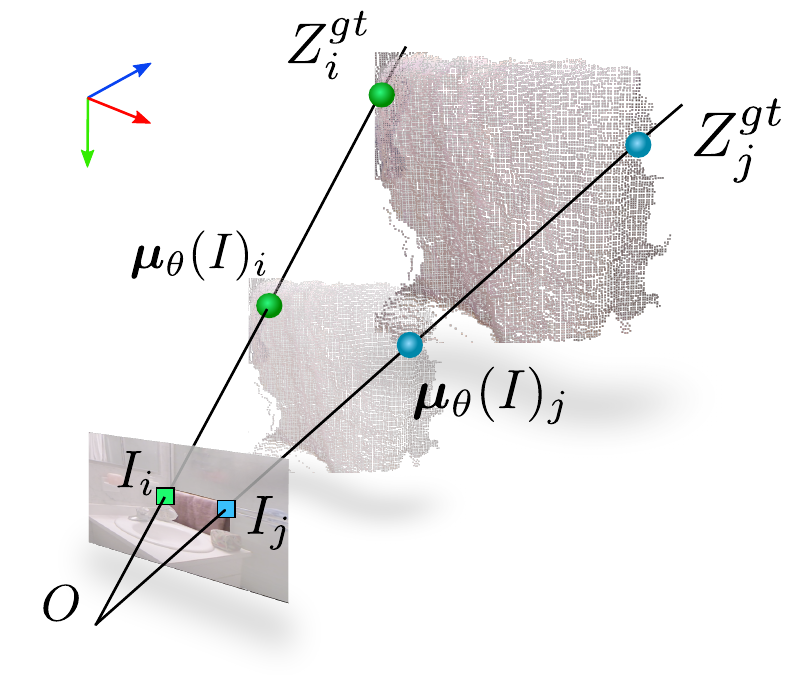}
\caption{Scale Ambiguity}\label{fig:scale_b}
\end{center}
\end{subfigure}
\begin{subfigure}[b]{0.24\textwidth}
\begin{center}
\includegraphics[width=1.0\textwidth]{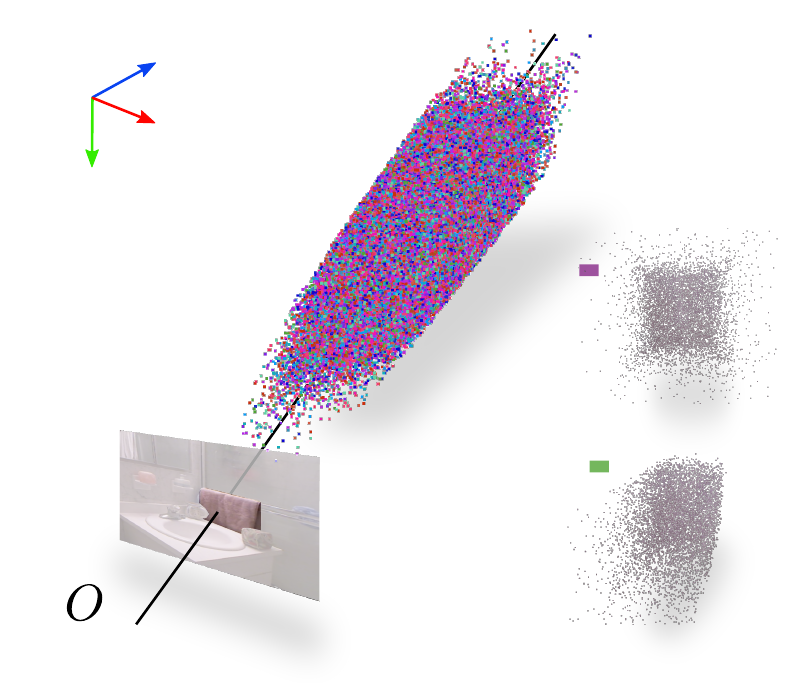}
\caption{Independent Modeling}\label{fig:independent_c}
\end{center}
\end{subfigure}
\begin{subfigure}[b]{0.24\textwidth}
\begin{center}
\includegraphics[width=1.0\textwidth]{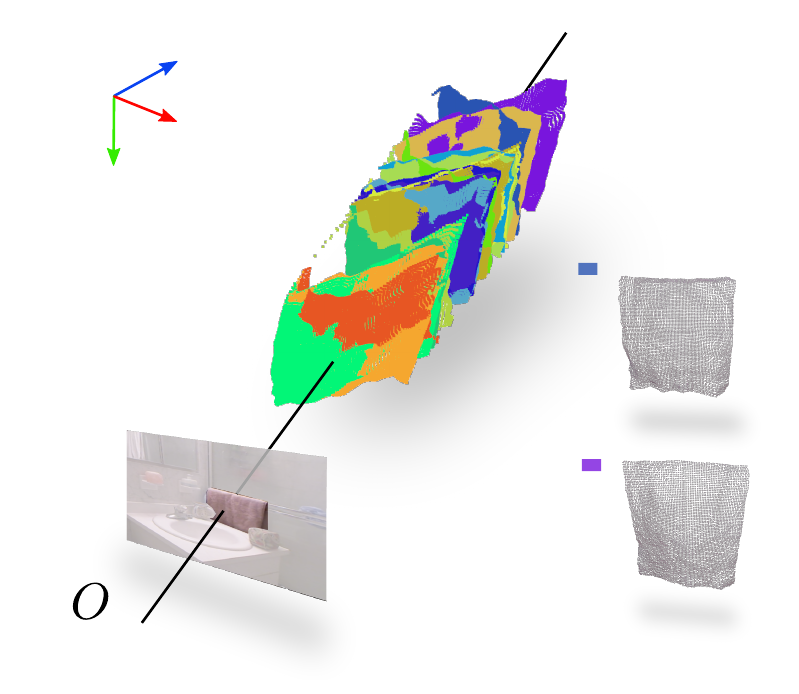}
\caption{Our Modeling}\label{fig:test_ours}
\end{center}
\end{subfigure}
\end{center}
\caption{(a) Test image of an indoor bathroom scene. (b) The problem of scale ambiguity: showing several possible 3D point-cloud configurations of the towel with same imaging region. (c) If the depth values from the towel region are back-projected in the scene space under the independent Gaussian distribution assumption of the depth map. Clearly, the 3D point cloud results are not encouraging. (d) Samples due to our multivariate Gaussian distribution modeling that constrain the pixel depth with learned covariance. We observe the samples drawn from our modeling provide better 3D point clouds. Note: depth map is transformed to point cloud for visualization.
}
\end{figure*}

\section{Related Work}
Predicting the scene depth from a single image is a popular problem in computer vision with long-list of approaches. To keep our review of the existing literature succinct and on-point, we discuss work of direct relevance to the proposed method. Roughly, we divide well-known methods into two sub-category based on their depth prediction modeling.

\smallskip
\formattedparagraph{\textit{(i)} General SIDP.} By general SIDP, we mean methods that predict one scalar depth value per image pixel at inference time. Earlier works include Markov Random Field (MRF) or Conditional Random Fields (CRF) modeling \cite{saxena2005learning, saxena2007learning, saxena2008make3d, wang2015depth}. With the advancement in neural network-based approaches, such classical modeling ideas are further improved using advanced deep-network design \cite{liu2015learning, li2015cvpr, yuan2022new, xu2018structured, xu2017cvpr}. A few other stretches along this line use piece-wise planar scene assumption \cite{lee2019big}. Other variations in deep neural network-based SIDP methods use ranking, ordinal relation constraint, or structured-guided sampling strategy \cite{zoran2015learning, chen2016single, xian2020structure, lienen2021plackett, fu2018deep}. The main drawback with the above deep-leaning methods is that they provide an over-smoothed depth solution, and most of them rely on some heuristic formulation for depth-map refinement as a post-refinement step.

Recently, transformer networks have been used for better feature aggregation via an increase in the network receptive field \cite{yang2021transformer, bhat2021adabins, chang2021transformer, yuan2022new} or with the use of attention supervision \cite{chang2021transformer} leading to better SIDP accuracy. Another mindful attempt is to exploit the surface normal and depth relation. To this end, \cite{hu2019revisiting} introduces both normal and depth loss for SIDP, whereas \cite{yin2019enforcing} proposes using virtual normal loss for imposing explicit 3D scene constraint and utilizing long-range scene dependencies. Long \etal \cite{long2021adaptive} improved over \cite{yin2019enforcing} by introducing an adaptive strategy to compute local patch surface normal by randomly sampling for candidate triplets. A short time ago, \cite{bae2022iron} showed commendable results using normal-guided depth propagation \cite{qi2018geonet} with a depth-to-normal and learnable normal-to-depth module.

\smallskip
\formattedparagraph{\textit{(ii)} Probabilistic SIDP.} In comparison, there are limited pieces of literature that directly target to solve SIDP in a probabilistic way, where the methods could predict the scene depth and simultaneously can reason about its prediction quality. Generally, popular methods from uncertainty modeling in deep-network are used as it is for such purposes. For instance, Kendall \etal \cite{kendall2017uncertainties} Bayesian uncertainty in deep-networks,  Lakshminarayanan \etal \cite{lakshminarayanan2017simple} deep ensemble-based uncertainty modeling, Amini \etal \cite{amini2020deep} deep evidential regression approach, is shown to work also for the depth prediction task. Yet, these methods are very general and can be used for most, if not all, computer vision problems \cite{gal2016uncertainty}. Moreover, these methods treat each pixel independently, which may lead to inferior SIDP modeling.

This brings us to the point that application-specific priors, constraints, and settings could be exploited to enhance the solution, and we must not wholly depend on general frameworks to tackle the problem with similar motivation \cite{kendall2017uncertainties, lakshminarayanan2017simple, amini2020deep}. Therefore, this paper advocate using per-pixel multivariate Gaussian covariance modeling with efficient low-rank covariance parametric representation to improve SIDP for its broader application. Furthermore, we show that the depth likelihood due to multivariate Gaussian distribution modeling can help define better loss function and allow depth covariance learning based on scene feature regularity. With our modeling formulation, the derived loss function naturally unifies the essence of L2 loss, scale-invariant loss, and gradient loss. These three losses can be derived as a special case of our proposed loss (cf. Sec. \ref{ssec:other_loss_relation}).

\section{Proposed Method}\label{sec:method}
To begin with, let's introduce problem setting and some general notations, which we shall be using in the rest of the paper. That will help concisely present our formulation, its useful mathematical insights, discussion, and application to Bayesian uncertainty estimation in deep networks.

\smallskip
\noindent
\textbf{Problem Setting.} Given an image $\mathbf{I} \in \mathbb{R}^{m \times n}$ at test time, our  goal is to predict the reliable per-pixel depth map $\mathbf{Z}  \in \mathbb{R}^{m \times n}$, where $m, n$ symbolize the number of image rows and cols, respectively. For this problem, we reshape the image and corresponding ground-truth depth map as a column vector represented as $I \in \mathbb{R}^{N \times 1}$ and $Z^{gt} \in \mathbb{R}^{N \times 1}$, respectively.
Here, $N = m \times n$ is the total number of pixels in the image and $\mathcal{D}$ denotes the train set.

\subsection{Multivariate Gaussian Modeling}\label{ssec:mgm}
Let's assume the depth map $Z$ corresponding to image $I$ follows a $N$-dimensional Gaussian distribution. Accordingly, we can write the distribution $\Phi$ given $I$ as 
\begin{equation}\label{eq:mvgm_definition}
    \Phi(Z|\theta, I)=\mathcal{N}\big(\bm{\mu}_{\theta}(I), \bm{\Sigma}_{\theta}(I, I)\big).
\end{equation}
Where, $\bm{\mu}_{\theta}(I) \in \mathbb{R}^{N \times 1}$ and $\bm{\Sigma}_{\theta}(I, I) \in \mathbb{R}^{N \times N}$ symbolize the mean and covariance of multivariate Gaussian distribution  $\mathcal{N}$ of predicted depth, respectively. The $\theta$ represents the parametric description of mean and covariance, which the neural network can learn at train time. It is important to note that with such network modeling, it is easy for the network to reliably reason about the scene depth distribution of similar-looking images at test time. Using the general form of multivariate Gaussian density function, the log probability density of Eq.\eqref{eq:mvgm_definition} could be elaborated as
\begin{equation}
\begin{split}
    \log \Phi(Z|\theta, I) = -\frac{N}{2}\log2\pi - \frac{1}{2}\log \det (\bm{\Sigma}_{\theta}(I, I)) -\\
    \frac{1}{2}(Z-\bm{\mu}_\theta)^T (\bm{\Sigma}_{\theta}(I, I))^{-1}(Z-\bm{\mu}_\theta).
\end{split}\label{eq:log_prob_density}
\end{equation}
Eq.\eqref{eq:log_prob_density} is precisely the formulation we strive to implement. Yet, computing the determinant and inverse of a $N \times N$ covariance matrix can be computationally expensive, \ie, $O(N^3)$, for a reasonable image resolution. Previous methods in this direction usually restrict covariance to be diagonal \cite{kendall2017uncertainties,lakshminarayanan2017simple}, \ie, $\bm{\Sigma}_{\theta}(I, I) = \textbf{diag}(\bm{\sigma}_{\theta}^2(I))$ with $\bm{\sigma}_{\theta}$ as the standard deviation learned by the network with parameter $\theta$. Even though such a simplification leads to computationally tractable algorithm $O(N)$, it leads to questionable depth prediction at test time. The reason for that is in SIDP, each pixel's depth could vary by a single scale value which must be the same for all the pixels under the rigid scene assumption (see Fig.\ref{fig:scale_b}). By assuming covariance to be zero, each pixel is modeled independently; hence the coherence among scene points is lost completely. It can also be observed from Fig.(\ref{fig:independent_c}) when the covariance matrix is restricted to a diagonal matrix, the sampled depth from $\Phi(Z|\theta, I)$ is incoherently scattered. Therefore, it is pretty clear that multivariate covariance modeling is essential (see Fig. \ref{fig:test_ours}) despite being computationally expensive.

To overcome the computational bottleneck in covariance modeling, we propose to exploit $\bm{\Sigma}_{\theta}$ parametric form with low-rank assumption. It is widely studied in statistics that multivariate data relation generally has low-dimensional structure \cite{wang2022low, zhou2022covariance,lazaro2010sparse}. Since, covariance matrix is symmetric and positive definite, we write  $\bm{\Sigma}_{\theta}$ in parametric form \ie,

\begin{equation}
    \bm{\Sigma}_{\theta}(I, I) = \Psi_{\theta}(I) \Psi_{\theta}(I)^T + \sigma^2 \textbf{eye}(N)
    \label{eq:cov_eig},
\end{equation}
where, $\Psi_{\theta}(I) \in \mathbb{R}^{N \times M}$ is learned by deep networks with parameter $\theta$ with $M \lll N$. $\textbf{eye}(N) \in \mathbb{R}^{N \times N}$ is a slang for identity matrix. $\Psi_{\theta}(I) \Psi_{\theta}(I)^T$ is symmetric and $\sigma^2 \textbf{eye}(N)$ guarantees positive definite matrix with $\sigma > 0$ as some positive constant. By using the popular matrix inversion lemma \cite{PresTeukVettFlan92} and Eq.\eqref{eq:cov_eig} parametric form, log probability density defined in Eq.\eqref{eq:log_prob_density} can be re-written as
\begin{equation}
\begin{split}
     \log \Phi(Z|&\theta, I)= -\frac{N}{2}\log 2\pi\sigma^2 -  \frac{1}{2}\log \det (\mathbf{A})  - \\ 
     & \frac{\sigma^{-2}}{2} \mathbf{r}^T\mathbf{r} + \frac{\sigma^{-4}}{2}\mathbf{r}^T\Psi_{\theta}(I)\mathbf{A}^{-1}\Psi_{\theta}(I)^T\mathbf{r},
\end{split}
\label{eq:wood_gaussian}
\end{equation}
with $\mathbf{r} = Z - \bm{\mu}_{\theta}(I)$, and $ \mathbf{A} = \sigma^{-2}\Psi_{\theta}(I)^T\Psi_{\theta}(I) + \textbf{eye}(M)$. It can be shown that the above form for modeling covariance is computationally tractable with complexity $O(NM+M^3)$ \cite{rasmussen2003gaussian} as compared to $O(N^3)$ since $M \lll N$ (refer to supplementary material for details). We use Eq.\eqref{eq:wood_gaussian} as negative log likelihood $(NLL)$ loss function, \ie, $\mathcal{L}_{NLL} = -\log \Phi(Z^{gt}| \theta, I)$ to train the network for learning per-pixel depth, and covariance \wrt all the pixels, hence overcoming the shortcomings with prior works in SIDP.


\begin{figure}[t]
\begin{center}
\begin{subfigure}[b]{0.115\textwidth}
\begin{center}
\includegraphics[width=1.0\textwidth]{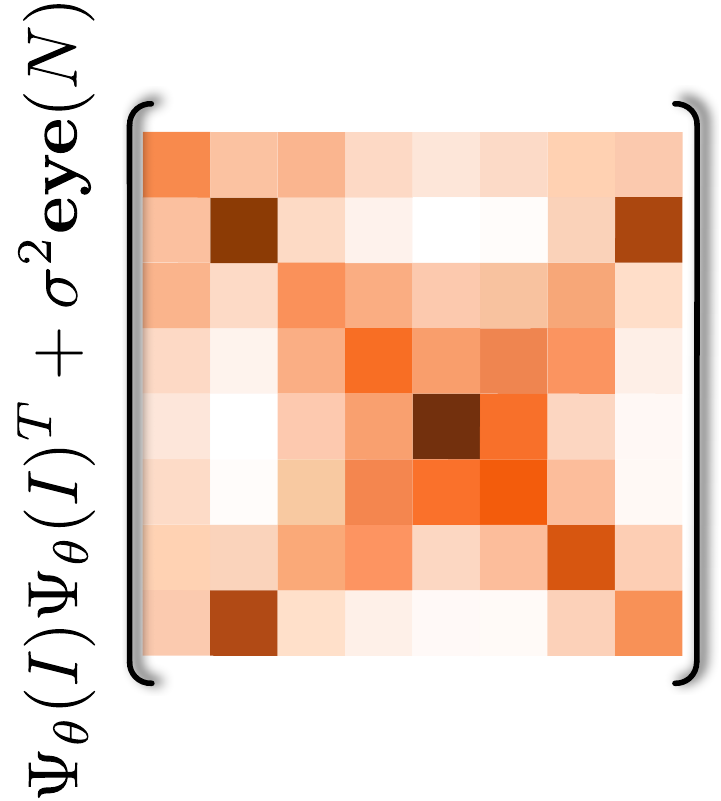}
\caption{Ours $(\mathcal{L}_{NLL})$}
\end{center}
\end{subfigure}
\begin{subfigure}[b]{0.115\textwidth}
\begin{center}
\includegraphics[width=1.0\textwidth]{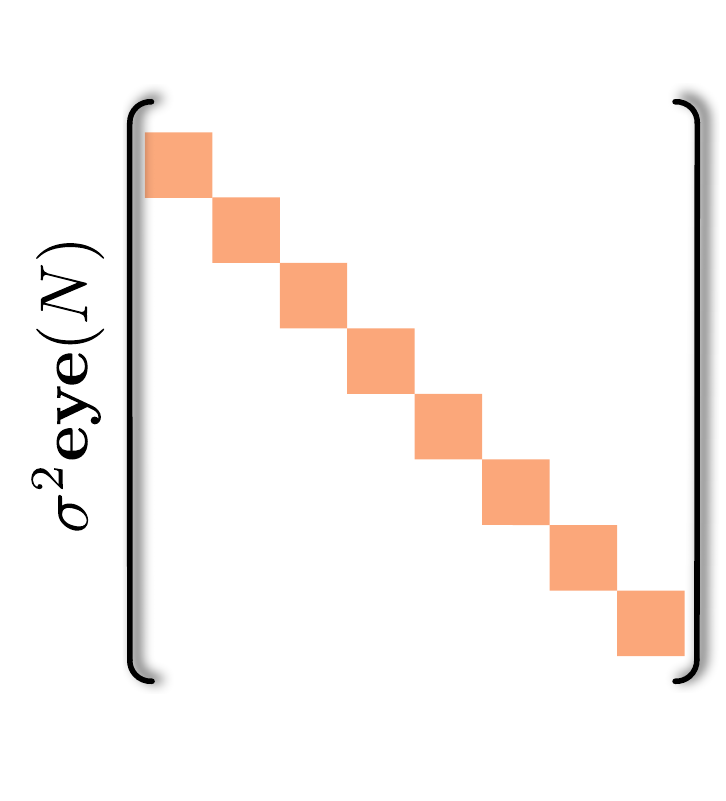}
\caption{L2 Loss}
\end{center}
\end{subfigure}
\begin{subfigure}[b]{0.115\textwidth}
\begin{center}
\includegraphics[width=1.0\textwidth]{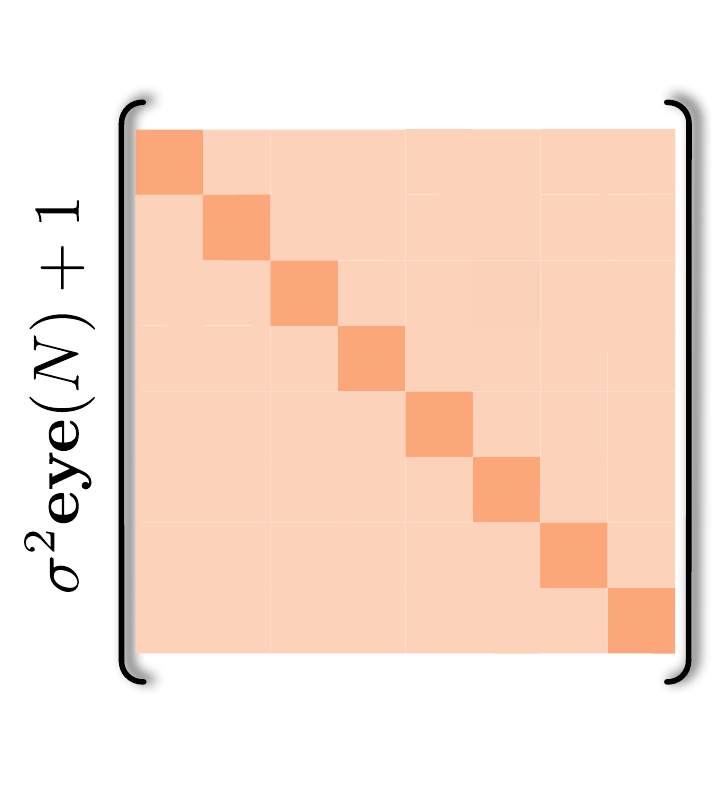}
\caption{SI Loss}
\end{center}
\end{subfigure}
\begin{subfigure}[b]{0.115\textwidth}
\begin{center}
\includegraphics[width=1.0\textwidth]{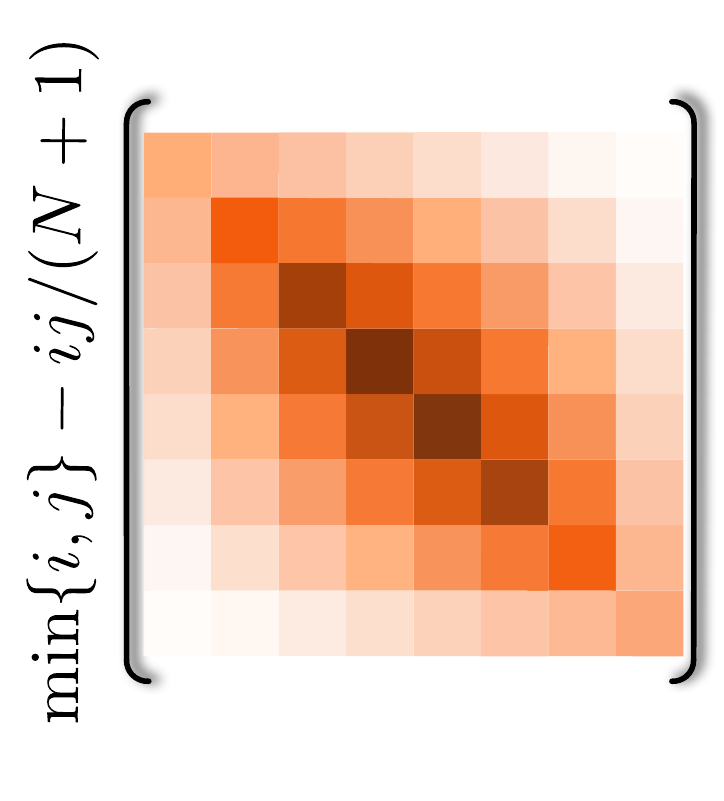}
\caption{G-Loss}
\end{center}
\end{subfigure}
\end{center}
\caption{\textbf{The covariance matrix of loss function.} (a) Ours $\bm{\Sigma}_{\theta}$. (b)-(d) shows the equivalent covariance matrix for the (b) L2 loss, (c) scale invariant loss, and (d) gradient loss. It can be observed that our covariance already contains most, if not all, information that could be recovered by employing different loss functions, hence showing the generality of our formulation.}
\label{fig:covariance_matrix}
\end{figure}

\subsection{Deeper Insights into the Formulation} \label{ssec:other_loss_relation}
A detailed analysis of Eq.\eqref{eq:wood_gaussian} and how it naturally encapsulates the notion of popular loss functions are presented for better understanding. Concretely, we show that ``L2 Loss'', ``Scale Invariant Loss (SI Loss)'', and ``Gradient Loss (G-Loss)'' as a special case of Eq.\eqref{eq:wood_gaussian}; thus, our formulation is more general. Later, in the subsection, we apply Eq.\eqref{eq:wood_gaussian} to well-known Bayesian uncertainty modeling in deep neural networks showing improved uncertainty estimation than independent Gaussian assumption.

\subsubsection{Relation to Popular Loss Function}\label{sss:rpl}
By taking Eq.\eqref{eq:wood_gaussian} $NLL$ form as the training loss, \ie, $-\log \Phi(Z^{gt}| \theta, I)$, we show that using the special values for $\Psi_{\theta}(I)$, the $NLL$ loss can be reduced to some widely-used losses (see Fig. \ref{fig:covariance_matrix}). Here, symbolizes $Z^{gt} \in \mathbb{R}^{N\times1}$.  Denoting $\mathbf{r} = Z^{gt}-\bm{\mu}_{\theta}(I)$, we derive the relation.

\smallskip
\noindent
\textit{(i) \textbf{Case I}.} Substituting $\Psi_{\theta}(I) = \mathbf{0}_N$ in Eq.\eqref{eq:wood_gaussian} will give 
\begin{equation}
-\log \Phi(Z^{gt}|\theta, I) \propto \mathbf{r}^T\mathbf{r},
\end{equation}
which is equivalent to the ``L2 loss'' function. Here, $\mathbf{0}_N$ is a column vector with $N$ elements, all set to 0.

\begin{figure*}[t]
\begin{center}
\includegraphics[width=0.9\linewidth]{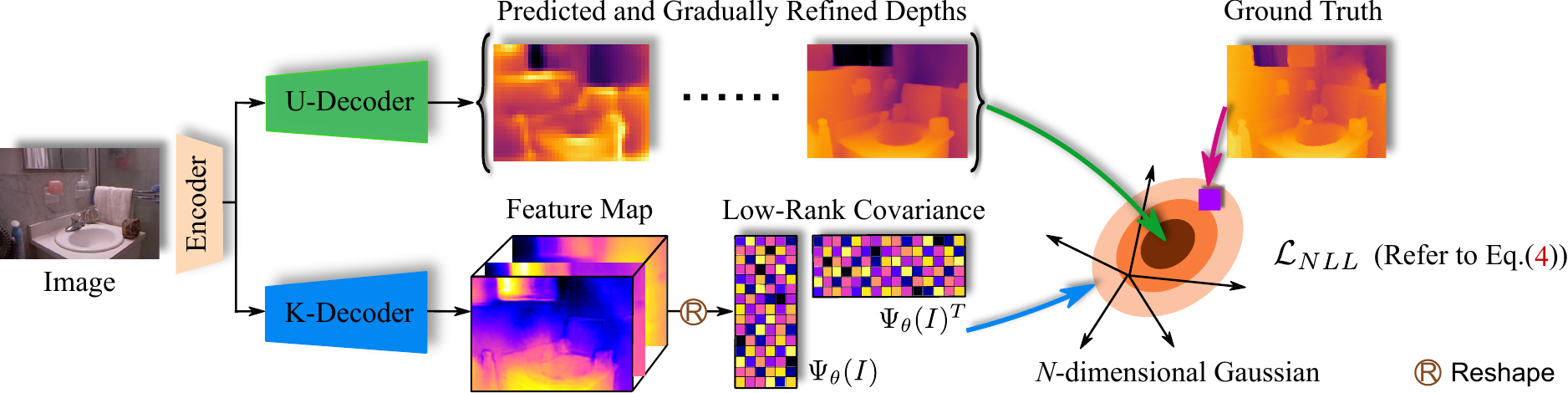}
\end{center}
\caption{\textbf{Overview of our framework}. Given an image, first an encoder is employed to extract features. Then the U-Decoder will predict and gradually refine the depth maps. And the K-Decoder is responsible for predicting the factor $\Psi_{\theta}(I)$ for  modeling the covariance. In the end, we compute the negative log likelihood of the $N$-dimensional Gaussian distribution as the loss function to supervise training.}
\label{fig:arch}
\end{figure*}

\smallskip
\noindent
\textit{(ii) \textbf{Case {II}}.}  Substituting  $\Psi_{\theta}(I) = \mathbf{1}_N$ in Eq.\eqref{eq:wood_gaussian} will give
\begin{equation}\label{eq:gaussian_si}
-\log \Phi(Z^{gt}|\theta, I) \propto \mathbf{r}^T\mathbf{r} - \frac{\alpha}{N} (\mathbf{r}^T\mathbf{1}_N)^2,
\end{equation}
where, $\alpha=(\sigma^{-2}N)/(\sigma^{-2}N+1)$ and $\mathbf{1}_N$ is a column vector with $N$ elements set to 1. Assuming $\sigma^{-2}N \ggg 1$, which is mostly the experimental setting in SIDP, then $\alpha \approx 1$ and Eq.\eqref{eq:gaussian_si} becomes equivalent to ``Scale Invariant Loss''.

\smallskip
\noindent
\textit{(iii) \textbf{Case {III}}.} Here, we want to show the relation between Eq.\eqref{eq:wood_gaussian} and gradient loss. But, unlike previous cases, it's a bit involved. So, for simplicity, consider the gradient of the flattened depth map (\ie, a column vector)\footnote{ignoring border pixels for simple 1D case}. The general squared gradient loss between the ground-truth and predicted depth can be computed as $(\nabla Z^{gt}-\nabla \bm{\mu}_{\theta}(I))^T (\nabla Z^{gt}-\nabla \bm{\mu}_{\theta}(I))$, where $\nabla \in \mathbb{R}^{N \times N}$ is the gradient operator for computing the first order difference of $Z^{gt}$ and $\bm{\mu}_{\theta}(I)$ \cite{renardy2004introduction}. Taking out the common factor, we can re-write the gradient loss as $\big(\nabla(Z^{gt}-\bm{\mu}_{\theta}(I)) \big)^T\big(\nabla(Z^{gt}-\bm{\mu}_{\theta}(I)\big)$. Simplifying using the matrix transpose property and it can be written in compact form as $\mathbf{r}^T(\nabla^T\nabla)\mathbf{r}$, which is equivalent to the Gaussian multivariate form in Eq.\eqref{eq:log_prob_density}. Let's denote $J \triangleq(\nabla^T \nabla)^{-1}$, where $J_{i, j} = \min\{i, j\} - ij / (N+1)$ \cite{dafonseca20017inverse}. However, $J$ is difficult to parameterize and decompose into low-dimensional form. Concretely, we want to factorize $J$  into $\Psi_{\theta}(I)\Psi_{\theta}(I)^T + \sigma^2 \textbf{eye}(N)$ 
that fits the notion developed in Eq.\eqref{eq:log_prob_density} and Eq.\eqref{eq:cov_eig}, with $\Psi_{\theta}(I)\in\mathbb{R}^{N\times M}$ and $M \lll N$.

Fortunately, it is possible to approximate $J$ as $J \approx \big(\Psi_{\theta}(I)\Psi_{\theta}(I)^T + \sigma^2 \textbf{eye}(N)\big)$ by using well-known Eigen approximation \cite{golub2013matrix}. To be precise, setting
$\Psi_{\theta}(I)$ to
\begin{equation}
    \Psi_{\theta}(I)_{k, l}=\sqrt{\bm{\lambda}(J)_l}\mathbf{U}(J)_{k, l}
\end{equation}
where $\bm{\lambda}(J)\in\mathbb{R}^{N \times1}$ and $\mathbf{U}(J)\in\mathbb{R}^{N \times N}$ are the sorted eigenvalues and corresponding eigenvectors of $J$, respectively that can be computed using $\bm{\lambda}(J)_l = (2-2\cos\frac{l \pi}{N + 1})^{-1}$ and $\mathbf{U}(J)_{k, l} = (-1)^{k + 1}\sin\frac{kl\pi}{N + 1}$ \cite{noschese2013tridiagonal}.

\subsubsection{Application in Uncertainty Estimation}\label{sss:rbu}
We apply Eq.\eqref{eq:wood_gaussian} to the popular Bayesian uncertainty modeling in neural networks. Given $\Phi(Z|\theta, I)$ as aleatoric uncertainty for depth map $Z$ \cite{kendall2017uncertainties}, we can compute the Bayesian uncertainty by marginalising over the parameters' posterior distribution using the following well-known equation:
\begin{equation}\label{eq:bayesian_um}
    \Phi(Z|I, \mathcal{D}) = \int \Phi(Z|\theta, I)\Phi(\theta|\mathcal{D}) d{\theta}
\end{equation}
where $\mathcal{D}$ is the train set. The analytic integration of Eq.\eqref{eq:bayesian_um} is difficult to compute in practice, and is usually approximated by Monte Carlo integration, such as ensemble \cite{lakshminarayanan2017simple} and dropout \cite{gal2016uncertainty}. Suppose we have sampled a set of parameters $\Theta\triangleq\{\theta^s\}_{s=1}^S$ from $\Phi(\theta|\mathcal{D})$. The integration is popularly approximated as
\begin{equation}
    \Phi(Z|I, \mathcal{D})=\frac{1}{S}\sum_s \Phi(Z|\theta^s, I).
    \label{eq:uncertainty}
\end{equation}
The $\Phi(Z|I, \mathcal{D})$ denotes the {mixture of Gaussian distributions} \cite{lakshminarayanan2017simple}. The mean and covariance of the distribution is computed as $\bar{\bm{\mu}}(I) = \frac{1}{S}\sum_s \bm{\mu}^{s}(I)$
and $\bar{\bm{\Sigma}}(I, I) = \bar{\Psi}(I)\bar{\Psi}(I)^T + \sigma^2\textbf{eye}(N)$, respectively \cite{weiss2006course}, which in fact has the same form as Eq.\eqref{eq:cov_eig}, where we compute $\bar{\Psi}$ using the following expression
\begin{equation}
    \bar{\Psi} = \frac{1}{\sqrt S}\mathbf{concat}(\Psi^1,\ldots,\Psi^S, \bm{\mu}^1-\bar{\bm{\mu}},\ldots,\bm{\mu}^S-\bar{\bm{\mu}}).
\end{equation}
So, from the derivations in Sec.(\ref{sss:rpl}) and Sec.(\ref{sss:rbu}), it is quite clear that our proposed Eq.\eqref{eq:wood_gaussian} is more general and encapsulates flavors of popular loss functions widely used in deep networks. For the SIDP problem we need such a loss function for deep neural network parameters learning. Next, we discuss the  implementation in our proposed pipeline and usefulness of our introduced loss function.

\subsection{Overall Pipeline}
To keep our pipeline description simple, let's consider the image and depth map in 2D form instead of a column vector. For brevity, we slightly abuse the notation hereafter. Here, we use the same notation we defined for the 1D Gaussian distribution case for simplicity. For a better understanding of our overall pipeline, we provide architectural implementation details following Fig.(\ref{fig:arch}) blueprint, \ie, \textit{(i)} Encoder details, \textit{(ii)} Decoder details, followed by \textit{(iii)} Train and test time settings. 

\begin{figure*}[t]
\begin{center}
\begin{subfigure}[b]{0.15\textwidth}
\begin{center}
\begin{minipage}[b]{1.0\textwidth}
\includegraphics[width=1.0\textwidth]{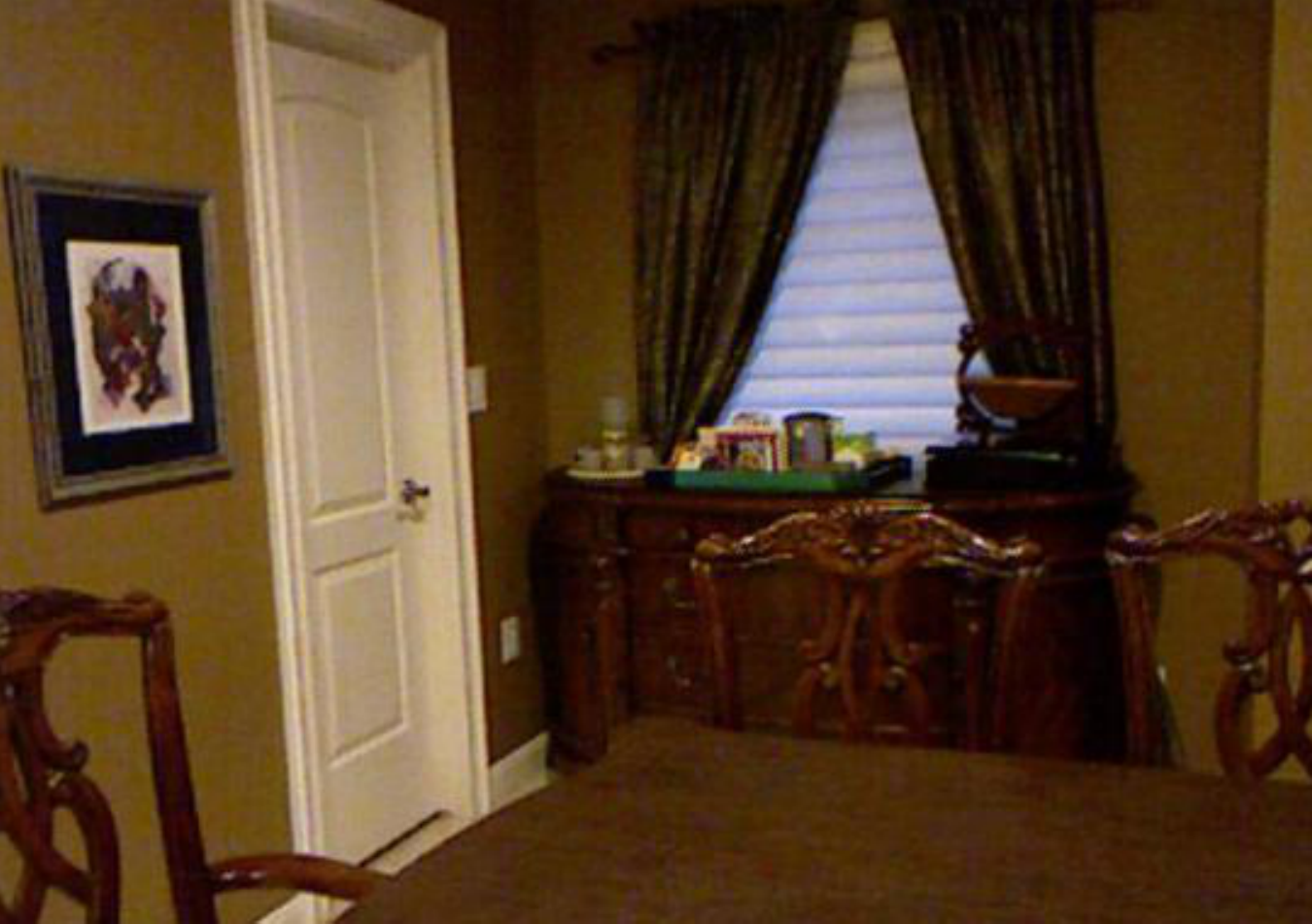}
\includegraphics[width=1.0\textwidth]{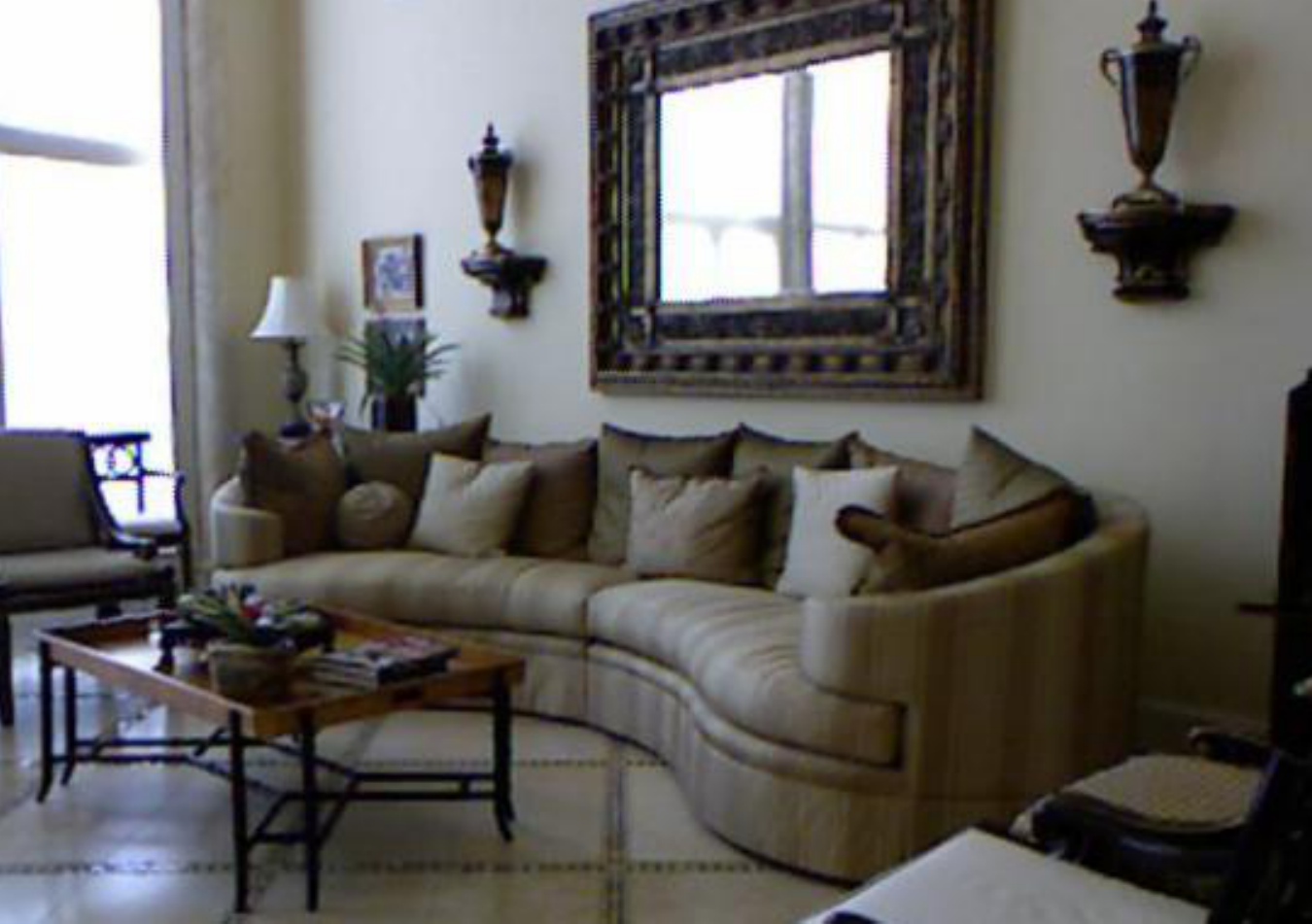}
\end{minipage}
\caption{Test}
\end{center}
\end{subfigure}
\begin{subfigure}[b]{0.15\textwidth}
\begin{center}
\begin{minipage}[b]{1.0\textwidth}
\includegraphics[width=1.0\textwidth]{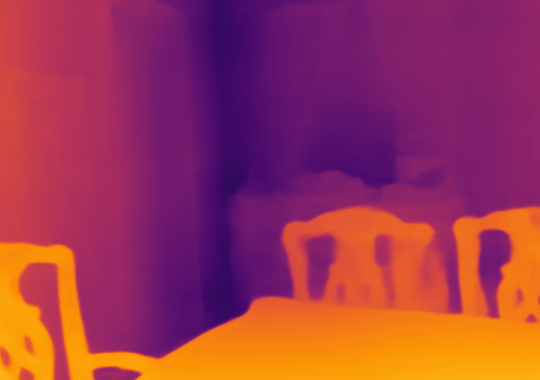}
\includegraphics[width=1.0\textwidth]{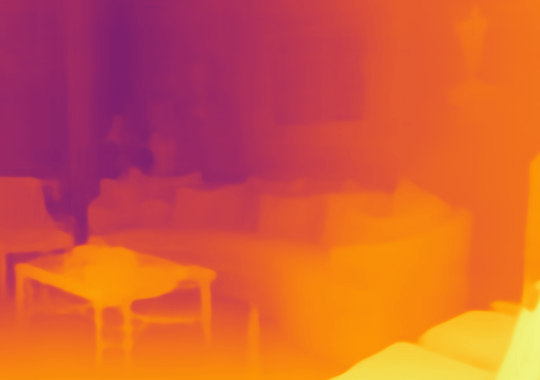}
\end{minipage}
\caption{DPT\cite{ranftl2021vision}}
\end{center}
\end{subfigure} 
\begin{subfigure}[b]{0.15\textwidth}
\begin{center}
\begin{minipage}[b]{1.0\textwidth}
\includegraphics[width=1.0\textwidth]{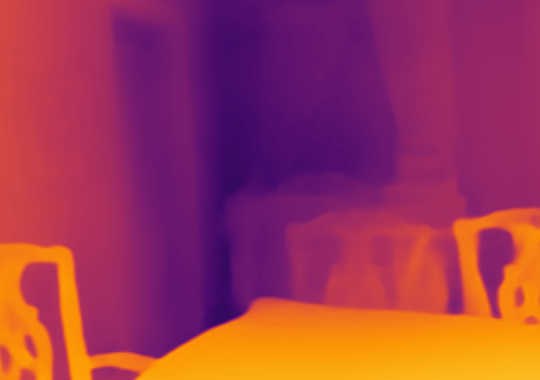}
\includegraphics[width=1.0\textwidth]{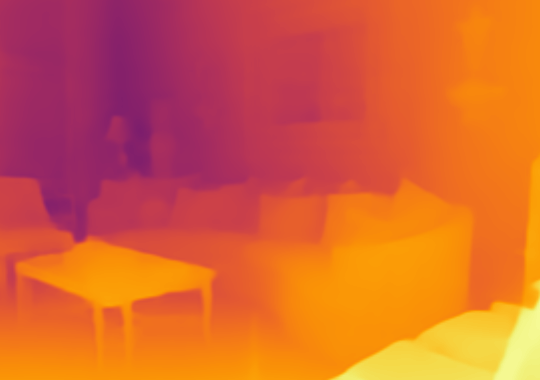}
\end{minipage}
\caption{AdaBins\cite{bhat2021adabins}}
\end{center}
\end{subfigure}  
\begin{subfigure}[b]{0.15\textwidth}
\begin{center}
\begin{minipage}[b]{1.0\textwidth}
\includegraphics[width=1.0\textwidth]{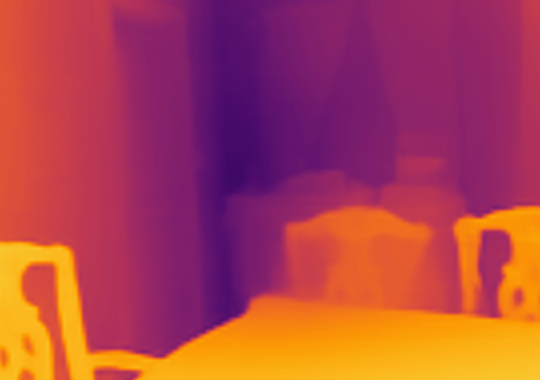}
\includegraphics[width=1.0\textwidth]{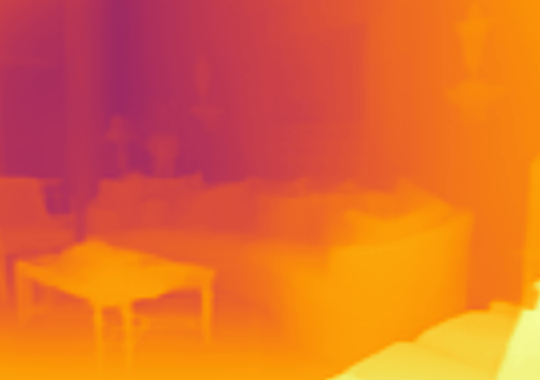}
\end{minipage}
\caption{NeWCRFs\cite{yuan2022new}}
\end{center}
\end{subfigure}
\begin{subfigure}[b]{0.15\textwidth}
\begin{center}
\begin{minipage}[b]{1.0\textwidth}
\includegraphics[width=1.0\textwidth]{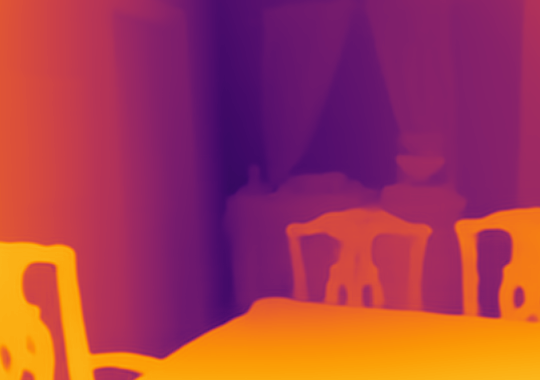}
\includegraphics[width=1.0\textwidth]{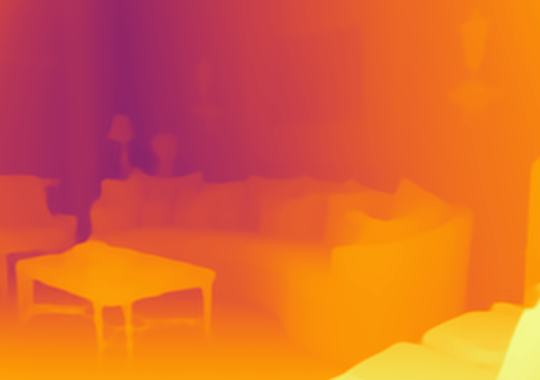}
\end{minipage}
\caption{\textbf{Ours}}
\end{center}
\end{subfigure}
\begin{subfigure}[b]{0.15\textwidth}
\begin{center}
\begin{minipage}[b]{1.0\textwidth}
\includegraphics[width=1.0\textwidth]{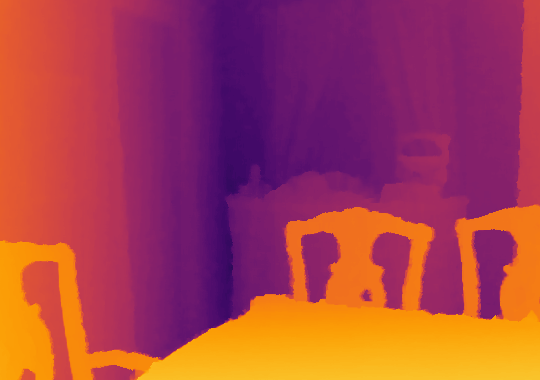}
\includegraphics[width=1.0\textwidth]{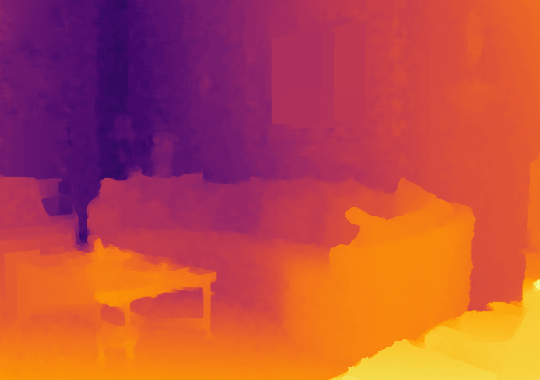}
\end{minipage}
\caption{Ground Truth}
\end{center}
\end{subfigure}
\end{center}
\caption{\textbf{Qualitative Comparison}. Our method recovers better depth even for complex scenes. Our depth results even qualitatively looks closer to the ground truth than the prior art such as (b) DPT \cite{ranftl2021vision}, (c) AdaBins \cite{bhat2021adabins}, (d) NeWCRFs \cite{yuan2022new} on NYU Depth V2 test set \cite{silberman2012indoor}.}
\vspace{-4mm}
\label{fig:quality_nyu}
\end{figure*}

\smallskip
\formattedparagraph{\textit{(i)} Encoder Details.} Our encoder takes the image $\mathbf{I}\in\mathbb{R}^{m \times n}$ as input and gives a hierarchical feature maps  $\mathbf{F}=\{\mathbf{F}^1, \mathbf{F}^2, \mathbf{F}^3, \mathbf{F}^4\}$ as output, where $\mathbf{F}^i\in\mathbb{R}^{C^i\times m^{i}\times n^{i}}$ denotes a feature map with channels $C^i$, and resolution $m^{i} \times n^{i}$. We adopt the Swin-Large \cite{liu2021swin} as the encoder. Specifically, it includes four stages of non-linear transformation to extract features from the input image, where each stage contains a series of transformer blocks to learn non-linear transformation and a downsampling block to reduce the resolution of the feature map by 2. We collect the output feature map from the last block of the $i$-th stage as  $\mathbf{F}_i$.

\smallskip
\formattedparagraph{\textit{(ii)} Decoder Details.}
The \emph{U-decoder} (see Fig.\ref{fig:arch}) estimates a set of depth maps $\{\bm{\mu}^i_{\theta}(\mathbf{I})\}_{i=1}^4$. The U-decoder first estimates $\bm{\mu}^4_{\theta}(\mathbf{I})$ only from $\mathbf{F}^4$ by a convolution layer, then upsamples and refines the depth map in a hierarchical manner. At the $i$-th stage, where $i$ decreases from 3 to 1, we first concatenate $\bm{\mu}^{i+1}_{\theta}(\mathbf{I})$ and $\mathbf{F}^{i+1}$ from the previous stage, and then feed into a stack of convolution layers to refine the depth map and feature map. The refined depth map is upsampled via bi-linear interpolation to double the resolution, and denoted as $\bm{\mu}^{i}_{\theta}(\mathbf{I})$. Similarly, the refined feature map is upsampled and added to $\mathbf{F}^{i}$. In the end, we upsample all the depth maps $\{\bm{\mu}^i_{\theta}(\mathbf{I})\}_{i=1}^4$ into $m\times n$ resolution via bi-linear interpolation as the final output of the U-decoder. 

The \emph{K-decoder} (see Fig. \ref{fig:arch}) estimates $\Psi_{\theta}(\mathbf{I})$. It first upsamples and refines the feature maps in $\mathbf{F}$. Specifically, at the $i$-th stage, where $i$ decreases from 3 to 1, it upsamples $\mathbf{F}^{i+1}$ from the previous stage and adds to the $\mathbf{F}^{i}$. We utilize a stack of convolution layers to further refine the added feature map. In the end, we upsample the refined feature map $\mathbf{F}^1$ to $m \times n$ resolution by bi-linear interpolation, and predict the $\Psi_{\theta}(\mathbf{I})$ by a convolution layer. 

\smallskip
\formattedparagraph{\textit{(iii)} Train and Test Time Setting.}
At train time, we collect $\{\bm{\mu}^i_{\theta}(\mathbf{I})\}_{i=1}^4$, and $\Psi_{\theta}(\mathbf{I})$ and compute loss using our proposed loss function (refer to Sec. \ref{ssec:loss_func}). At test time, our approach provides $\bm{\mu}^1_{\theta}(\mathbf{I})$ as the final depth map prediction. Furthermore, we can query $\Psi_{\theta}(\mathbf{I})$ to infer the distribution of the depth map if necessary depending on the application.

\subsection{Loss Function}\label{ssec:loss_func}
As shown in Sec. \ref{sss:rpl}, the negative log likelihood loss can approximate the scale invariant loss and the gradient loss when $\Psi_{\theta}(I)$ and $\sigma$ take special values. Consequently, we propose the following overall loss function:
\begin{equation}
    \mathcal{L}_{total} =\sum_{j=1}^4 \mathcal{L}^{j}_{NLL} + \frac{1}{N}\sum_i(\bm{\mu}_{\theta}^1(I)_i-Z^{gt}_i)^2
    \label{eq:total_loss}
\end{equation}
where $\mathcal{L}^{j}_{NLL}$ is the negative log likelihood loss applying to $\bm{\mu}_{\theta}^j(I)$ and $\Psi_{\theta}(I)$. Note, however, Eq.\eqref{eq:total_loss} second term is optional. Yet, it is added to provide train time improvement. 

\begin{table*}[th]
\begin{center}
\scriptsize
\begin{tabular*}{1.0\textwidth}{l@{\extracolsep{\fill}}ccccccc}
\hline
Method & Backbone & SILog $\downarrow$ & Abs Rel $\downarrow$& RMS$\downarrow$ & RMS log$\downarrow$ &  $\delta_1$ $\uparrow$ \\
\hline
GeoNet\cite{qi2018geonet} & ResNet-50 & - & 0.128 & 0.569 & - & 0.834 \\
DORN \cite{fu2018deep} & ResNet-101 & - & 0.115 & 0.509 & - & 0.828  \\
VNL\cite{yin2019enforcing} & ResNeXt-101 & - & 0.108 & 0.416 & - & 0.875 \\
TransDepth \cite{yang2021transformer} & ViT-B & - & 0.106 & 0.365 & - & 0.900 \\
ASN \cite{long2021adaptive} & HRNet-48 & - & 0.101 & 0.377 & - & 0.890  \\
BTS \cite{lee2019big} & DenseNet-161 & 11.533 & 0.110 & 0.392 & 0.142 & 0.885  \\
DPT-Hybrid \cite{ranftl2021vision} & ViT-B & 9.521 & 0.110 & 0.357 & 0.129 & 0.904  \\
AdaBins \cite{bhat2021adabins} & EffNet-B5+ViT-mini & 10.570 & 0.103 & 0.364 & 0.131 & 0.903  \\
ASTrans \cite{chang2021transformer} & ViT-B & 10.429 & 0.103 & 0.374 & 0.132 & 0.902  \\
NeWCRFs \cite{yuan2022new} & Swin-L & 9.102 & 0.095 & 0.331 & 0.119 & 0.922  \\
\hline
\textbf{Ours} & Swin-L & \textbf{8.323}  & \textbf{0.087} & \textbf{0.311} & \textbf{0.110} & \textbf{0.933} \\
\textbf{\% Improvement} & &
\textcolor{ao}{{8.56\%}} & \textcolor{ao}{{8.42\%}} & \textcolor{ao}{{6.04\%}} & \textcolor{ao}{{7.56\%}} & \textcolor{ao}{{1.18\%}} \\
\hline
\end{tabular*}
\caption{Comparison with the state-of-the-art methods on the NYU test set \cite{silberman2012indoor}. Please refer to Sec. \ref{sec:exp_sota} for details.
}\label{tab:nyu_sota}
\vspace{-6mm}
\end{center}
\end{table*}

\begin{table}[th]
\begin{center}
\scriptsize
\begin{tabular*}{0.5\textwidth}{l@{\extracolsep{\fill}}cccc}
\hline
Method & SILog$\downarrow$ & Abs Rel$\downarrow$ & Sq Rel$\downarrow$ & iRMS$\downarrow$\\
\hline
DLE \cite{liu2021deep} &  11.81 &   9.09 &  2.22 & 12.49\\
DORN \cite{fu2018deep} & 11.80 &  8.93  &  2.19 & 13.22 \\
BTS \cite{lee2019big} &  11.67 & 9.04 &  2.21 & 12.23 \\
BANet \cite{aich2021bidirectional} &  11.55 &  9.34 &  2.31 & 12.17 \\
PWA \cite{lee2021patch} &  11.45 &   9.05 &  2.30 & 12.32 \\
ViP-DeepLab \cite{qiao2021vip} & 10.80 &   8.94 &  2.19 & 11.77 \\
NeWCRFs \cite{yuan2022new} & 10.39 & 8.37 & 1.83 & 11.03 \\
\hline
\textbf{Ours}  &\textbf{9.93} &  \textbf{7.99} &  \textbf{1.68} & \textbf{10.63} \\
\textbf{\% Improvement} &   \textcolor{ao}{{4.43\%}} &  \textcolor{ao}{{4.54\%}} &  \textcolor{ao}{{8.20\%}} & \textcolor{ao}{{3.63\%}} \\
\hline
\end{tabular*}
\caption{Comparison with the state-of-the-art methods on the the KITTI official test set \cite{geiger2012we}. We only list the results from the published methods. 
}\label{tab:kitti_off_sota}
\vspace{-6mm}
\end{center}
\end{table}

\section{Experiments and Results}
\paragraph{Implementation Details.} We implemented our framework in PyTorch 1.7.1 and Python 3.8 with CUDA 11.0. All the experiments and statistical results shown in the draft are simulated on a computing machine with Quadro RTX 6000 (24GB Memory) GPU support. 
We use evaluation metrics including SILog, Abs Rel, RMS, RMS log, $\delta_i$, Sq Rel, iRMS to report our results on the benchmark dataset. For exact definition of the metrics we refer to \cite{lee2019big}.

\begin{table*}
\begin{center}
\scriptsize
\begin{tabular*}{1.0\textwidth}{l@{\extracolsep{\fill}}ccccccc}
\hline
Method & Backbone & SILog $\downarrow$ & Abs Rel $\downarrow$& RMS$\downarrow$ & RMS log$\downarrow$ &  $\delta_1$ $\uparrow$ \\
\hline
DORN \cite{fu2018deep} & ResNet-101 &  - & 0.072 & 0.273 & 0.120 & 0.932  \\
VNL \cite{yin2019enforcing} & ResNeXt-101 & - & 0.072 & 0.326 & 0.117 & 0.938 \\
TransDepth \cite{yang2021transformer} & ViT-B & 8.930 & 0.064 & 0.275 & 0.098 & 0.956  \\
BTS \cite{lee2019big} & DenseNet-161 & 8.933 & 0.060 & 0.280 & 0.096 & 0.955  \\
DPT-Hybrid \cite{ranftl2021vision} & ViT-B & 8.282 & 0.062 & 0.257 & 0.092 & 0.959  \\
AdaBins \cite{bhat2021adabins} & EffNet-B5+ViT-mini & 8.022 & 0.058 & 0.236 & 0.089 & 0.964  \\
ASTrans \cite{chang2021transformer} & ViT-B & 7.897 & 0.058 & 0.269 & 0.089 & 0.963  \\
NeWCRFs \cite{yuan2022new} & Swin-L & 6.986 & 0.052 & 0.213 & 0.079 & 0.974 \\
\hline
\textbf{Ours} & Swin-L& \textbf{6.757}  & \textbf{0.050} & \textbf{0.202} & \textbf{0.075} & \textbf{0.976} \\
\textbf{\% Improvement} & & \textcolor{ao}{3.28\%} & \textcolor{ao}{3.85\%} & \textcolor{ao}{5.16\%} & \textcolor{ao}{5.06\%} & \textcolor{ao}{0.21\%}  \\
\hline
\end{tabular*}
\caption{Comparison with the state-of-the-art methods on the KITTI Eigen test set \cite{eigen2014depth}. 
Please refer to Sec. \ref{sec:exp_sota} for details.}\label{tab:eigen_sota}
\vspace{-3mm}
\end{center}
\end{table*}

\begin{table*}
\begin{center}
\scriptsize
\begin{tabular*}{1.0\textwidth}{l@{\extracolsep{\fill}}ccccccc}
\hline
Method & Backbone & SILog $\downarrow$ & Abs Rel $\downarrow$& RMS$\downarrow$ & RMS log$\downarrow$ &  $\delta_1$ $\uparrow$\\
\hline
AdaBins\cite{bhat2021adabins}& EffNet-B5+ViT-mini & 13.652 & 0.110 & 0.321 & 0.137 & 0.906\\
NeWCRFs \cite{yuan2022new} & Swin-L& 13.695 & 0.105 & 0.322 & 0.138 & 0.920 \\
\hline
\textbf{Ours} & Swin-L & \textbf{11.985}  & \textbf{0.090} & \textbf{0.282} & \textbf{0.120} & \textbf{0.936} \\
\textbf{\% Improvement} & &
\textcolor{ao}{12.49\%} & \textcolor{ao}{14.29\%} & \textcolor{ao}{12.42\%}
& \textcolor{ao}{13.04\%} & 
\textcolor{ao}{1.74\%}  \\
\hline
\end{tabular*}\caption{Comparison with AdaBins \cite{bhat2021adabins} and NeWCRFs \cite{yuan2022new} on SUN RGB-D test set \cite{song2015sun}. All methods are trained on NYU Depth V2 train set {without} fine-tuning on SUN RGB-D. Please refer to Sec. \ref{sec:exp_sota} for details.}
\label{tab:sun_sota}
\vspace{-3mm}
\end{center}
\end{table*}

\smallskip
\formattedparagraph{Datasets.} We performed experiments and statistical comparisons with the prior art on benchmark datasets such as NYU Depth V2 \cite{silberman2012indoor}, KITTI \cite{geiger2012we}, and SUN RGB-D \cite{song2015sun}. 

\smallskip
\noindent
\textbf{(a) NYU Depth V2:} This dataset contains images of indoor scenes with $480 \times 640$ resolution  \cite{silberman2012indoor}. We follow the standard train and test split setting used by previous works for experiments \cite{lee2019big}. Precisely, we use $24,231$ image-depth pairs for training the network and $654$ images for testing the performance. Note that the depth map evaluation for this dataset has an upper bound of $10$ meters. 

\smallskip
\noindent
\textbf{(b) KITTI:} This dataset contains images and depth data of outdoor driving scenarios. The official experimental split contains $42,949$ training images, $1,000$ validation images, and $500$ test images with $352 \times 1216$ resolution \cite{geiger2012we}. Here, the depth map accuracy can be evaluated up to an upper bound of $80$ meters. In addition, there are few works following the split from Eigen \cite{eigen2014depth}, which includes $23, 488$ images for training and $697$ images for the test. 

\smallskip
\noindent
\textbf{(c) SUN RGB-D:} It contains data of indoor scenes captured by different cameras \cite{song2015sun}. The depth values range from $0$ up to $10$ meters. The images are resized to $480 \times 640$ resolution for consistency. We use the official test set \cite{song2015sun} of 5050 images to evaluate the generalization of the frameworks.

\smallskip
\formattedparagraph{Training Details.} We use Adam optimizer \cite{kingma2014adam} to minimize our proposed loss function and learn network parameters. At train time, the learning rate is decreased from $3e^{-5}$ to $1e^{-5}$ by the cosine annealing scheduler. Our encoder--which is inspired from \cite{liu2021swin}, is initialized by pre-training the network on ImageNet \cite{deng2009imagenet}. For the KITTI dataset, we train our framework for 10 and 20 epochs on the official split \cite{geiger2012we} and Eigen \cite{eigen2014depth} split, respectively. For the NYU dataset \cite{silberman2012indoor}, our framework is trained for 20 epochs. We randomly apply horizontal flipping on the image and depth map pair at train time for data augmentation.

\subsection{Performance Comparison with Prior Works}
\label{sec:exp_sota}
\cref{tab:nyu_sota}, \cref{tab:kitti_off_sota}, \cref{tab:eigen_sota}, and \cref{tab:sun_sota} show our method's statistical performance comparison with popular state-of-the-art (SOTA) methods on  NYU Depth V2 \cite{silberman2012indoor}, KITTI official \cite{geiger2012we} and Eigen \cite{eigen2014depth} split, and SUN RGB-D \cite{song2015sun}. From the tables, it is easy to infer that our approach consistently performs better on all the popular evaluation metrics. The percentage improvement over the previous SOTA is indicated in green for better exposition. In particular, on the NYU test set, which is a challenging dataset, we reduce the SILog error from the previous best result of 9.102 to 8.323 and increase $\delta_1$ metric from 0.922 to 0.933\footnote{At the time of submission, our method's performance on the KITTI official leaderboard was the best among all the published works.}. Fig. \ref{fig:quality_nyu} shows qualitative comparison results. It can be observed that our method's predicted depth is better at low and high-frequency scene details. 
For the SUN RGB-D test set, all competing models, including ours, are trained on the NYU DepthV2 train set without fine-tuning on SUN RGB-D \cite{song2015sun}. In addition, we align the predictions from all the models with the ground truth by a scale and shift following \cite{Ranftl2020}. \cref{tab:sun_sota} results indicate our method’s better generalization capability than other approaches. More results are provided in the supplementary material.

\subsection{Bayesian Uncertainty Estimation Comparison}
\label{sec:exp_uncertainty}
\begin{figure}[h]
\begin{centering}
\includegraphics[width=1.0\linewidth]{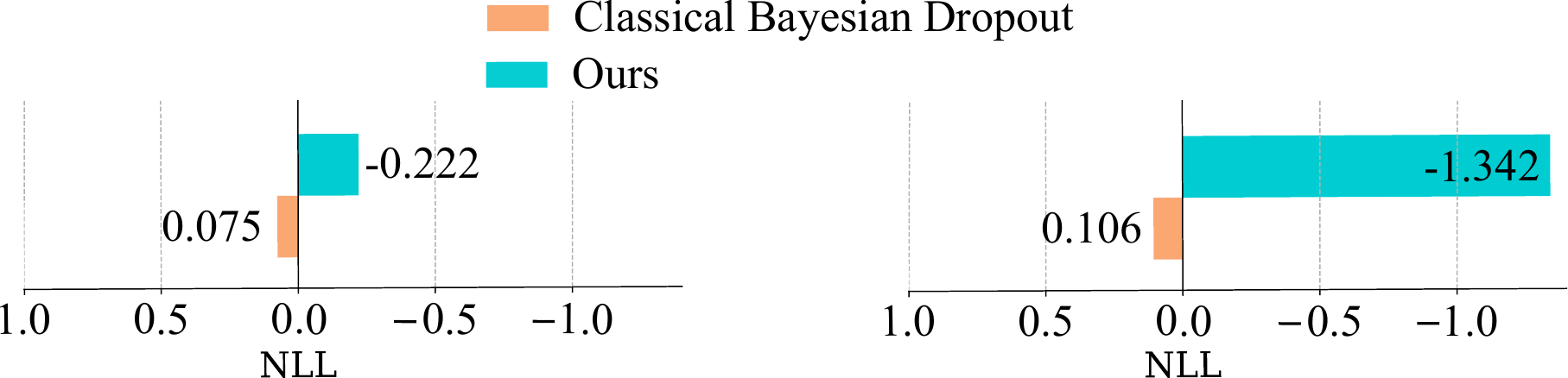}
\end{centering}
\caption{Comparison with the classical Bayesian dropout for uncertainty estimation. The left and right figures present the Negative Log Likelihood (NLL) of the predicted depth map distribution on KITTI Eigen \cite{geiger2012we,eigen2014depth} split and NYU test set \cite{silberman2012indoor} respectively. Our multivariate Gaussian distribution achieves lower NLL than the independent Gaussian distribution in classical Bayesian dropout.
}
\vspace{-2mm}
\label{fig:exp_uncertainty}
\end{figure}

In this part, we compare with the classical Bayesian dropout \cite{kendall2017uncertainties}, which uses independent Gaussian distribution to quantify uncertainty. As for our approach, we also use dropout to sample multiple depth predictions, and compute the negative log likelihood following the distribution in Eq.\eqref{eq:uncertainty}. More specifically, in each block of Swin transformer \cite{liu2021swin}, we randomly drop feature channels before the layer normalization \cite{ba2016layer} operation with probability $0.01$. We first sample $S=10$ predictions for each test image, then compute the mean and covariance of the mixture of Gaussian distributions in Eq.\eqref{eq:uncertainty}, and further approximate the entire distribution as single Gaussian following \cite{lakshminarayanan2017simple}. We present the comparison results of the negative log likelihood in \cref{fig:exp_uncertainty}. Our multivariate Gaussian distribution achieves much lower negative log likelihood cost.

\subsection{Ablations and Further Analysis}
To better understand our introduced approach, we performed ablations on the NYU Depth V2 dataset \cite{silberman2012indoor} and studied our trained model's inference time and memory footprint for its practical suitability.

\smallskip
\formattedparagraph{\textit{(i)} Effect of NLL Loss.} To realize the significance of NLL loss in Eq.\eqref{eq:total_loss}, we replaced it with $L2$ loss, SI loss \cite{eigen2014depth}, gradient loss, and virtual normal loss \cite{yin2019enforcing} one by one, while keeping the remaining term in Eq.\eqref{eq:total_loss} fixed. The statistical results are shown in \cref{tab:exp_loss}. The stats show that our proposed NLL loss achieves the best performance over all the widely used metrics.

\begin{table}[th]
\begin{center}
\scriptsize
\begin{tabular*}{0.5\textwidth}{l@{\extracolsep{\fill}}ccccccc}
\hline
Loss & SILog $\downarrow$ & Abs Rel $\downarrow$& RMS $\downarrow$ & $\delta_1$ $\uparrow$ \\
\hline
L2 & 8.912 & 0.090 & 0.324 & 0.929 \\
SI \cite{eigen2014depth} & 8.762 & 0.089 & 0.322 & 0.929\\
Gradient  & 8.886 & 0.090 & 0.323 & 0.929 \\
VNL \cite{yin2019enforcing} & 8.543 & 0.090 & 0.325 & 0.926\\
\hline
\textbf{Ours} & \textbf{8.323} & \textbf{0.087} & \textbf{0.311} & \textbf{0.933}\\
\hline
\end{tabular*}
\caption{Comparison of our NLL loss function with widely used loss functions for solving SIDP task.}
\label{tab:exp_loss}
\vspace{-3mm}
\end{center}
\end{table}

\begin{figure}[t]
\begin{center}
\begin{subfigure}[b]{0.162\textwidth}
\begin{center}
\includegraphics[width=1.0\textwidth]{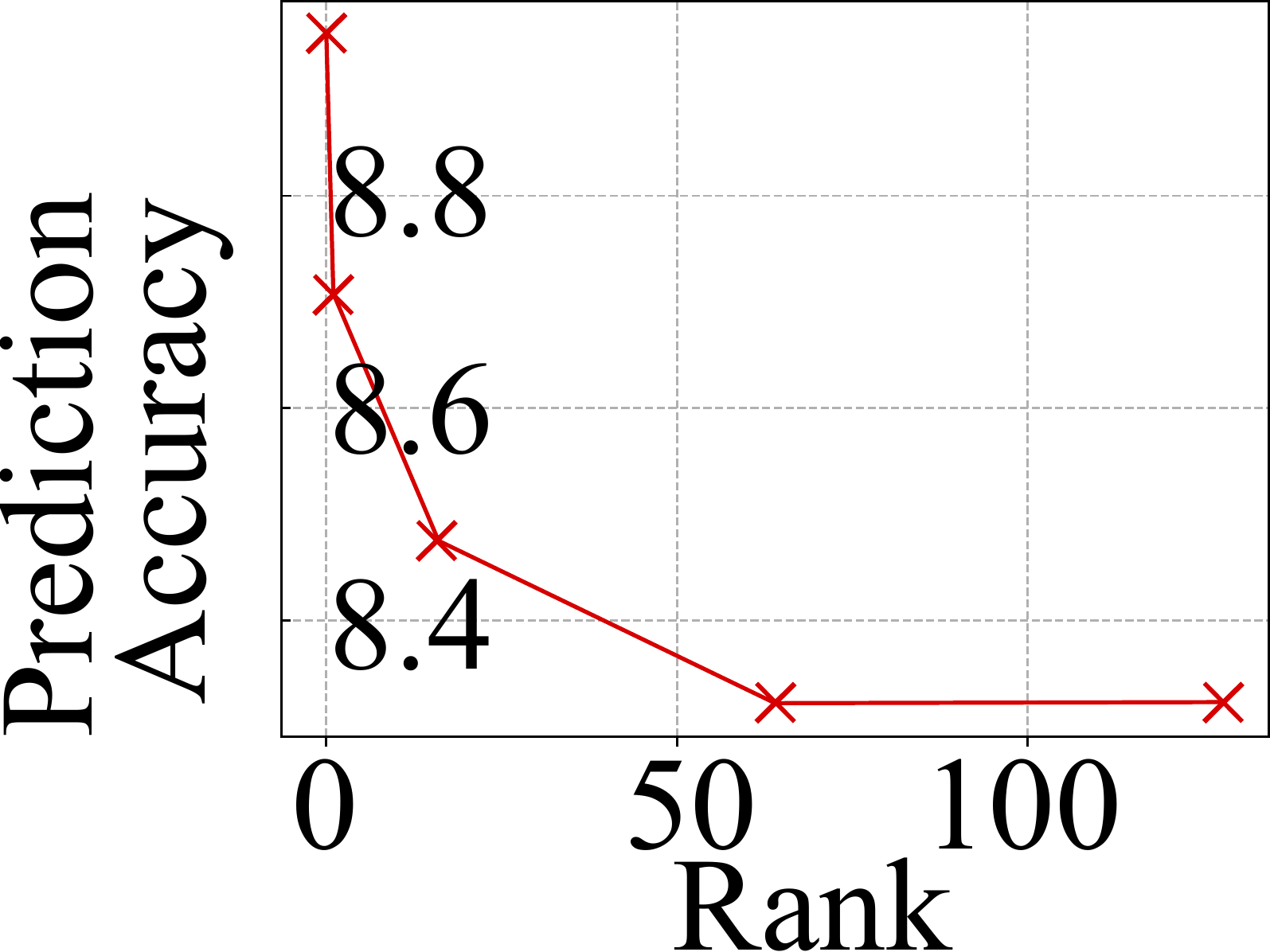}
\caption{SILog Metric}
\end{center}
\end{subfigure}
\begin{subfigure}[b]{0.14\textwidth}
\begin{center}
\includegraphics[width=0.9\textwidth]{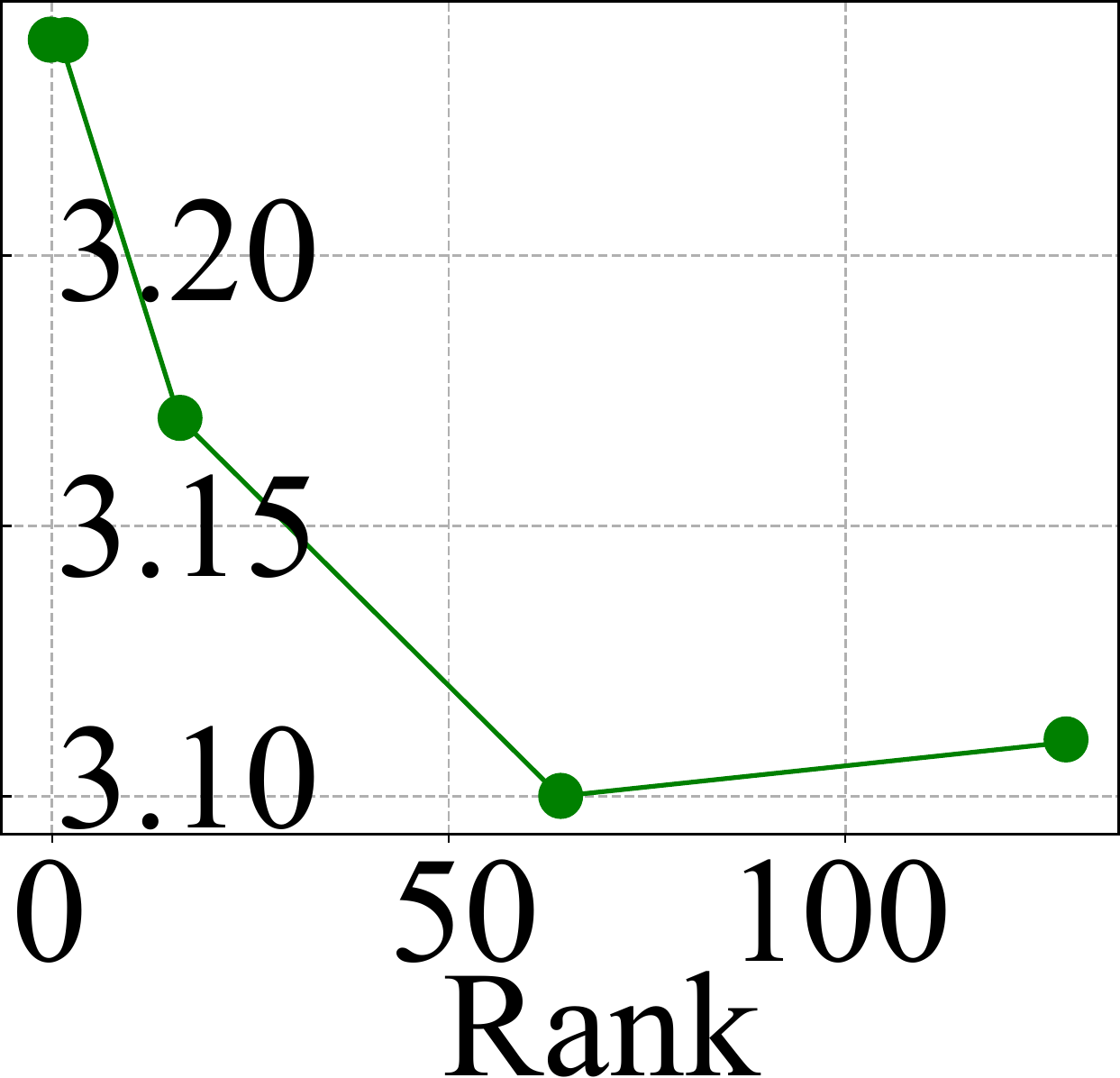}
\caption{RMS Metric $\times10$}
\end{center}
\end{subfigure} 
\begin{subfigure}[b]{0.14\textwidth}
\begin{center}
\includegraphics[width=0.9\textwidth]{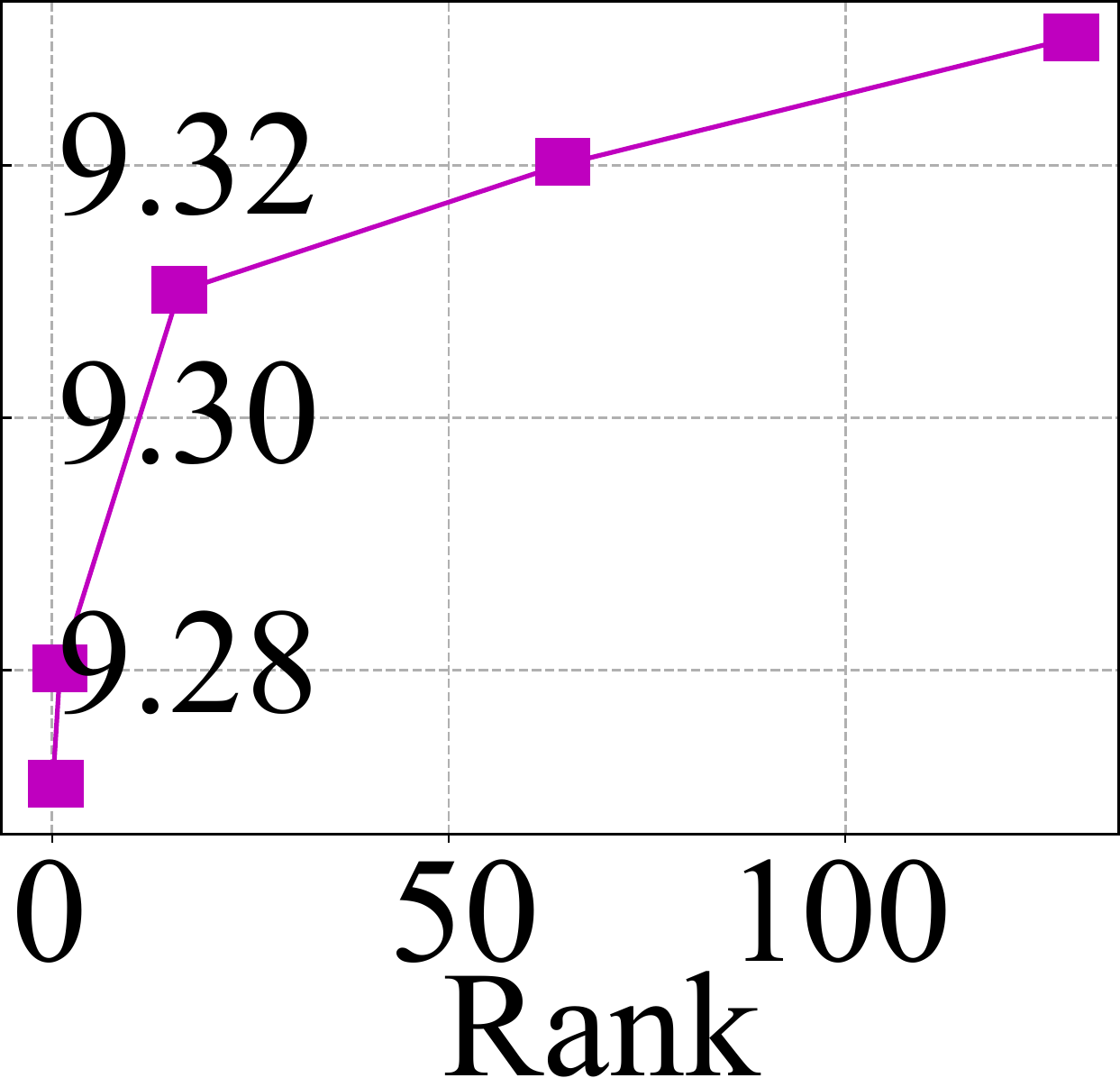}
\caption{$\delta_1$ Metric$\times10$}
\end{center}
\end{subfigure} 
\end{center}
\caption{Depth prediction accuracy of our method using different evaluation metrics w.r.t change in the rank of the covariance matrix. The increase in the rank improves prediction accuracy and shows saturation at 128, thereby showing the effectiveness of our low-dimensional modeling.}\label{fig:exp_rank}
\vspace{-3mm}
\end{figure}

\smallskip
\formattedparagraph{\textit{(ii)} Performance with the change in the Rank of Covariance.}
We vary the rank of $\Psi_{\theta}(I)$, and observe our method's prediction accuracy. We present the accuracy under various evaluation metrics in Fig. \ref{fig:exp_rank}. With increase in the rank, the distribution is better approximated, and the performance improves, but saturates later showing the suitability of its low-dimensional representation.

\smallskip
\formattedparagraph{\textit{(iii)} Evaluation on NeWCRFs.}
We evaluate our loss function on NeWCRFs \cite{yuan2022new} network design using their proposed training strategies. The depth prediction accuracy is shown in Tab. \ref{tab:newcrfs}. The results convincingly indicate the benefit of our proposed loss on a different SIDP network.
\begin{table}[th]
\begin{center}
\scriptsize
\begin{tabular*}{0.5\textwidth}{l@{\extracolsep{\fill}}ccccccc}
\hline
Method & SILog $\downarrow$ & Abs Rel $\downarrow$& RMS $\downarrow$ & $\delta_1$ $\uparrow$ \\
\hline
NeWCRFs & 9.102 & 0.095 & 0.331 & 0.922 \\
\textbf{+Our Loss} & \textbf{8.619} & \textbf{0.086} & \textbf{0.316} & \textbf{0.935}\\
\hline
\end{tabular*}
\caption{Results using our loss on \cite{yuan2022new} network on NYU Depth.}
\label{tab:newcrfs}
\vspace{-3mm}
\end{center}
\end{table}

\smallskip
\formattedparagraph{\textit{(iv)} Inference Time \& Parameter Comparison.}
We compared our method’s inference time, and the number of model parameters to the recent state-of-the-art NeWCRFs \cite{yuan2022new}. The inference time is measured on the NYU Depth V2 test set \cite{silberman2012indoor} with batch size 1. As shown in \cref{tab:speed}, our method achieves lower scale invariant logarithmic error (SILog) with fewer network parameters and comparable FPS. Such a statistical performance further endorse our approach's practical adequacy.


\begin{table}[th]
\begin{center}
\scriptsize
\begin{tabular*}{0.5\textwidth}{l@{\extracolsep{\fill}}l@{\extracolsep{\fill}}ccc}
\hline
Method & SILog$\downarrow$ & Speed (FPS)$\uparrow$ & Param (M)$\downarrow$ \\
\hline
NeWCRFs & 9.171 & \textbf{10.551} & 258 \\
\hline
\textbf{Ours} & \textbf{8.323} & 9.909 & \textbf{244} \\
\hline
\end{tabular*}
\caption{Comparison of the inference time and parameters with NeWCRFs \cite{yuan2022new} on NYU Depth V2 \cite{silberman2012indoor}.}
\label{tab:speed}
\vspace{-7mm}
\end{center}
\end{table}

\textit{Refer supplementary for more experiments and results}

\section{Conclusion}
This work suitably formalizes the connection between robust statistical modeling techniques, \ie, multivariate covariance modeling with low-rank approximation, and popular loss functions in neural network-based SIDP problem. The novelty presented in this paper arises from the fact that the proposed pipeline and loss term turns out to be more general, hence could be helpful in the broader application of SIDP in several tasks, such as depth uncertainty for robot vision, control and others. Remarkably, the proposed formulation is not only theoretically compelling but observed to be practically beneficial, resulting in a loss function that is used to train the proposed network showing state-of-the-art SIDP results on several benchmark datasets.

\formattedparagraph{Acknowledgments.} This work was partly supported by ETH General Fund (OK), Chinese Scholarship Council (CSC), and The Alexander von Humboldt Foundation.

{\small
\bibliographystyle{ieee_fullname}
\bibliography{egbib}
}


\twocolumn[\section*{\centering \Large Single Image Depth Prediction Made Better: A Multivariate Gaussian Take \\ ---Supplementary Material---\\}]

\begin{abstract}
Our supplementary material accompanies the main paper and is organized as follows. Firstly, it contains detailed mathematical derivations of the proposed loss function and the covariance of the mixture of Gaussian. Secondly, details related to our neural network design and likelihood computation are presented to understand our implementation better. Next, more ablation studies are presented. Finally, visualization of our learned covariance and qualitative SIDP results on several benchmark datasets are presented for completeness.
\end{abstract}
    
\section{Derivations}
We present the detailed derivations for the negative log likelihood, and the covariance matrix for the mixture of Gaussian distributions.
\subsection{Low-Rank Negative Log Likelihood}
We start with the standard multivariate Gaussian distribution with mean $\bm{\mu}_{\theta}(I)\in\mathbb{R}^{N\times 1}$ and covariance $\bm{\Sigma}_{\theta}(I, I)\in\mathbb{R}^{N\times N}$:
\begin{equation}
\label{eq:mvgm_definition_supp}
\Phi(Z|\theta, I)=\mathcal{N}\big(\bm{\mu}_{\theta}(I), \bm{\Sigma}_{\theta}(I, I)\big).
\end{equation}
The log probability density function is:
\begin{equation}
\begin{split}
\log \Phi(Z|\theta, I) = -\frac{N}{2}\log {2\pi} - \frac{1}{2}\log \det (\bm{\Sigma}_{\theta}(I, I)) -\\
\frac{1}{2}(Z-\bm{\mu}_\theta)^T (\bm{\Sigma}_{\theta}(I, I))^{-1}(Z-\bm{\mu}_\theta).
\end{split}
\label{eq:log_prob_density_supp}
\end{equation}
We make the low-rank assumption:
\begin{equation}
\bm{\Sigma}_{\theta}(I, I) = \Psi_{\theta}(I) \Psi_{\theta}(I)^T + \sigma^2 \textbf{eye}(N)
\label{eq:cov_eig_supp}
\end{equation}
where $\Psi_{\theta}(I)\in\mathbb{R}^{N\times M}$ and $M\lll N$, to ease the computing of the determinant and the inversion term.

\formattedparagraph{Determinant:} We follow the matrix determinant lemma\cite{PresTeukVettFlan92} to simplify the computation of the determinant term in Eq.\eqref{eq:log_prob_density_supp}. The determinant of the matrix $\Psi_{\theta}(I) \Psi_{\theta}(I)^T + \sigma^2 \textbf{eye}(N)\in\mathbb{R}^{N\times N}$ is computed by
\begin{equation}
\begin{split}
&\det (\Psi_{\theta}(I) \Psi_{\theta}(I)^T + \sigma^2 \textbf{eye}(N))\\
= & \det(\sigma^2 \textbf{eye}(N))\det(\textbf{eye}(M)+\Psi_{\theta}^T(\sigma^{-2}\textbf{eye}(N))\Psi_{\theta})\\
= & \sigma^{2N} \det(\textbf{eye}(M)+\sigma^{-2}\Psi_{\theta}^T(I)\Psi_{\theta}(I))
\end{split}
\label{eq:wood_determinant_supp}
\end{equation}
The complexity of time and space for computing the determinant of the matrix  $\textbf{eye}(M)+\sigma^{-2}\Psi_{\theta}^T(I)\Psi_{\theta}(I)\in\mathbb{R}^{M\times M}$ is $O(M^3)$ \cite{rasmussen2003gaussian}.

\formattedparagraph{Inversion:} Then we use the matrix inversion lemma \cite{PresTeukVettFlan92} to ease the computation of the inversion of the matrix $\Psi_{\theta}(I) \Psi_{\theta}(I)^T + \sigma^2 \textbf{eye}(N)\in\mathbb{R}^{N\times N}$:
\begin{equation}
\begin{split}
&(\Psi_{\theta}(I) \Psi_{\theta}(I)^T + \sigma^2 \textbf{eye}(N))^{-1}\\
= &
-(\sigma^2\textbf{eye}(N))^{-1}\Psi_{\theta}(\textbf{eye}(M)+\sigma^{-2}\Psi_{\theta}^T\Psi_{\theta})^{-1}
\\& \Psi_{\theta}^T(\sigma^2\textbf{eye}(N))^{-1} + (\sigma^2\textbf{eye}(N))^{-1}\\
= & \sigma^{-2}\textbf{eye}(N) - \sigma^{-4}\Psi_{\theta}(\textbf{eye}(M)+\sigma^{-2}\Psi_{\theta}^T\Psi_{\theta})^{-1}\Psi_{\theta}^T
\end{split}
\label{eq:wood_inversion_supp}
\end{equation}
Again, computing the inversion of the term $\textbf{eye}(M)+\sigma^{-2}\Psi_{\theta}^T\Psi_{\theta}\in\mathbb{R}^{M\times M}$ requires time and space complexity $O(M^3)$.

\formattedparagraph{Total:} We put the Eq.\eqref{eq:wood_determinant_supp} and Eq.\eqref{eq:wood_inversion_supp} into Eq.\eqref{eq:log_prob_density_supp}, then we can easily obtain:
\begin{equation}
\begin{split}
	\log \Phi(Z|&\theta, I)= -\frac{N}{2}\log 2\pi\sigma^2 -  \frac{1}{2}\log \det (\mathbf{A})  - \\ 
	& \frac{\sigma^{-2}}{2} \mathbf{r}^T\mathbf{r} + \frac{\sigma^{-4}}{2}\mathbf{r}^T\Psi_{\theta}(I)\mathbf{A}^{-1}\Psi_{\theta}(I)^T\mathbf{r}
\end{split}
\label{eq:wood_gaussian_supp}
\end{equation}
where, $\mathbf{r} = Z - \bm{\mu}_{\theta}(I)$, and $ \mathbf{A} = \sigma^{-2}\Psi_{\theta}(I)^T\Psi_{\theta}(I) + \textbf{eye}(M)$.
\begin{figure*}[h]
\centering
\includegraphics{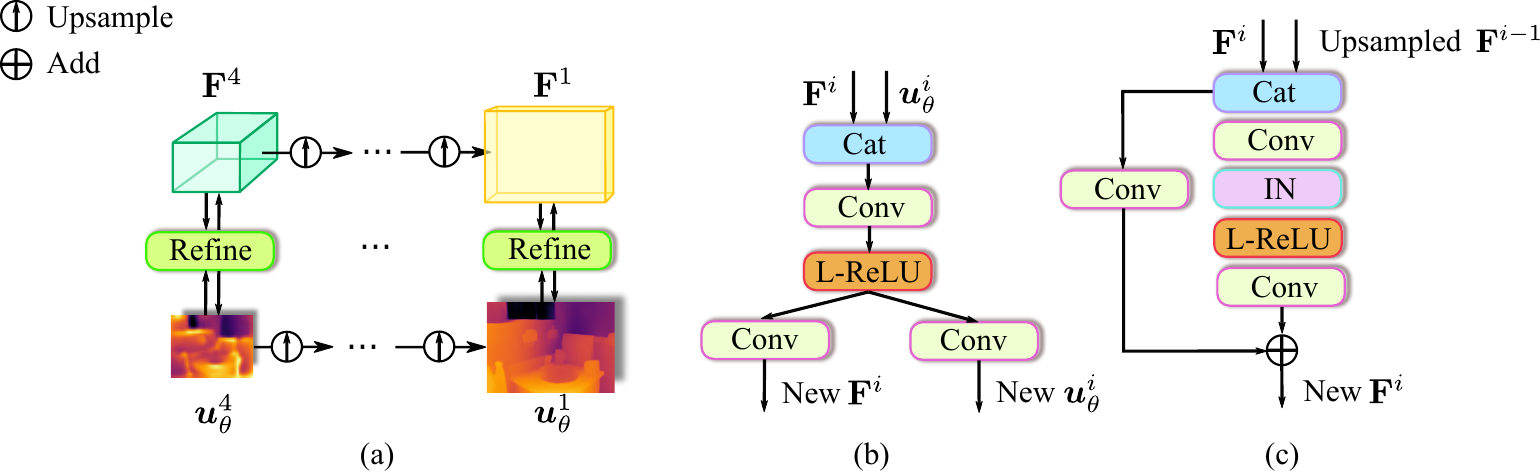}
\caption{(a) The architecture of the U-decoder. (b) The detailed structure of the refine module. (c) The detailed structure of the fusion module. We use L-ReLU to represent the LeakyReLU operation, and IN to represent the instance normalization layer.}
\label{fig:arch_supp}
\end{figure*}
\subsection{Covariance of Mixture of Gaussian}
We present the covariance matrix of the mixture of Gaussian distributions:
\begin{equation}
\Phi(Z|I, \mathcal{D})=\frac{1}{S}\sum_s \Phi(Z|\theta^s, I)
\label{eq:uncertainty_supp}
\end{equation}
where $\Phi(Z|\theta^s, I)$ is the probability density function for a single Gaussian distribution.
By the law of total variance \cite{weiss2006course}, we obtain:
\begin{equation}
\begin{split}
\mathrm{Var}(Z) = & \mathrm{E}[\mathrm{Var}(Z|s)] + \mathrm{Var}(\mathrm{E}[Z|s]) \\
=& \frac{1}{S} \sum_{s}(\Psi^{(s)}\Psi^{(s)T}+\sigma^2\textbf{eye}(M))\\
& + \frac{1}{S}\sum_s (\bm{u}^{(s)}-\bar{\bm{\mu}})(\bm{u}^{(s)}-\bar{\bm{\mu}})^T\\
=& \sigma^{2}\textbf{eye}(N) + \sum_{s}\frac{1}{\sqrt{S}}\Psi^{(s)}\frac{1}{\sqrt{S}}\Psi^{(s)T}\\
	& + \sum_s \frac{1}{\sqrt{S}}(\bm{u}^{(s)}-\bar{\bm{\mu}})\frac{1}{\sqrt{S}}(\bm{u}^{(s)}-\bar{\bm{\mu}})^T
\end{split}
\end{equation}
By constructing the following matrix:
\begin{equation}
\bar{\Psi} = \frac{1}{\sqrt S}\mathbf{concat}(\Psi^{(1)},\ldots,\Psi^{(S)}, \bm{\mu}^{(1)}-\bar{\bm{\mu}},\ldots,\bm{\mu}^{(S)}-\bar{\bm{\mu}}),    
\end{equation}
the covariance matrix can be written as $\bar{\Psi}\bar{\Psi}^T + \sigma^2\textbf{eye}(N)$, which shares the same form as Eq.\eqref{eq:cov_eig_supp}.

\section{Implementation Details}
In this section, we present the details for the network architecture design and the likelihood computation.
\subsection{Network Architecture}
We introduce the details about the network architecture. The network is comprised of (a) encoder, (b) U-decoder, and (c) K-decoder. In general, we set the kernel size of the convolution layers to be 3 unless otherwise stated.

\formattedparagraph{(a) Encoder.} We adopt the standard Swin-Large \cite{liu2021swin} as our encoder. More specifically, the patch size is 4, the window size is 12, and the embedding dim is 192. The numbers of feature channels in four stages are 192,  384, 768, 1536, respectively. And there are 2, 2, 18, 2 blocks in the four stages, respectively. We collect the output feature map from the last block in each stage into $\mathbf{F}=\{\mathbf{F}^1, \mathbf{F}^2, \mathbf{F}^3, \mathbf{F}^4\}$, where $\mathbf{F}^1$ has 192 channels and stride 4, $\mathbf{F}^2$ has 384 channels and stride 8, $\mathbf{F}^3$ has 768 channels and stride 16, $\mathbf{F}^4$ has 1536 channels and stride 32.

\formattedparagraph{(b) U-Decoder.} The input to the U-decoder is $\mathbf{F}=\{\mathbf{F}^i\}_{i=1}^{4}$. From the input, the U-decoder will predict a set of depth maps $\{\bm{\mu}^i_{\theta}\}_{i=1}^4$. The network architecture of U-decoder is shown in Fig.\ref{fig:arch_supp} (a). We start with $\mathbf{F}^4$, which has 1536 channels and stride 32. We first predict the $\bm{\mu}_{\theta}^4$ though a convolution layer, which has 1536 input channels and 128 output channels. We utilize a refine module to refine the $\mathbf{F}^4$ and $\bm{\mu}_{\theta}^4$. The refine module is shown in Fig. \ref{fig:arch_supp} (b). Then we upsample the $\mathbf{F}^4$ via bi-linear interpolation. The upsampled $\mathbf{F}^4$ will be concatenated with the $\mathbf{F}^3$ from the encoder. Then we adopt a fusion module to fuse the information from the $\mathbf{F}^3$ and the upsampled $\mathbf{F}^4$. The fusion module is shown in Fig. \ref{fig:arch_supp} (c). The fused $\mathbf{F}^3$ has 512 channels and stride 16. We upsample the $\bm{\mu}_{\theta}^4$ to $\bm{\mu}_{\theta}^3$ via bi-linear interpolation. Similar to the above procedures, the $\mathbf{F}^3$ will be refined with $\bm{\mu}_{\theta}^3$, and then upsampled and fused with $\mathbf{F}^2$ from the encoder. The fused $\mathbf{F}^2$ has 256 channels and stride 8. The $\bm{\mu}_{\theta}^3$ is also upsampled to $\bm{\mu}_{\theta}^2$ via bi-linear interpolation. With the same operations, we can further obtain the $\mathbf{F}^1$, which has 64 channels and stride 4. And we can also obtain $\bm{\mu}_{\theta}^1$. Now $\bm{\mu}_{\theta}^1$, $\bm{\mu}_{\theta}^2$, $\bm{\mu}_{\theta}^3$, $\bm{\mu}_{\theta}^4$ all have 128 channels. We upsample them to stride 1 via bi-linear operation, and compress the number of channels to 1 via a convolution layer.

\formattedparagraph{(c) K-Decoder.} The K-decoder aims to predict the $\Psi_{\theta}$. The input to the K-decoder is $\mathbf{F}=\{\mathbf{F}^i\}_{i=1}^{4}$. The architecture of K-decoder is similar to U-decoder, except for there is no depth map predictions and refine modules. More specifically, we first upsample $\mathbf{F}^4$ via bi-linear interpolation, then fuse with the $\mathbf{F}^3$ though the fusion module. The fusion module is the same as the one in the U-decoder. The fused $\mathbf{F}^3$ has 512 channels and stride 16. Similar to the above procedures, we can further obtain the fused $\mathbf{F}^2$ and the fused $\mathbf{F}^1$. The fused $\mathbf{F}^2$ has 256 channels, and the fused  $\mathbf{F}^1$ has 128 channels. We predict $\Psi_{\theta}$ from $\mathbf{F}^1$ by a convolution layer that has 128 input channels and 128 output channels.

\subsection{Likelihood Computation}
We provide the pseudo code to compute the log likelihood in \cref{alg:cap}.
\begin{algorithm}
\caption{Log Likelihood Computation}\label{alg:cap}
\hspace*{\algorithmicindent} \textbf{Input:} $\bm{\mu}_{\theta}(I)\in \mathbb{R}^{N \times 1}$, $\Psi_{\theta}(I)\in \mathbb{R}^{N \times M}$, $\sigma\in \mathbb{R}^+$, \hspace*{\algorithmicindent} and $Z^{gt}\in \mathbb{R}^{N \times 1}$\\
\hspace*{\algorithmicindent} \textbf{Output:} $\log \Phi(Z^{gt}|\theta, I)$
\begin{algorithmic}[1]
\State $\mathbf{r} = Z^{gt} - \bm{\mu}_{\theta}(I)$
\State $\mathbf{p} = \Psi_{\theta}(I)^T\mathbf{r}$
\State $ \mathbf{A} = \sigma^{-2}\Psi_{\theta}(I)^T\Psi_{\theta}(I) + \textbf{eye}(M)$
\State $\mathbf{L}\mathbf{L}^T=\text{cholesky}(\mathbf{A})$
\State $\mathbf{q}=\mathbf{L}\backslash\mathbf{p}$ \Comment{Or: $\mathbf{q}=\text{inv}(\mathbf{L})*\mathbf{p}$}
\State \textbf{Return} $-\frac{N}{2}\log 2\pi\sigma^2 -  \sum_{i}\log\mathbf{L}_{ii}  - \frac{\sigma^{-2}}{2} \mathbf{r}^T\mathbf{r} + \frac{\sigma^{-4}}{2} \mathbf{q}^T\mathbf{q}$
\end{algorithmic}
\end{algorithm}

\begin{figure}[t]
\begin{center}
\begin{subfigure}[b]{0.15\textwidth}
\begin{center}
\begin{minipage}[b]{1.0\textwidth}
\includegraphics[width=1.0\textwidth]{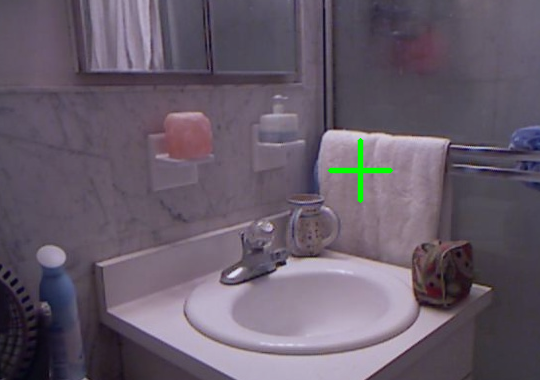}
\includegraphics[width=1.0\textwidth]{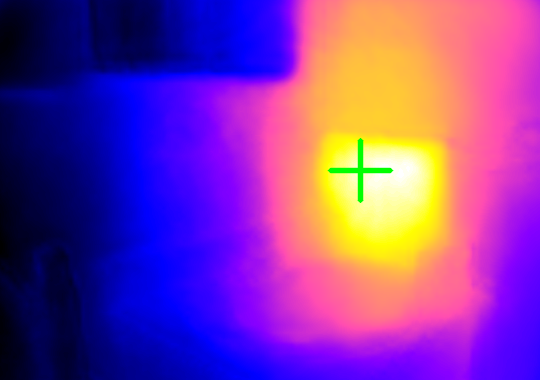}
\end{minipage}
\caption{Image 1}
\end{center}
\end{subfigure}
\begin{subfigure}[b]{0.15\textwidth}
\begin{center}
\begin{minipage}[b]{1.0\textwidth}
\includegraphics[width=1.0\textwidth]{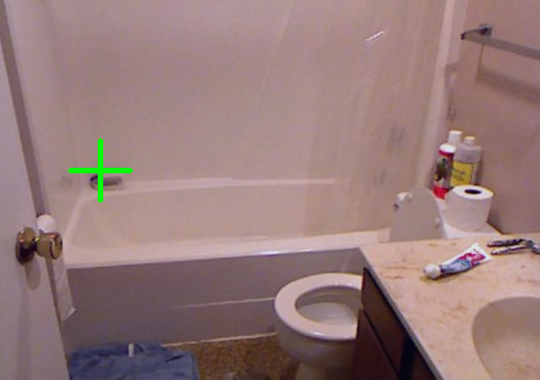}
\includegraphics[width=1.0\textwidth]{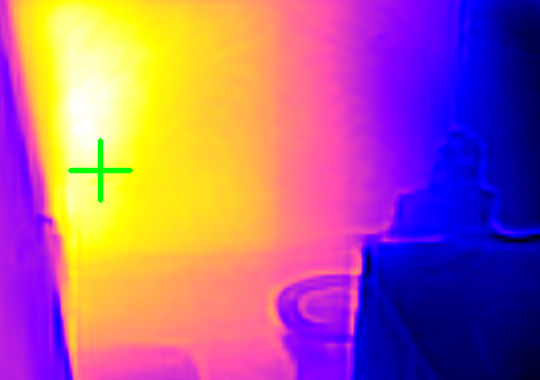}
\end{minipage}
\caption{Image 2}
\end{center}
\end{subfigure} 
\begin{subfigure}[b]{0.15\textwidth}
\begin{center}
\begin{minipage}[b]{1.0\textwidth}
\includegraphics[width=1.0\textwidth]{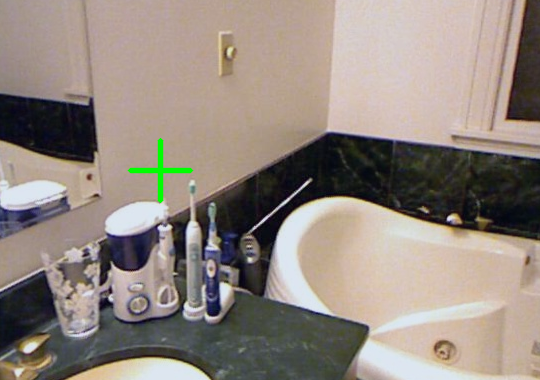}
\includegraphics[width=1.0\textwidth]{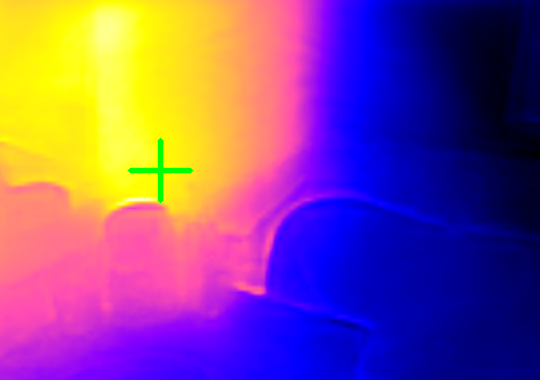}
\end{minipage}
\caption{Image 3}
\end{center}
\end{subfigure}  
\end{center}
\caption{\textbf{Visualization of Covariance}. Top: test image. Bottom: covariance with respect to the pixel which is marked as a green cross. The yellow and light regions have higher covariance than the blue and dark ones.}
\vspace{-4mm}
\label{fig:vis_cov}
\end{figure}

\section{More Ablations}
In this section, we provide more ablation studies.
\subsection{Comparison with Deep Evidential Regression}
We compare with the Deep Evidential Regression \cite{amini2020deep} on NYU Depth V2 test set \cite{silberman2012indoor} and KITTI Eigen split \cite{eigen2014depth}. We present the experimental results in Tab. \ref{tab:DER_comparison}. Our approach achieves better depth prediction accuracy and uncertainty estimation.
\begin{table}[h]
\centering
\scriptsize
\begin{tabular}{c|c|c|c|c|c}
\hline
	Dataset & Loss & SILog $\downarrow$ & NLL $\downarrow$ & RMS $\downarrow$ & $\delta_1$ $\uparrow$\\
	\hline
	\multirow{2}{*}{NYU}& DER & 9.253  & 0.118 & 0.330 & 0.927\\
	& Ours & \textbf{8.323} & \textbf{-1.342} & \textbf{0.311} & \textbf{0.933}\\
	\hline
	\multirow{2}{*}{KITTI}& DER & 7.500 & 1.072 & 0.225 &  0.971\\
	& Ours & \textbf{6.757} & \textbf{-0.222} & \textbf{0.202} & \textbf{0.976}\\
	\hline
\end{tabular}
\caption{Comparison with Deep Evidential Regression (DER).}
\label{tab:DER_comparison}
\end{table} 
\subsection{FPS with K-Decoder}
In general K-Decoder is used only at train time. The K-Decoder can be abandoned at test time for SIDP if uncertainty information is not required. For completeness, we present the FPS information at test time in Tab. \ref{tab:test_with_fps}.
\begin{table}[h]
\centering
\scriptsize
\begin{tabular}{c|c|c}
\hline
K-Decoder & SI Log $\downarrow$  & FPS\\
\hline
w/o & \textbf{8.323} & \textbf{9.909}  \\
\hline
w/ & \textbf{8.323} & 8.445\\
\hline 
\end{tabular}
\caption{SI Log error and corresponding FPS on NYU Dataset.}
\label{tab:test_with_fps}
\end{table}
\section{Visualization of Learned Covariance}
To understand the covariance learned by the proposed negative log likelihood loss function, we visualize the covariance for selected pixels. More specifically, for each image we select a pixel (marked as a green cross), and visualize the covariance between the pixel and all other pixels. The results are shown in Fig. \ref{fig:vis_cov}. We observe that the pixels from nearby regions or the same objects usually have higher covariance.

\section{Qualitative Results}
\begin{figure*}[t]
\begin{center}
\begin{subfigure}[b]{0.18\textwidth}
\begin{center}
\begin{minipage}[b]{1.0\textwidth}
\includegraphics[width=1.0\textwidth]{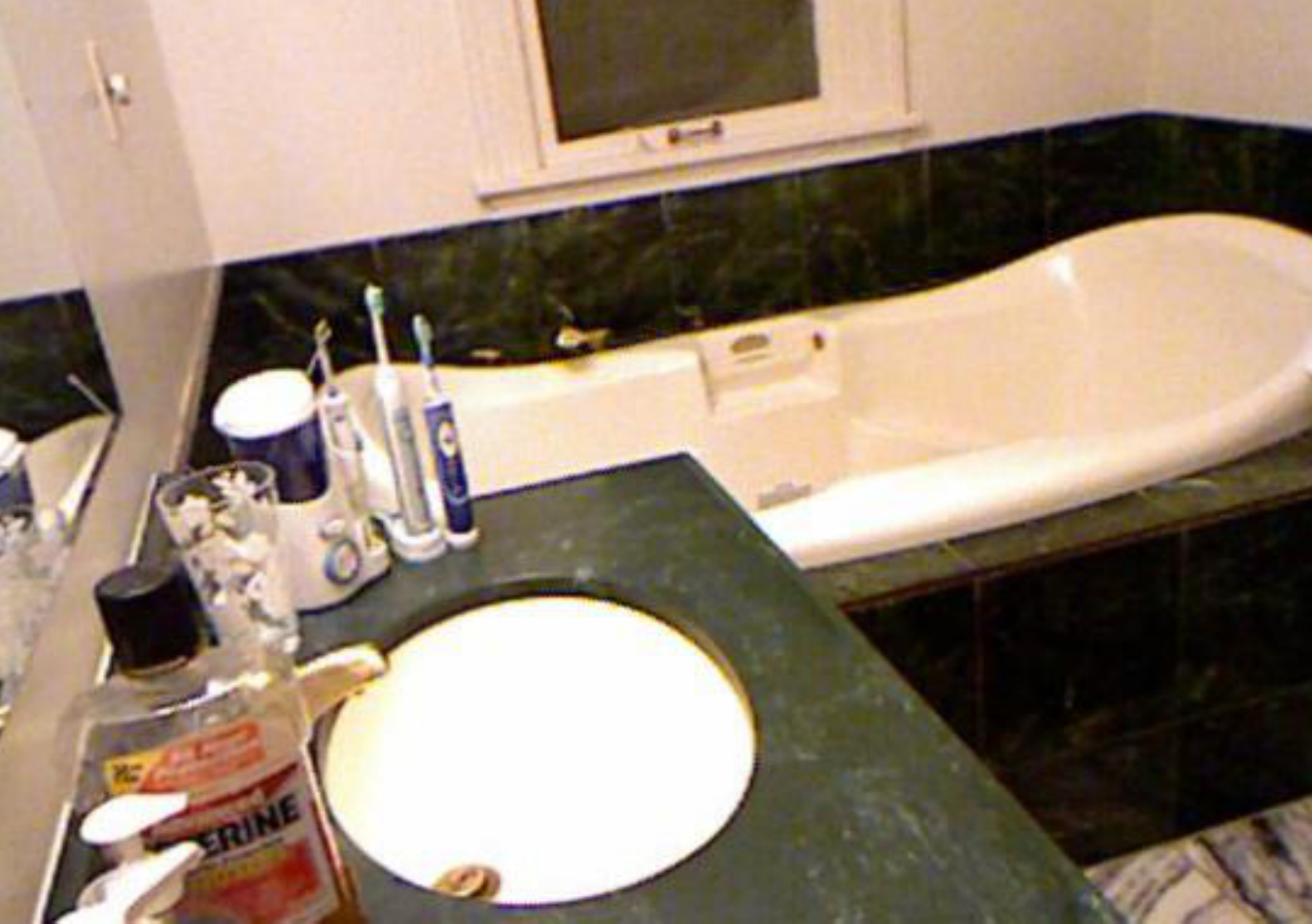}
\includegraphics[width=1.0\textwidth]{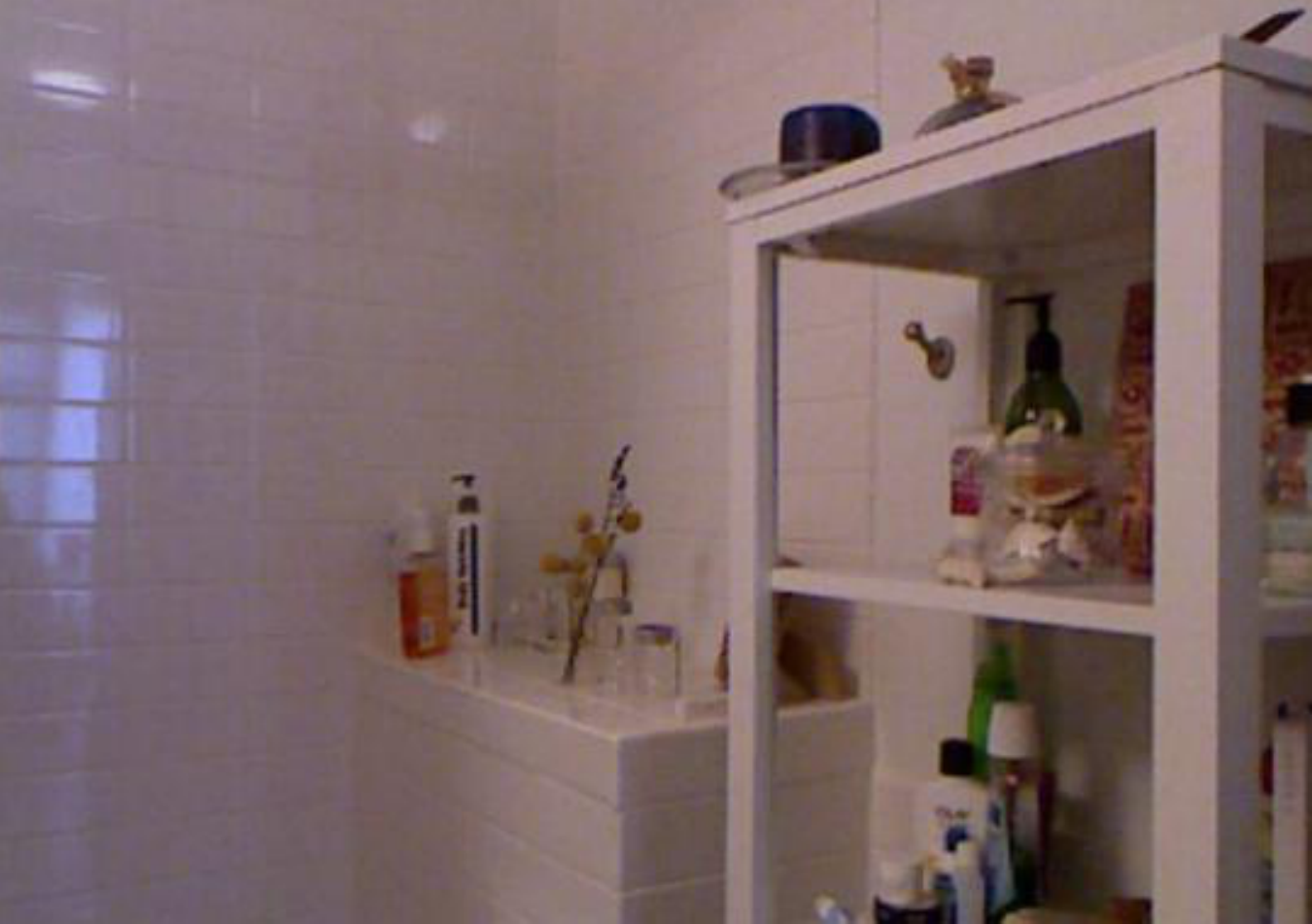}
\includegraphics[width=1.0\textwidth]{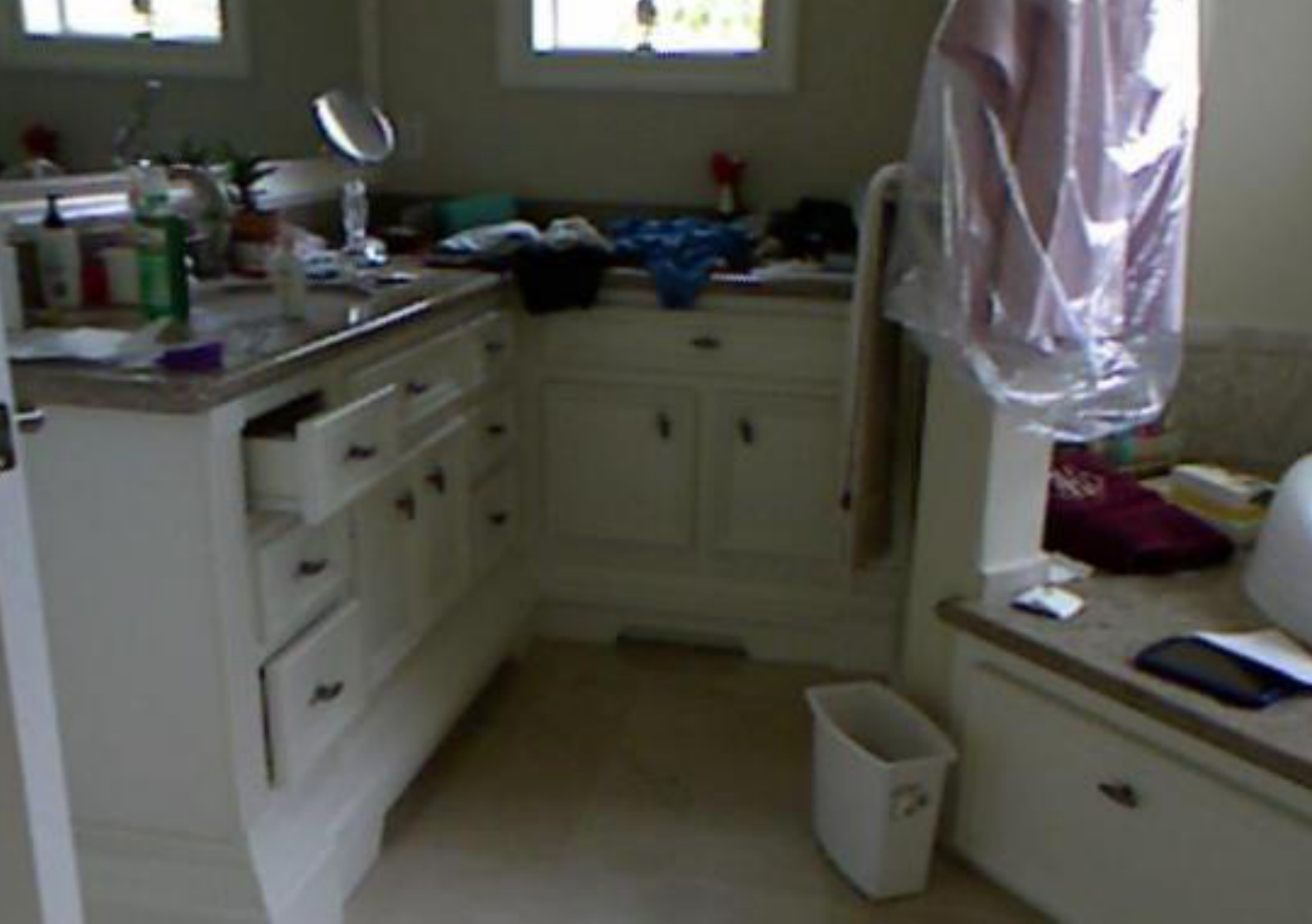}
\includegraphics[width=1.0\textwidth]{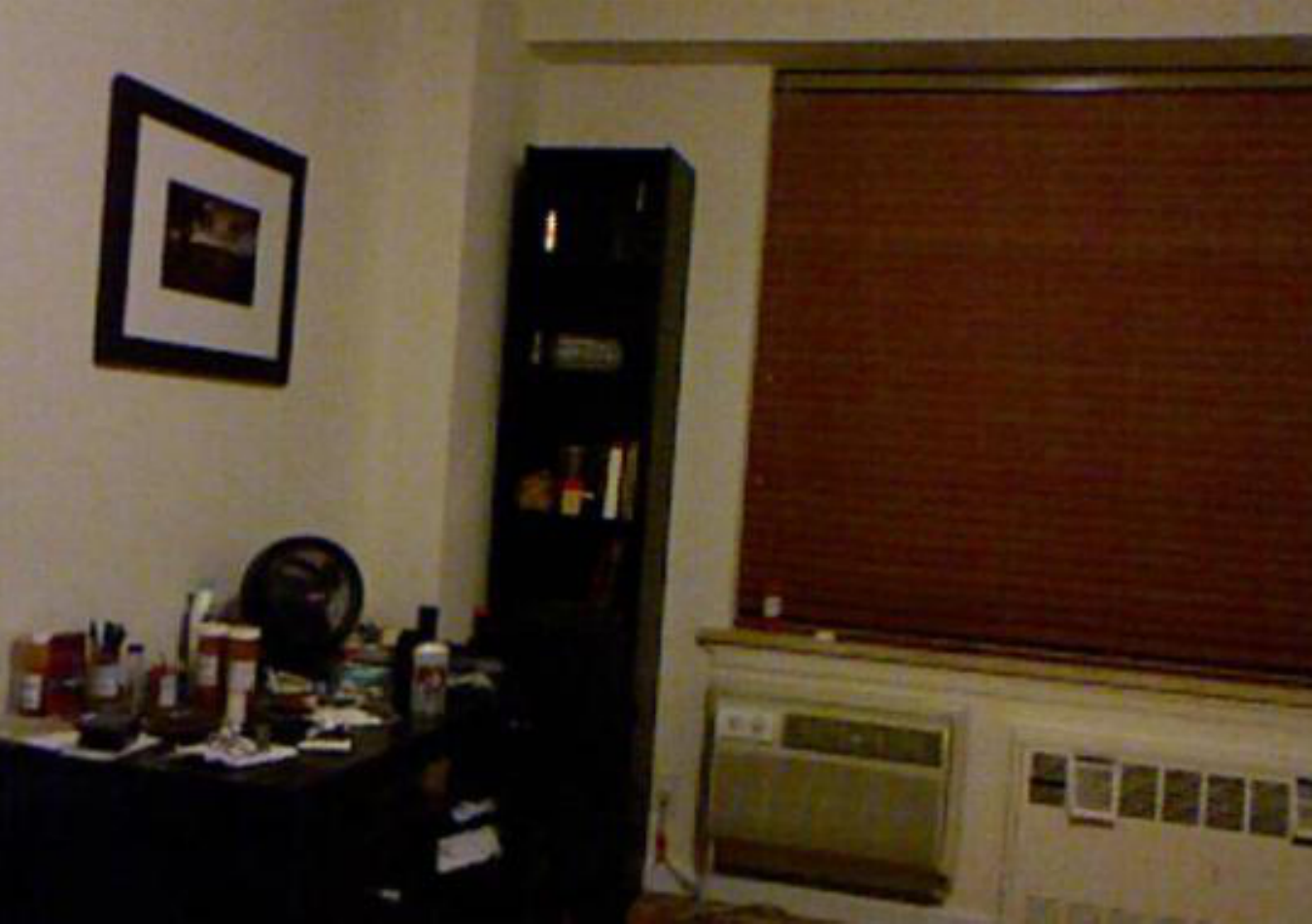}
\includegraphics[width=1.0\textwidth]{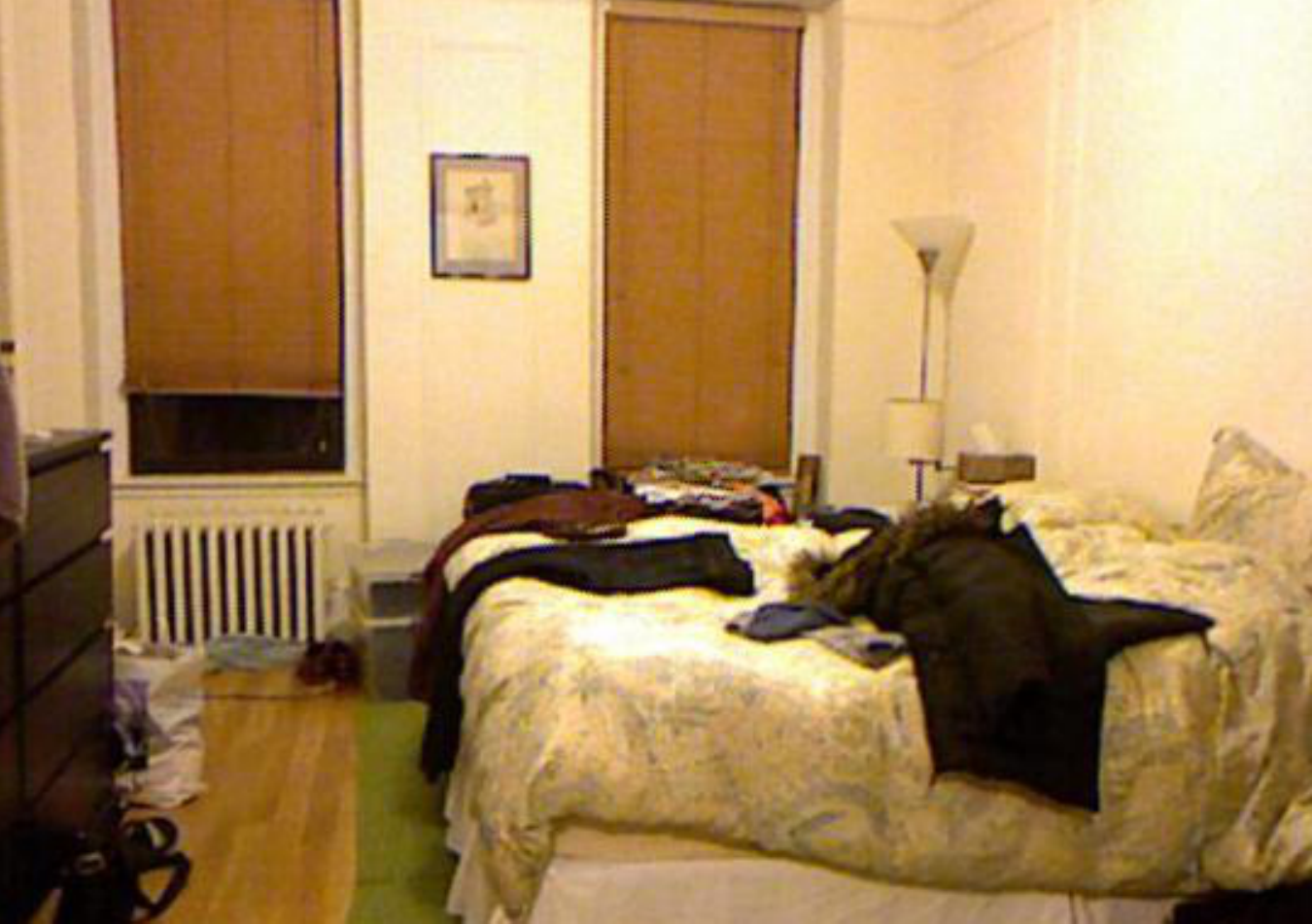}
\includegraphics[width=1.0\textwidth]{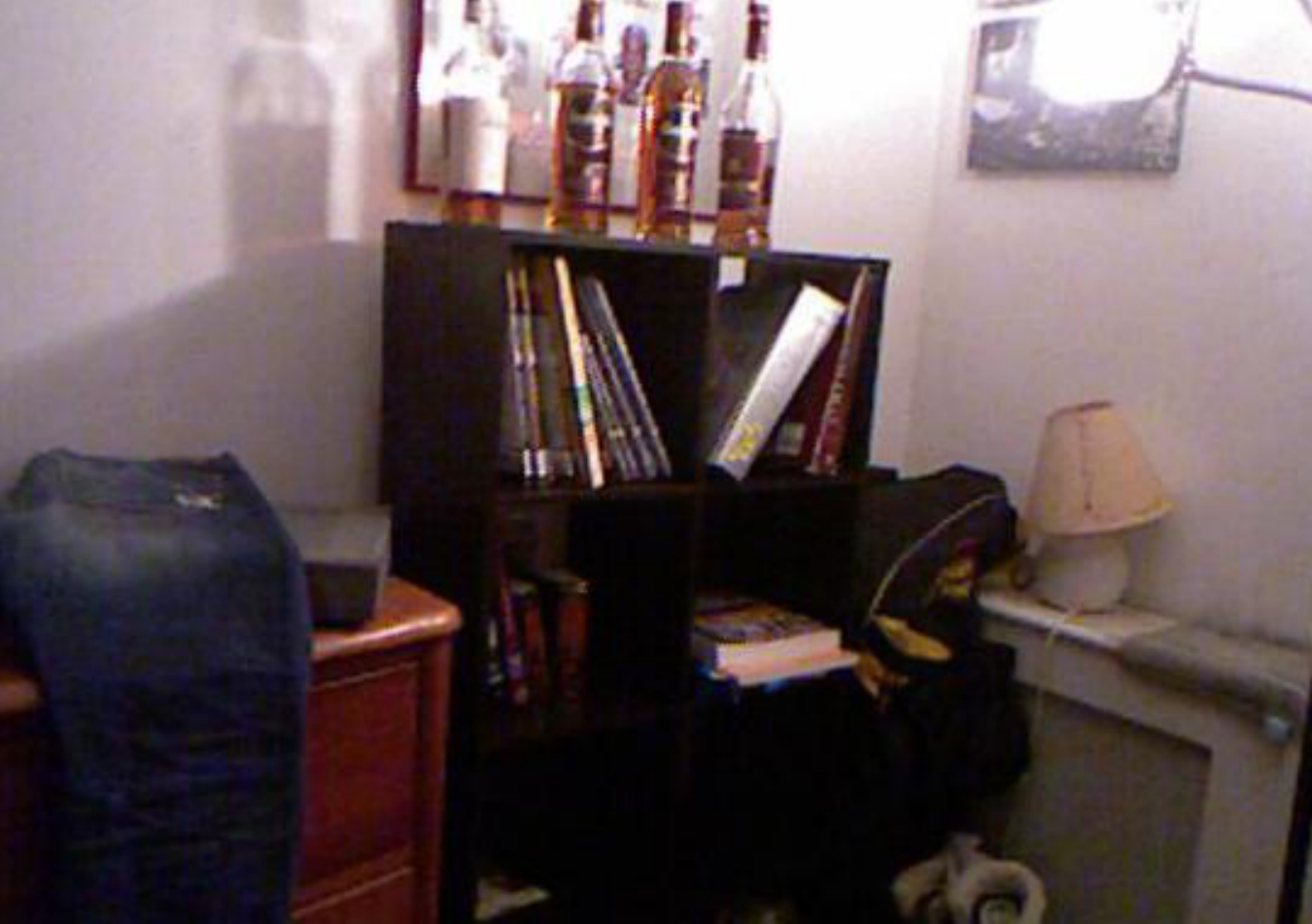}
\includegraphics[width=1.0\textwidth]{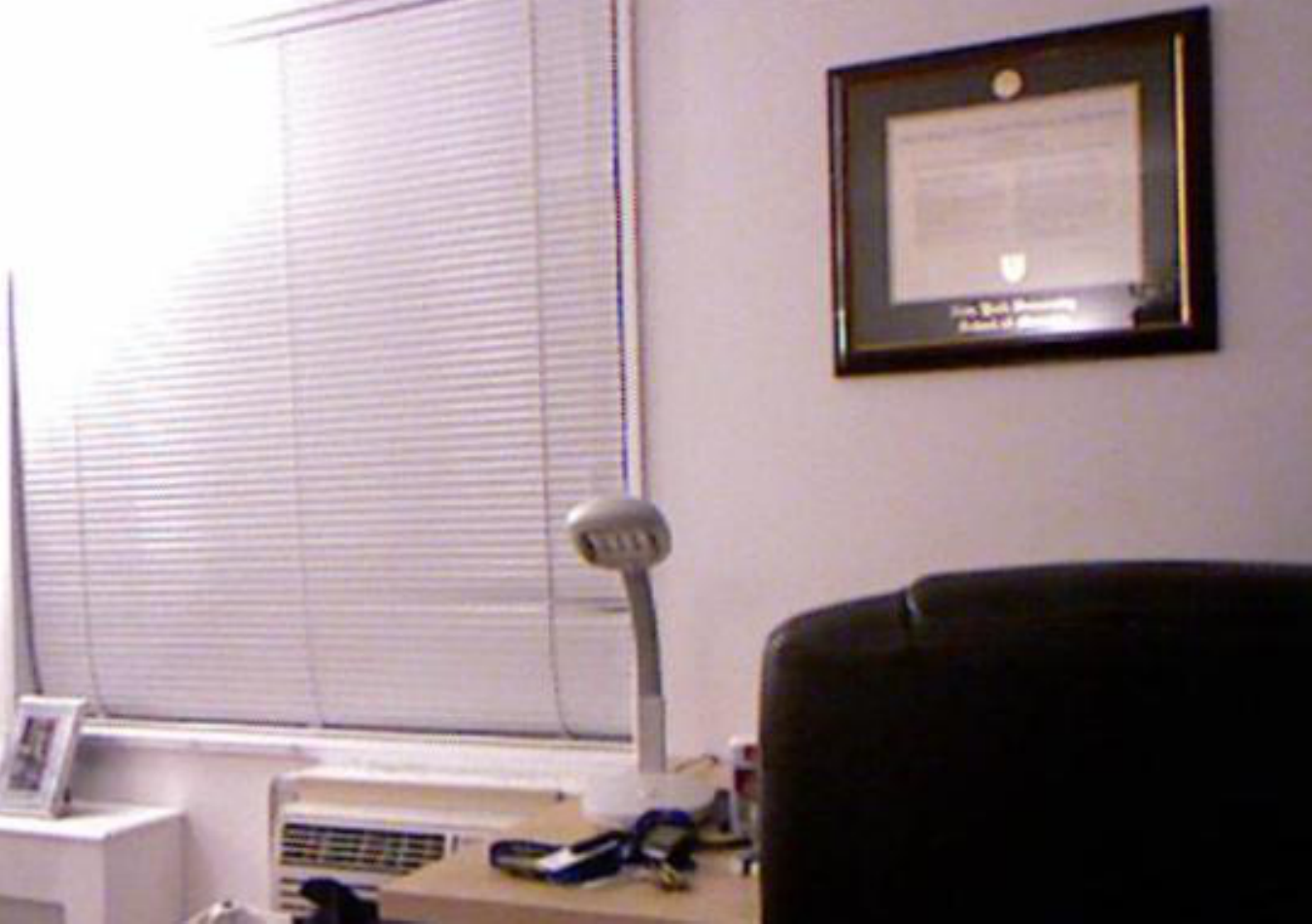}
\includegraphics[width=1.0\textwidth]{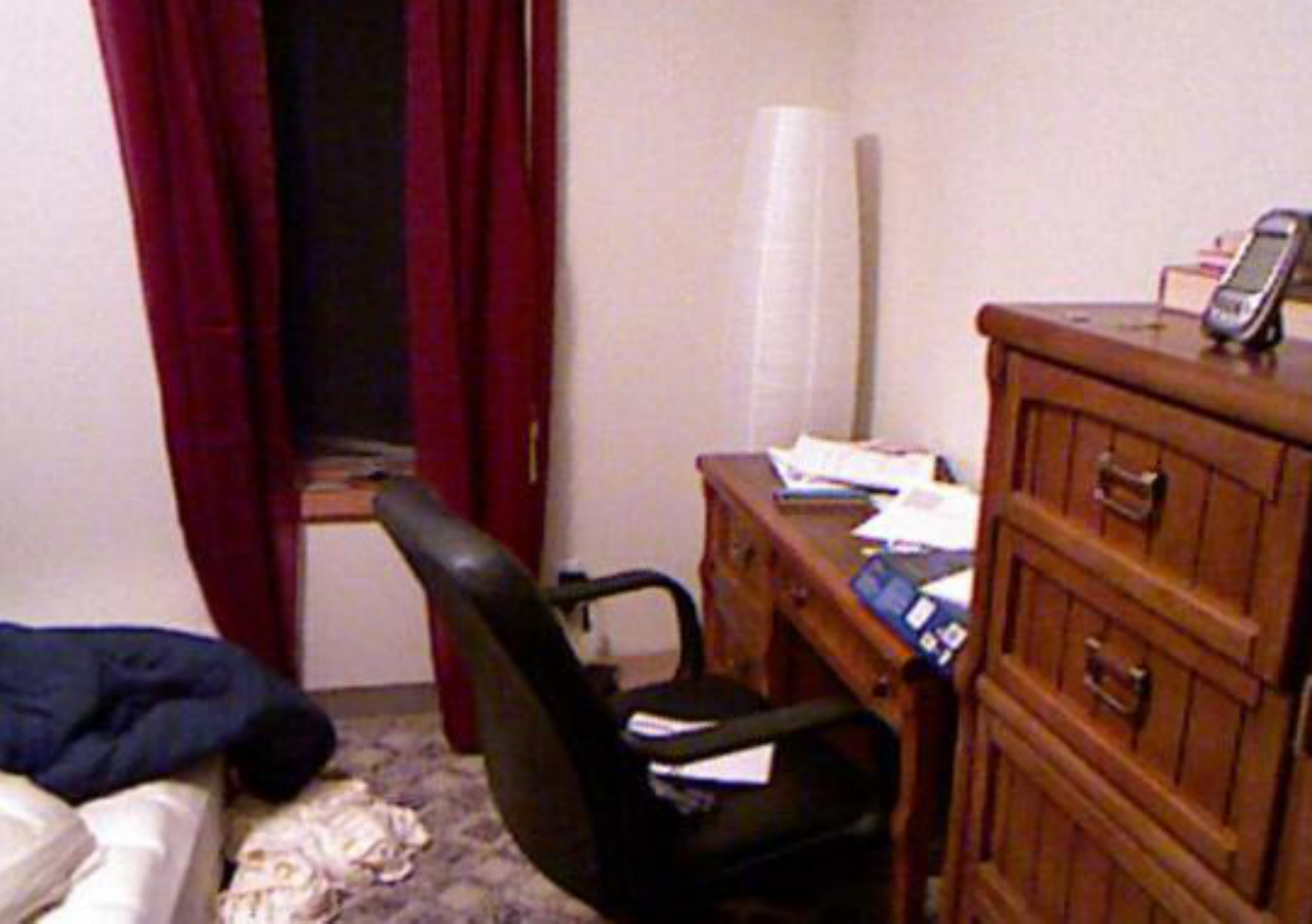}
\includegraphics[width=1.0\textwidth]{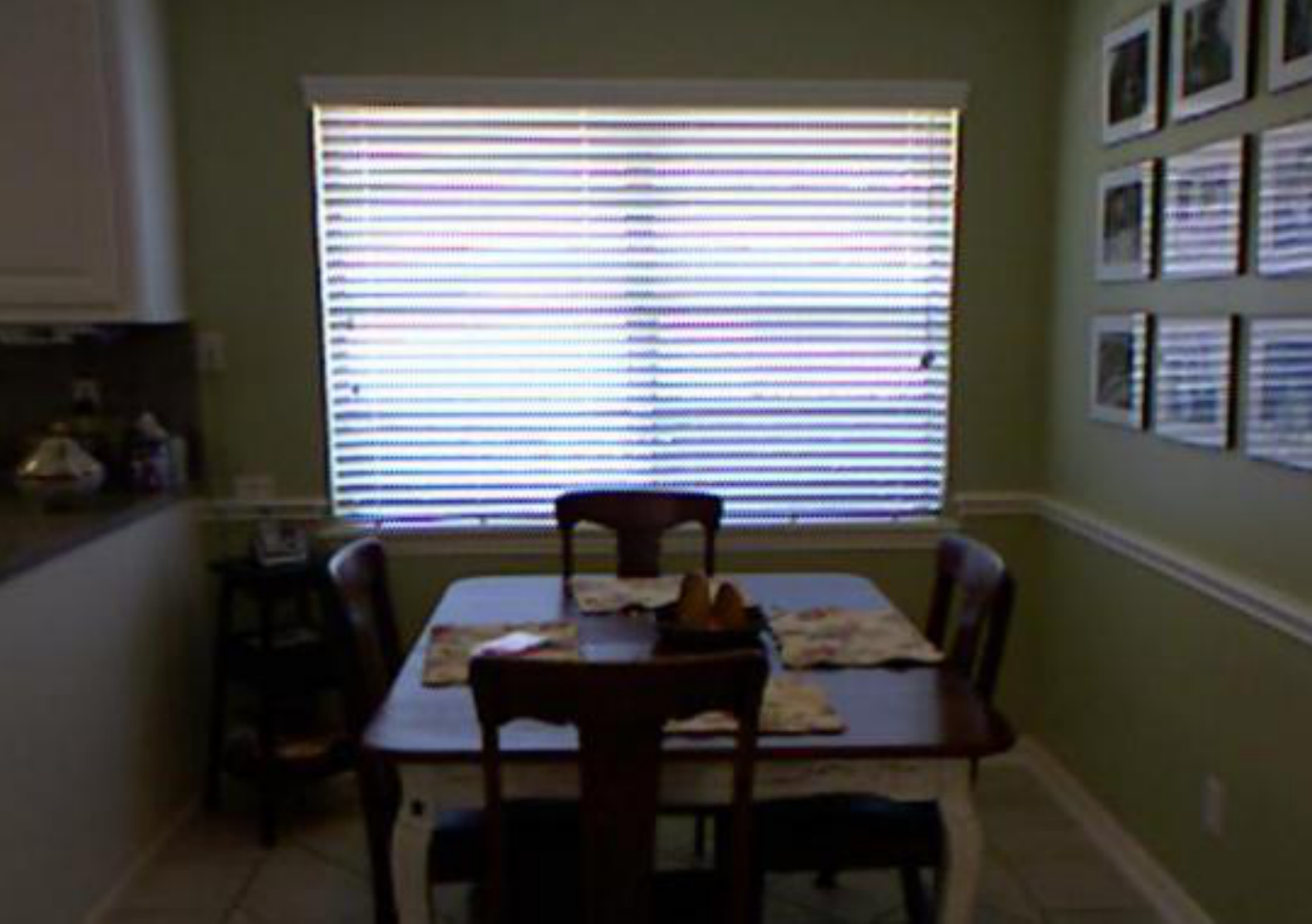}
\end{minipage}
\caption{Test}
\end{center}
\end{subfigure}
\begin{subfigure}[b]{0.18\textwidth}
\begin{center}
\begin{minipage}[b]{1.0\textwidth}
\includegraphics[width=1.0\textwidth]{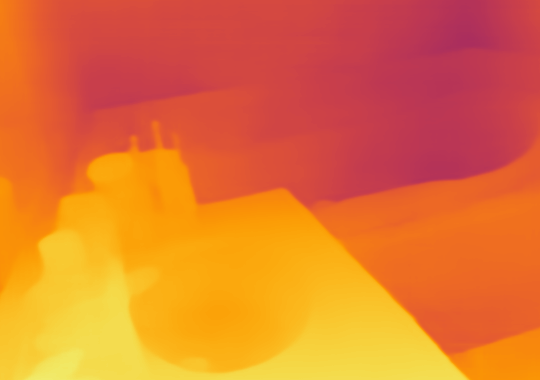}
\includegraphics[width=1.0\textwidth]{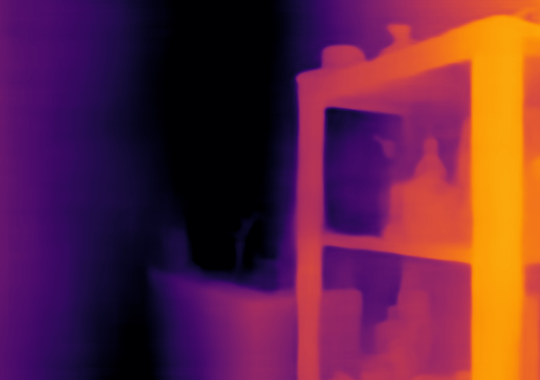}
\includegraphics[width=1.0\textwidth]{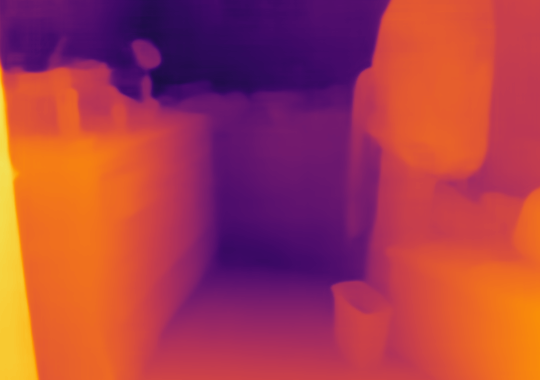}
\includegraphics[width=1.0\textwidth]{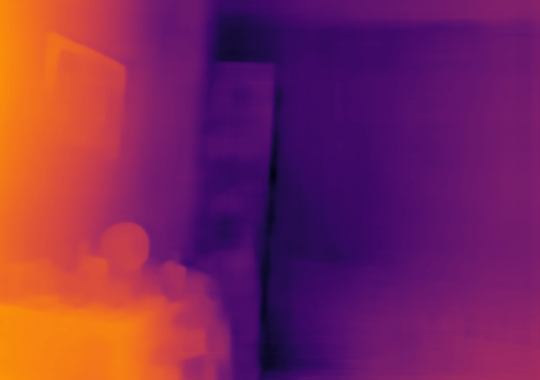}
\includegraphics[width=1.0\textwidth]{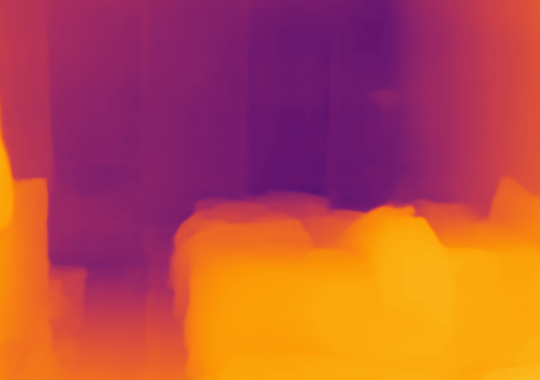}
\includegraphics[width=1.0\textwidth]{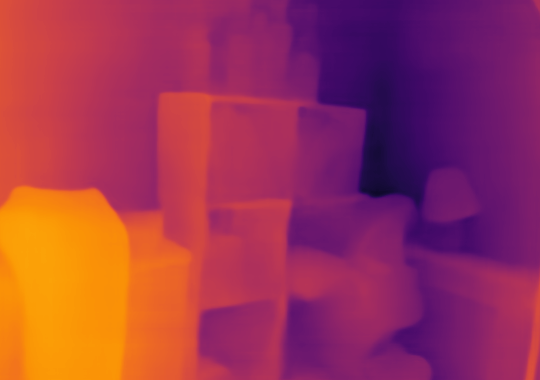}
\includegraphics[width=1.0\textwidth]{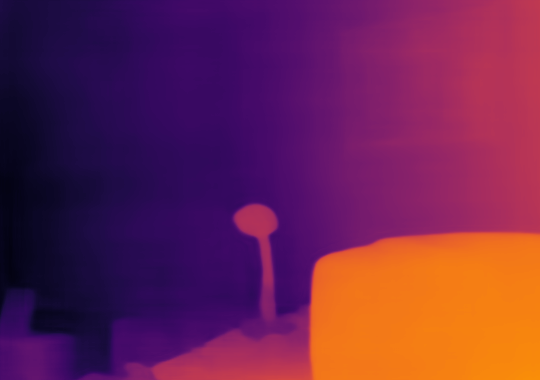}
\includegraphics[width=1.0\textwidth]{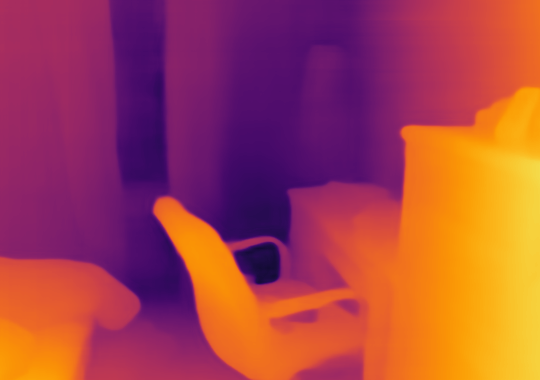}
\includegraphics[width=1.0\textwidth]{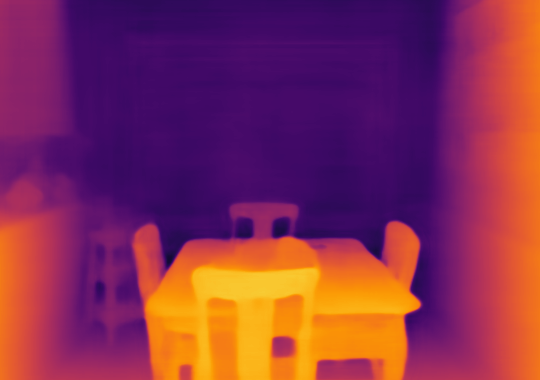}
\end{minipage}
\caption{DPT\cite{ranftl2021vision}}
\end{center}
\end{subfigure} 
\begin{subfigure}[b]{0.18\textwidth}
\begin{center}
\begin{minipage}[b]{1.0\textwidth}
\includegraphics[width=1.0\textwidth]{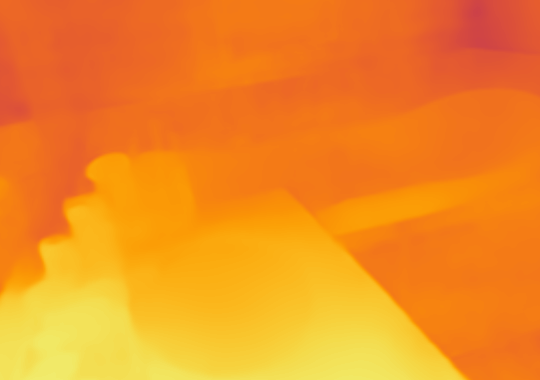}
\includegraphics[width=1.0\textwidth]{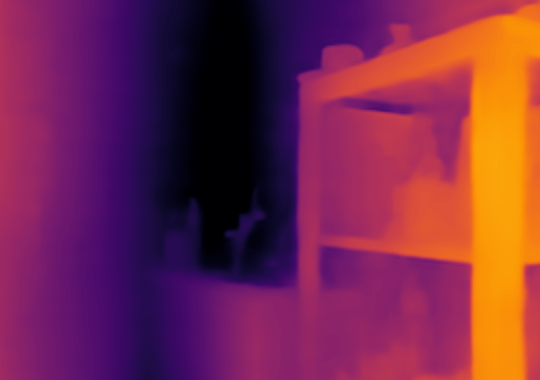}
\includegraphics[width=1.0\textwidth]{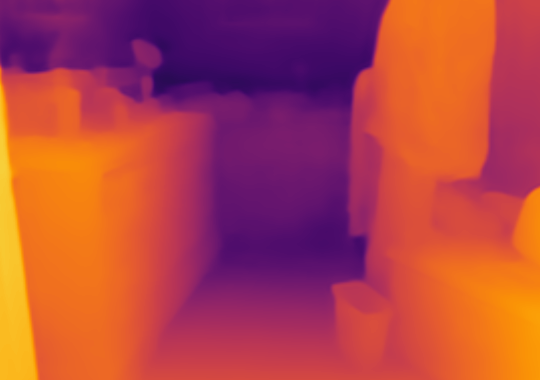}
\includegraphics[width=1.0\textwidth]{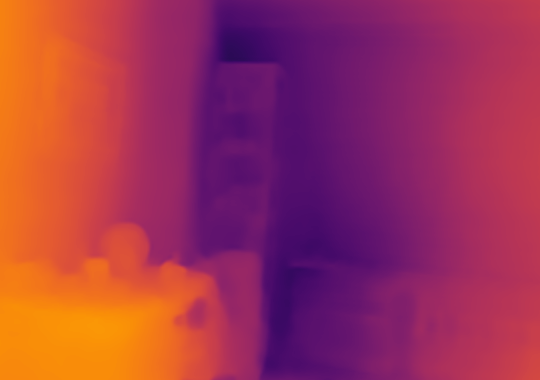}
\includegraphics[width=1.0\textwidth]{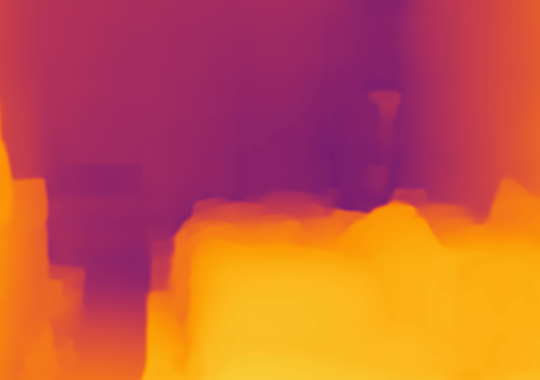}
\includegraphics[width=1.0\textwidth]{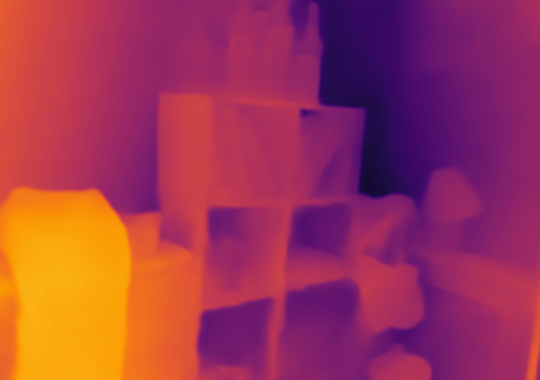}
\includegraphics[width=1.0\textwidth]{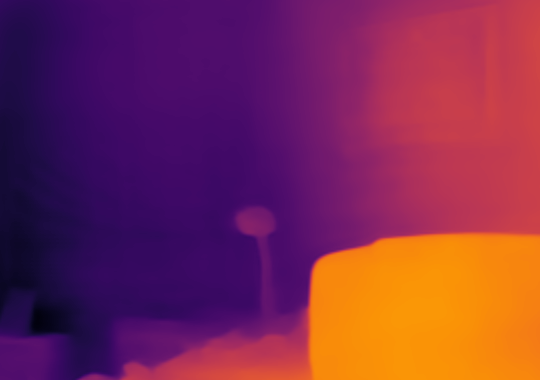}
\includegraphics[width=1.0\textwidth]{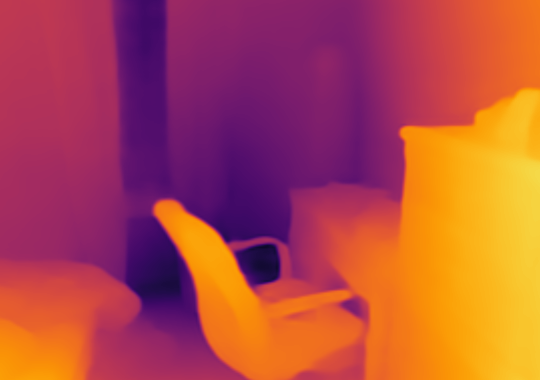}
\includegraphics[width=1.0\textwidth]{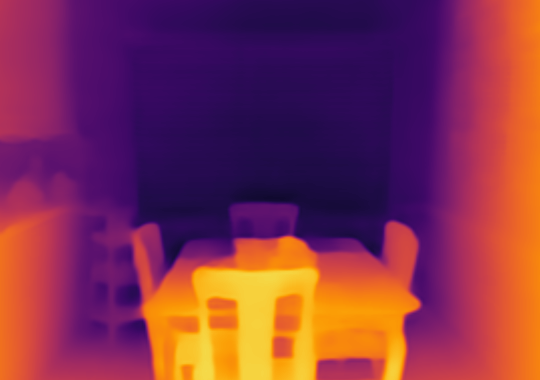}
\end{minipage}
\caption{AdaBins\cite{bhat2021adabins}}
\end{center}
\end{subfigure}  
\begin{subfigure}[b]{0.18\textwidth}
\begin{center}
\begin{minipage}[b]{1.0\textwidth}
\includegraphics[width=1.0\textwidth]{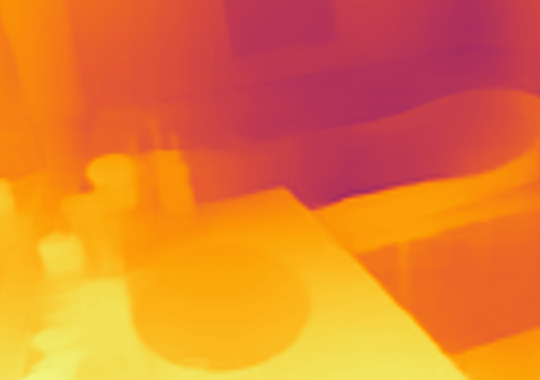}
\includegraphics[width=1.0\textwidth]{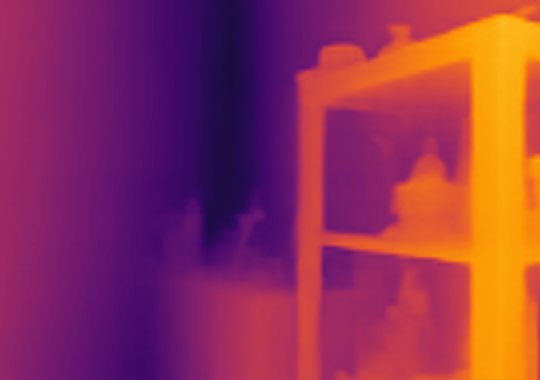}
\includegraphics[width=1.0\textwidth]{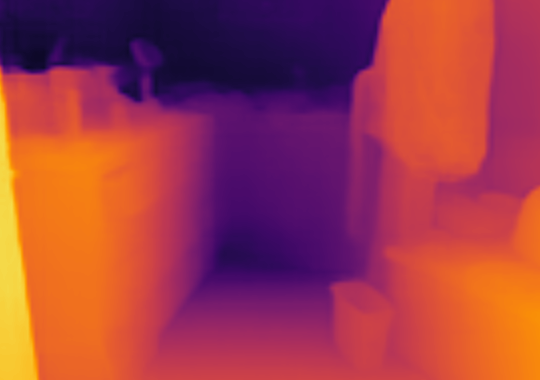}
\includegraphics[width=1.0\textwidth]{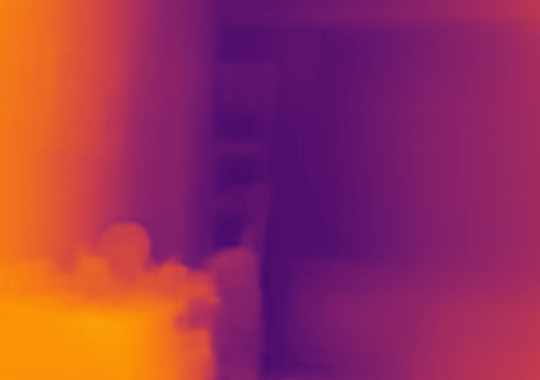}
\includegraphics[width=1.0\textwidth]{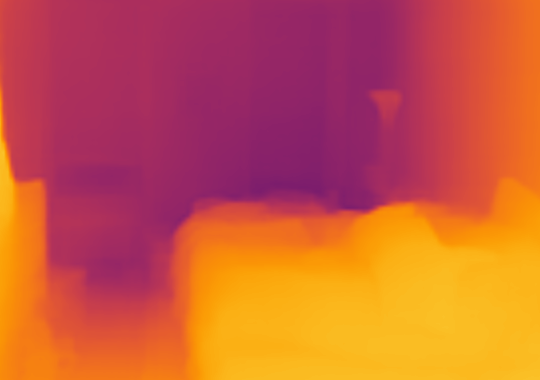}
\includegraphics[width=1.0\textwidth]{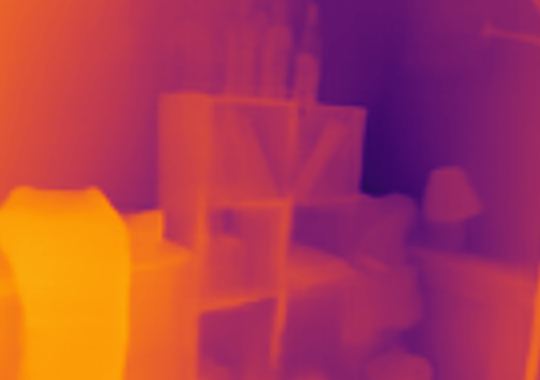}
\includegraphics[width=1.0\textwidth]{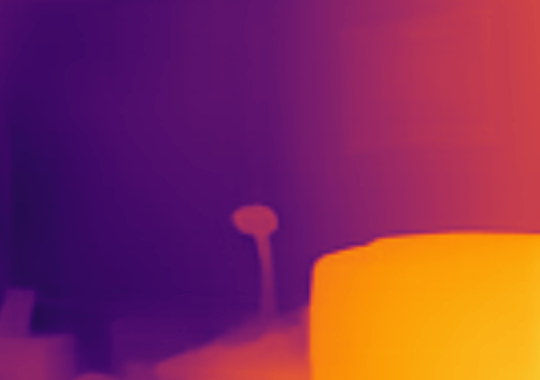}
\includegraphics[width=1.0\textwidth]{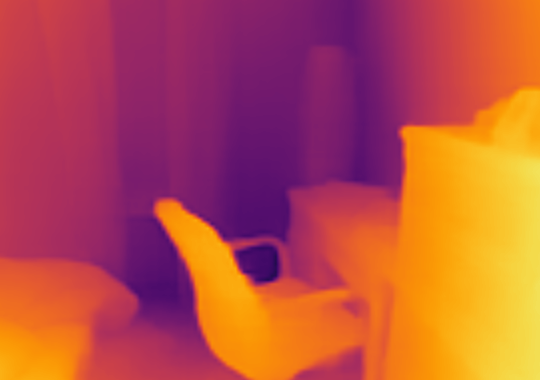}
\includegraphics[width=1.0\textwidth]{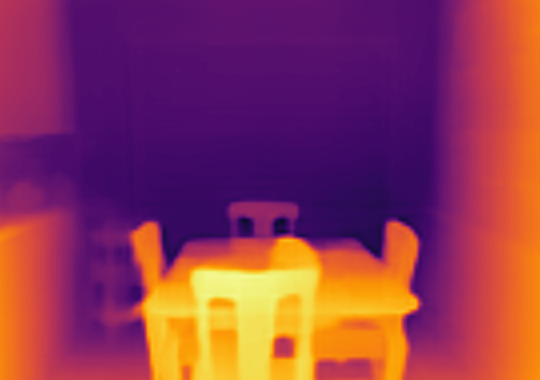}
\end{minipage}
\caption{NeWCRFs\cite{yuan2022new}}
\end{center}
\end{subfigure}
\begin{subfigure}[b]{0.18\textwidth}
\begin{center}
\begin{minipage}[b]{1.0\textwidth}
\includegraphics[width=1.0\textwidth]{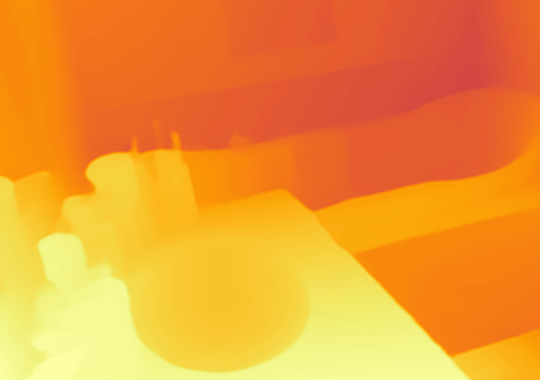}
\includegraphics[width=1.0\textwidth]{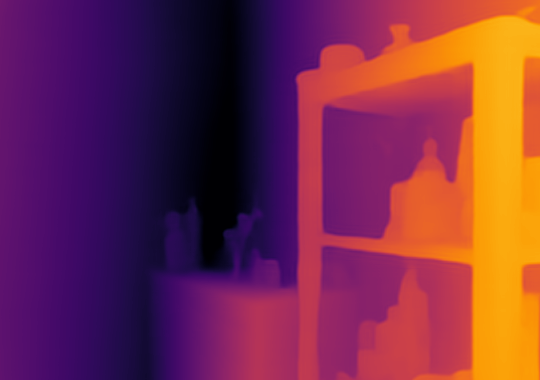}
\includegraphics[width=1.0\textwidth]{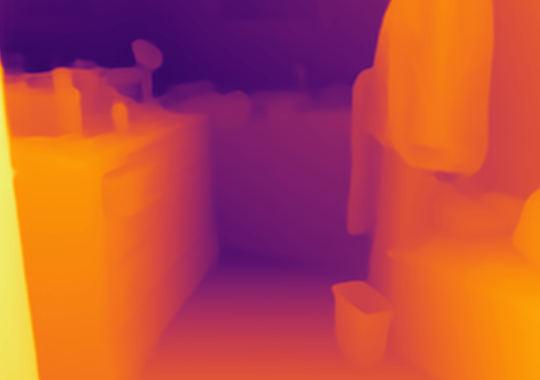}
\includegraphics[width=1.0\textwidth]{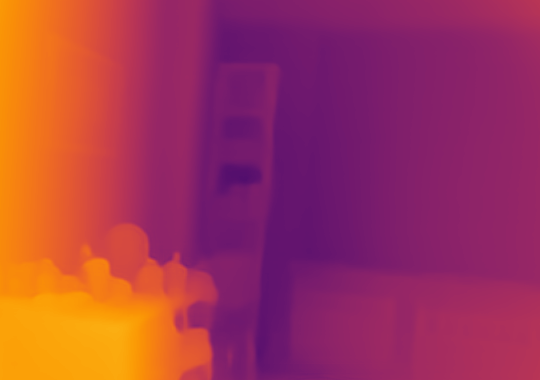}
\includegraphics[width=1.0\textwidth]{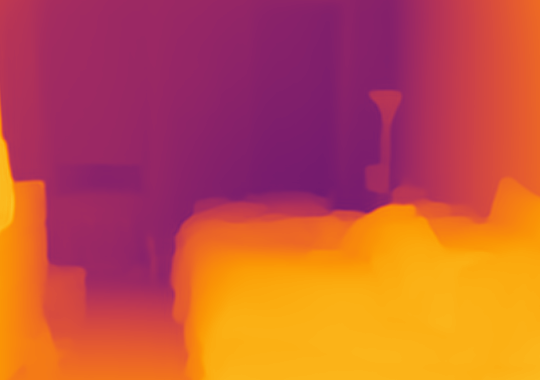}
\includegraphics[width=1.0\textwidth]{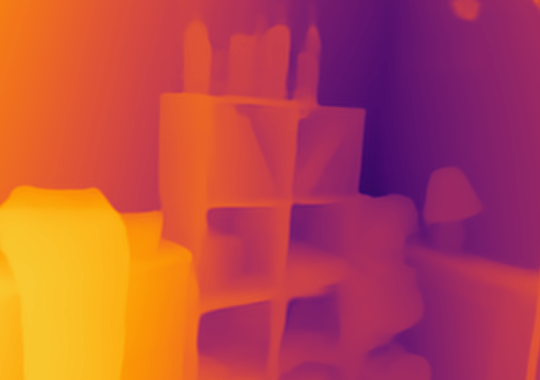}
\includegraphics[width=1.0\textwidth]{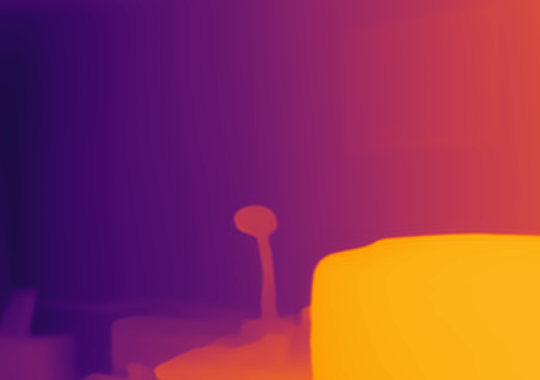}
\includegraphics[width=1.0\textwidth]{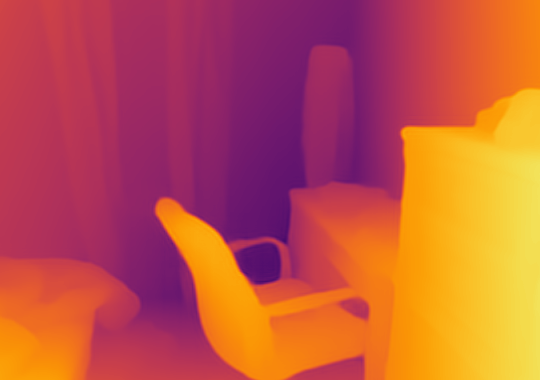}
\includegraphics[width=1.0\textwidth]{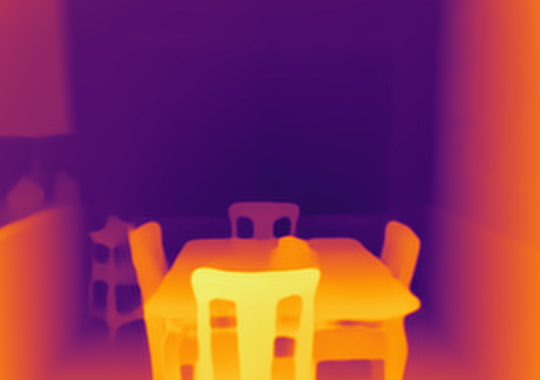}
\end{minipage}
\caption{\textbf{Ours}}
\end{center}
\end{subfigure}
\end{center}
\caption{\textbf{Qualitative Comparison on NYU Depth V2 test set} \cite{silberman2012indoor}. Our method recovers better depth even for complex scenes than the prior art such as (b) DPT \cite{ranftl2021vision}, (c) AdaBins \cite{bhat2021adabins}, (d) NeWCRFs \cite{yuan2022new}.}
\vspace{-4mm}
\label{fig:quality_nyu_supp}
\end{figure*}

\begin{figure*}[tb]
\centering
\begin{subfigure}[b]{0.3\textwidth}
\begin{minipage}[b]{1.0\textwidth}
\includegraphics[width=1\textwidth]{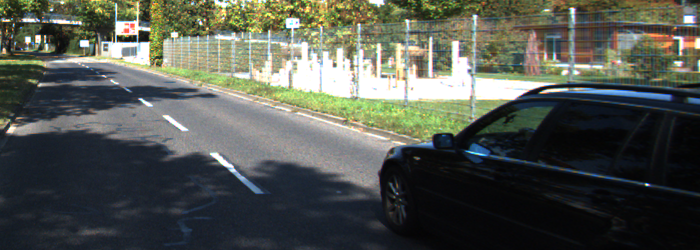}\\
\includegraphics[width=1\textwidth]{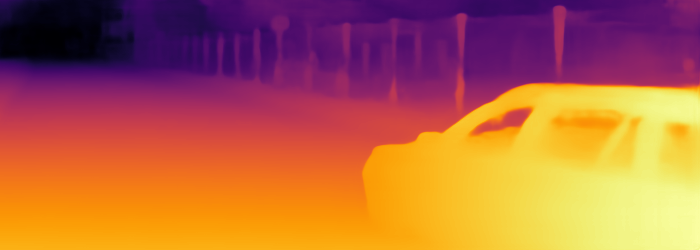}\\
\includegraphics[width=1\textwidth]{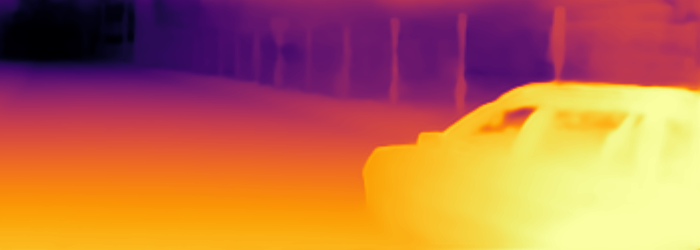}\\
\includegraphics[width=1\textwidth]{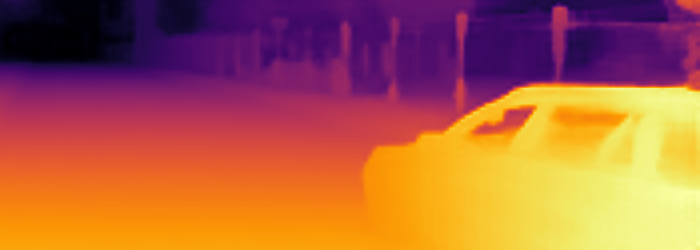}\\
\includegraphics[width=1\textwidth]{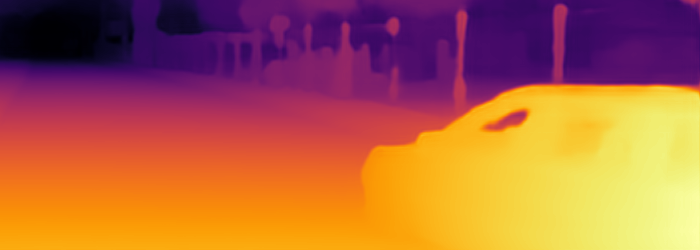}
\end{minipage}
\caption{Image 1}
\end{subfigure}\hspace{2pt}%
\begin{subfigure}[b]{0.3\textwidth}
\begin{minipage}[b]{1.0\textwidth}
\includegraphics[width=1\textwidth]{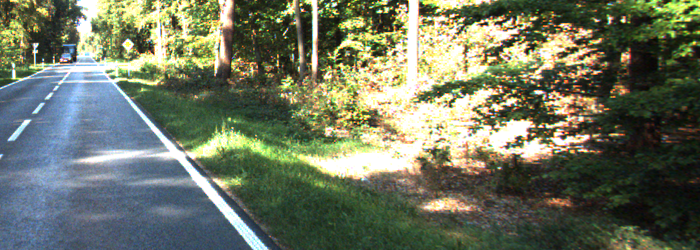}\\
\includegraphics[width=1\textwidth]{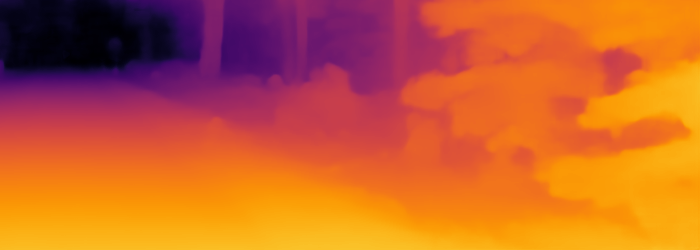}\\
\includegraphics[width=1\textwidth]{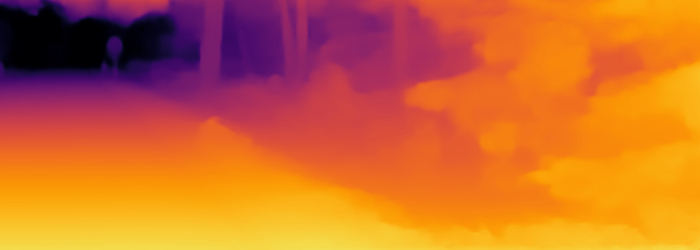}\\
\includegraphics[width=1\textwidth]{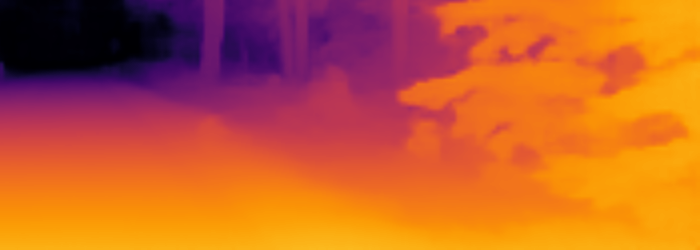}\\
\includegraphics[width=1\textwidth]{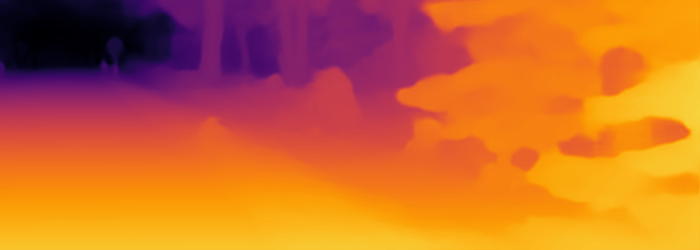}
\end{minipage}
\caption{Image 2}
\end{subfigure}\hspace{2pt}%
\begin{subfigure}[b]{0.3\textwidth}
\begin{minipage}[b]{1.0\textwidth}
\includegraphics[width=1\textwidth]{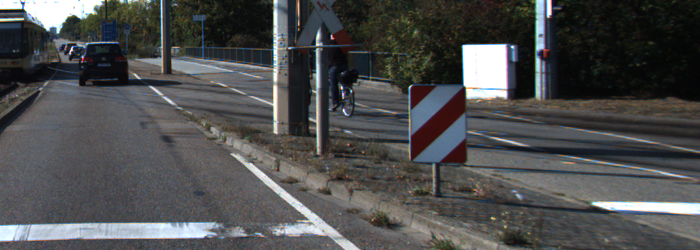}\\
\includegraphics[width=1\textwidth]{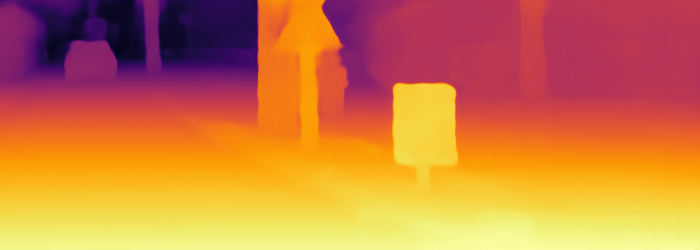}\\
\includegraphics[width=1\textwidth]{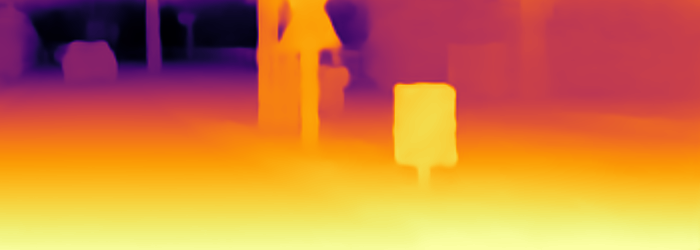}\\
\includegraphics[width=1\textwidth]{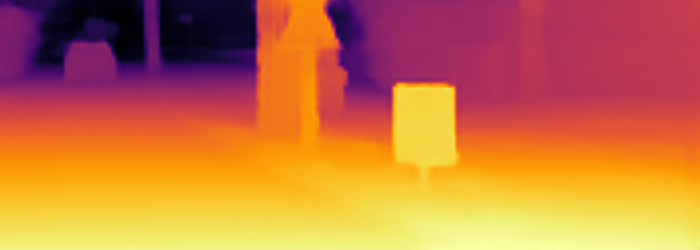}\\
\includegraphics[width=1\textwidth]{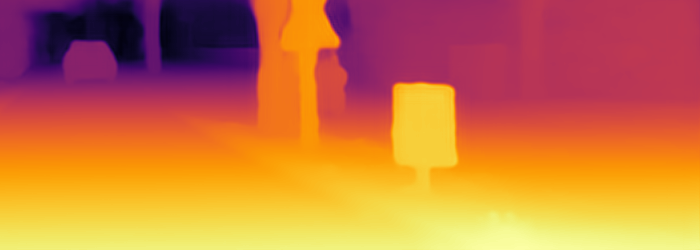}
\end{minipage}
\caption{Image 3}
\end{subfigure}

\begin{subfigure}[b]{0.3\textwidth}
\begin{minipage}[b]{1.0\textwidth}
\includegraphics[width=1\textwidth]{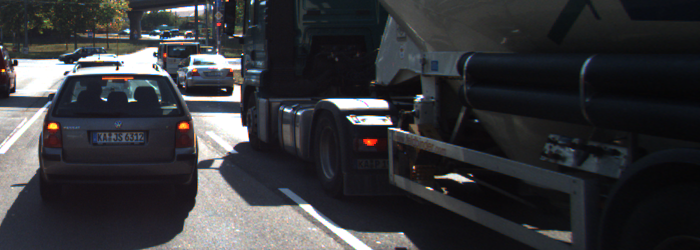}\\
\includegraphics[width=1\textwidth]{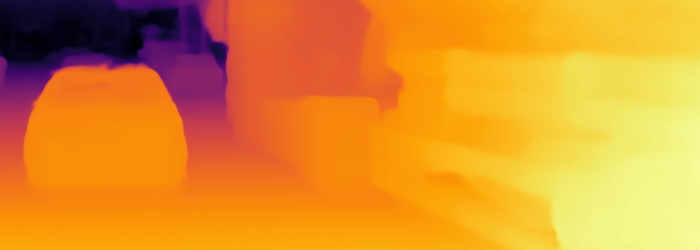}\\
\includegraphics[width=1\textwidth]{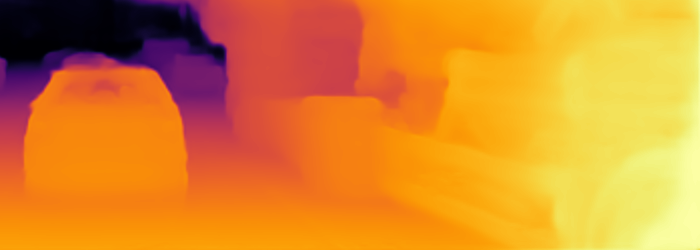}\\
\includegraphics[width=1\textwidth]{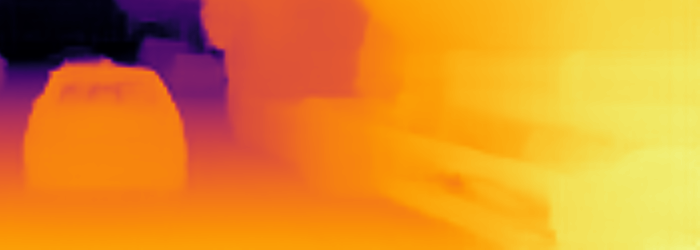}\\
\includegraphics[width=1\textwidth]{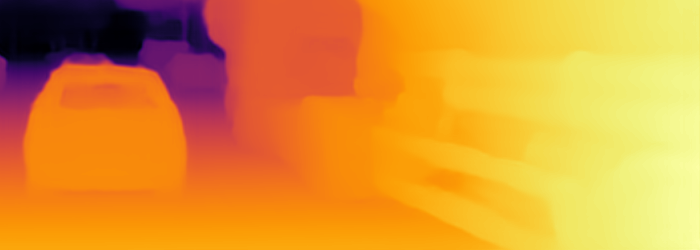}
\end{minipage}
\caption{Image 4}
\end{subfigure}\hspace{2pt}%
\begin{subfigure}[b]{0.3\textwidth}
\begin{minipage}[b]{1.0\textwidth}
\includegraphics[width=1\textwidth]{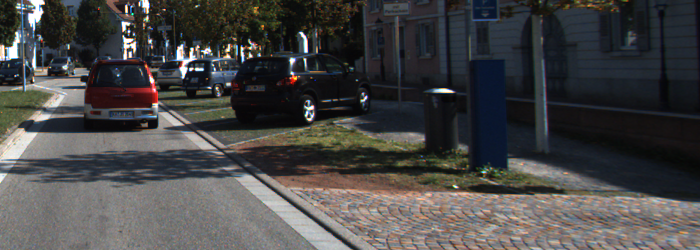}\\
\includegraphics[width=1\textwidth]{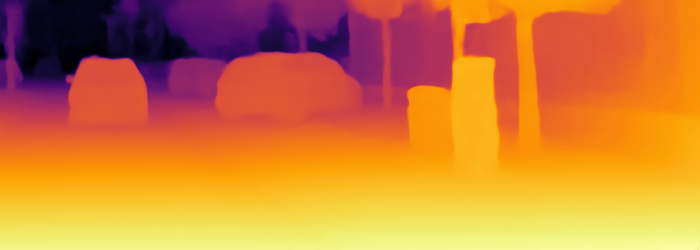}\\
\includegraphics[width=1\textwidth]{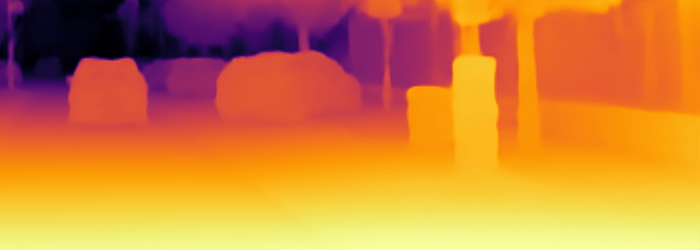}\\
\includegraphics[width=1\textwidth]{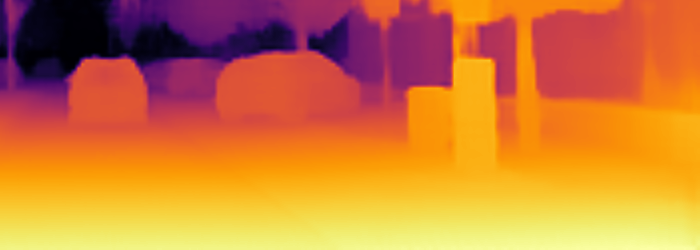}\\
\includegraphics[width=1\textwidth]{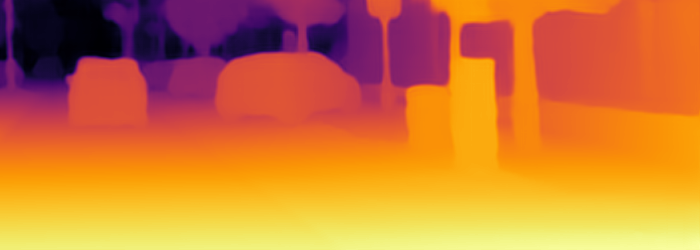}
\end{minipage}
\caption{Image 5}
\end{subfigure}\hspace{2pt}%
\begin{subfigure}[b]{0.3\textwidth}
\begin{minipage}[b]{1.0\textwidth}
\includegraphics[width=1\textwidth]{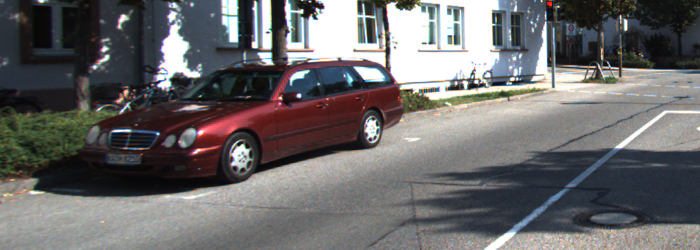}\\
\includegraphics[width=1\textwidth]{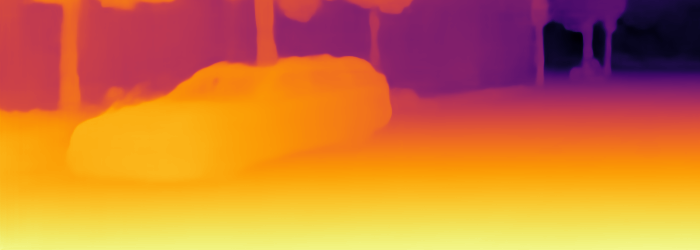}\\
\includegraphics[width=1\textwidth]{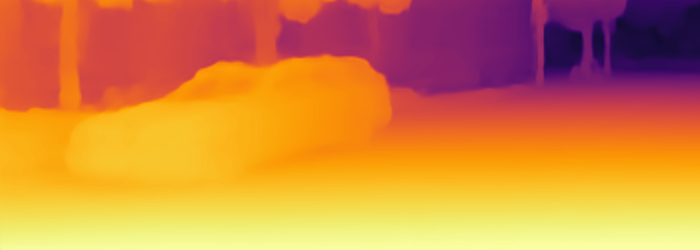}\\
\includegraphics[width=1\textwidth]{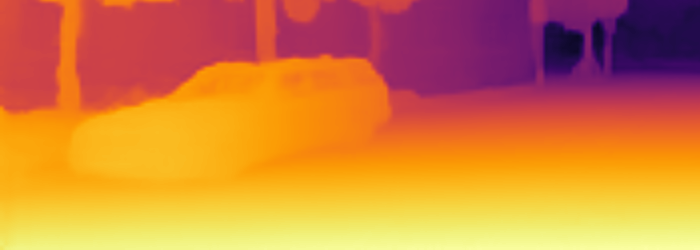}\\
\includegraphics[width=1\textwidth]{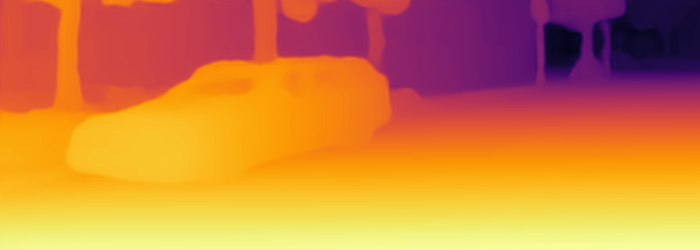}
\end{minipage}
\caption{Image 6}
\end{subfigure}
\caption{\textbf{Qualitative comparison on KITTI Eigen split}~\cite{eigen2014depth}. For each column, from top to bottom we present the input image, the prediction from DPT~\cite{ranftl2021vision}, AdaBins~\cite{bhat2021adabins}, NeWCRFs~\cite{yuan2022new}, and our framework respectively.}
\label{fig:quality_kitti_supp}
\end{figure*}

\begin{figure*}[t]
\begin{center}
\begin{subfigure}[b]{0.2\textwidth}
\begin{center}
\begin{minipage}[b]{1.0\textwidth}
\includegraphics[width=1.0\textwidth]{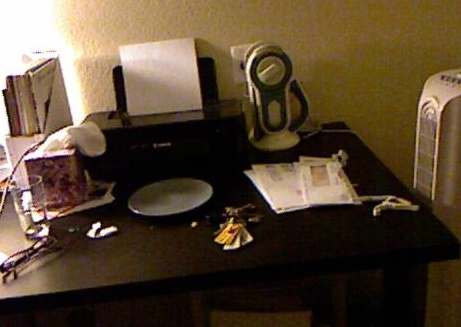}
\includegraphics[width=1.0\textwidth]{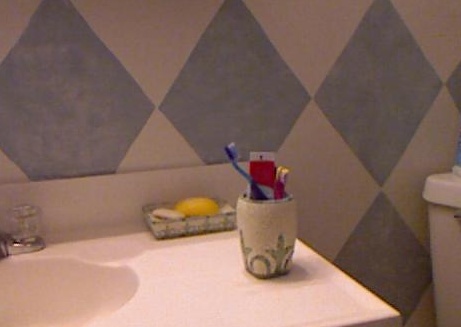}
\includegraphics[width=1.0\textwidth]{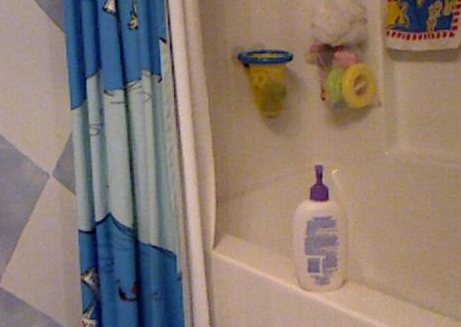}
\includegraphics[width=1.0\textwidth]{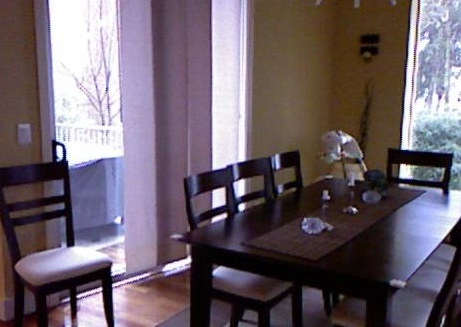}
\includegraphics[width=1.0\textwidth]{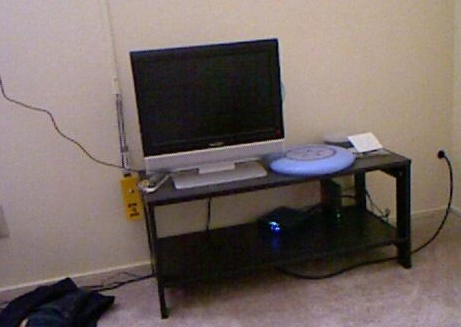}
\includegraphics[width=1.0\textwidth]{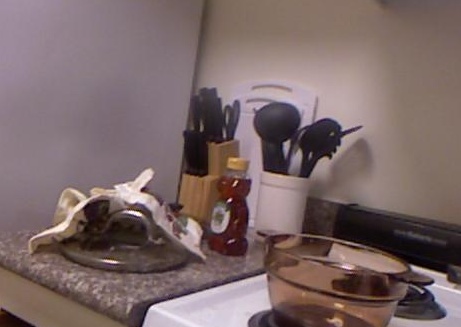}
\includegraphics[width=1.0\textwidth]{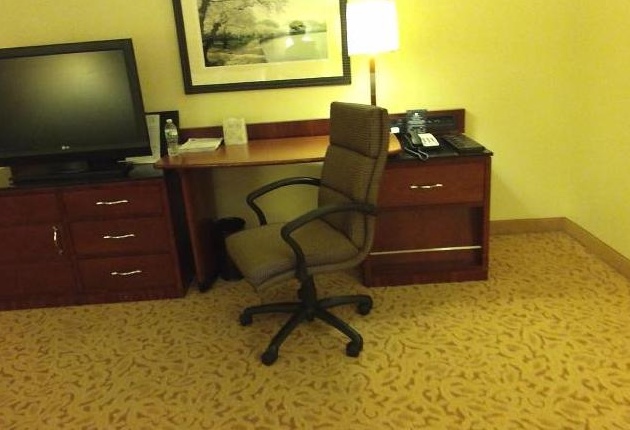}
\includegraphics[width=1.0\textwidth]{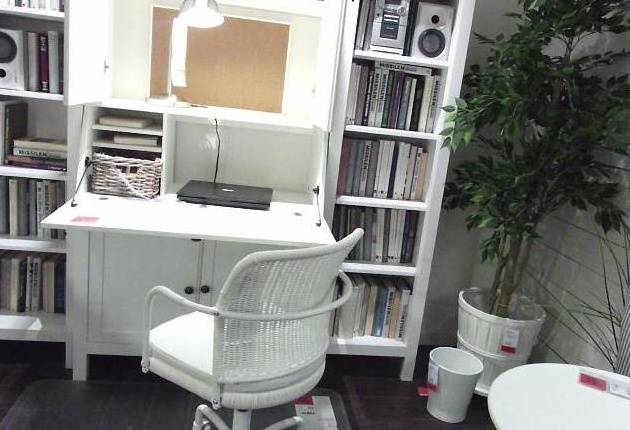}
\end{minipage}
\caption{Test}
\end{center}
\end{subfigure}
\begin{subfigure}[b]{0.2\textwidth}
\begin{center}
\begin{minipage}[b]{1.0\textwidth}
\includegraphics[width=1.0\textwidth]{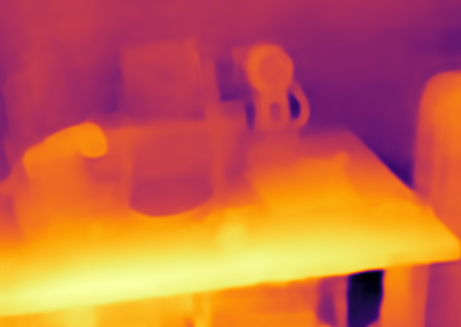}
\includegraphics[width=1.0\textwidth]{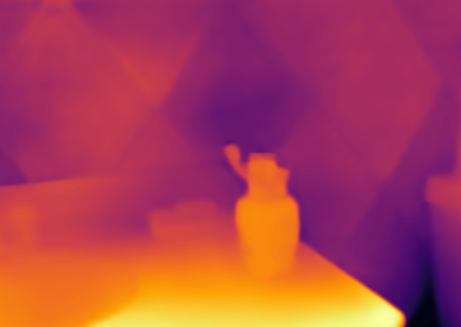}
\includegraphics[width=1.0\textwidth]{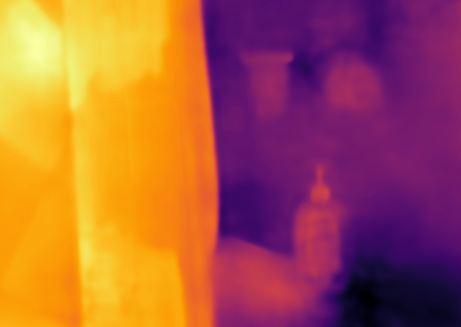}
\includegraphics[width=1.0\textwidth]{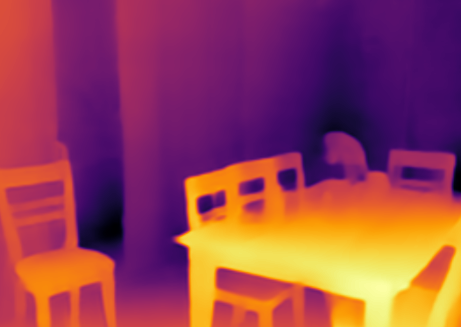}
\includegraphics[width=1.0\textwidth]{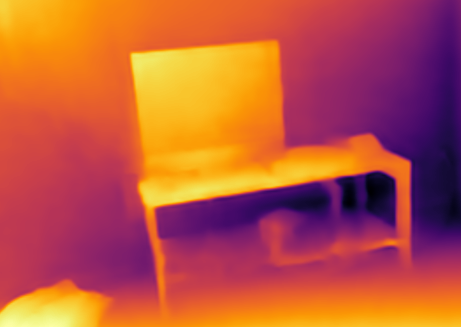}
\includegraphics[width=1.0\textwidth]{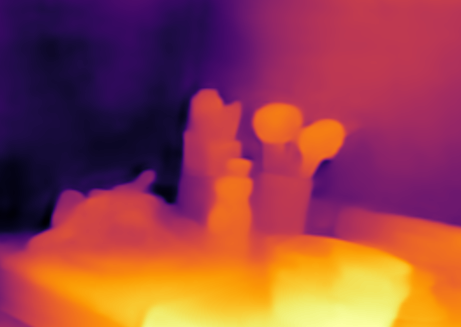}
\includegraphics[width=1.0\textwidth]{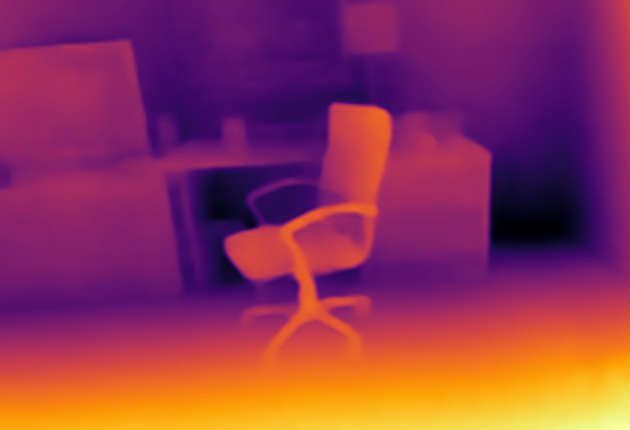}
\includegraphics[width=1.0\textwidth]{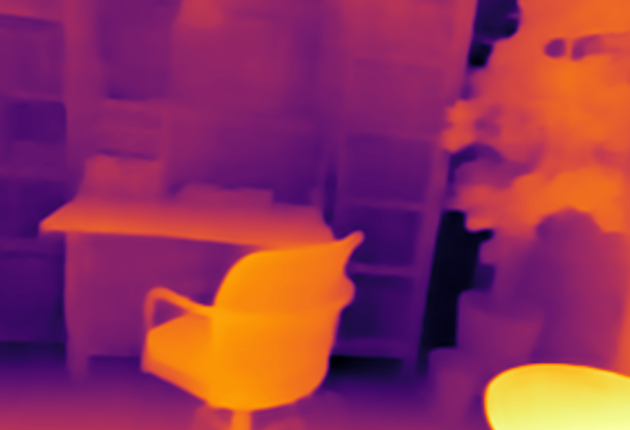}
\end{minipage}
\caption{AdaBins\cite{bhat2021adabins}}
\end{center}
\end{subfigure}  
\begin{subfigure}[b]{0.2\textwidth}
\begin{center}
\begin{minipage}[b]{1.0\textwidth}
\includegraphics[width=1.0\textwidth]{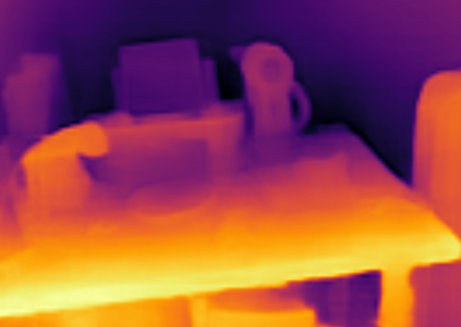}
\includegraphics[width=1.0\textwidth]{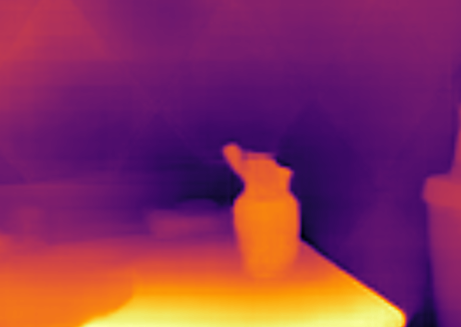}
\includegraphics[width=1.0\textwidth]{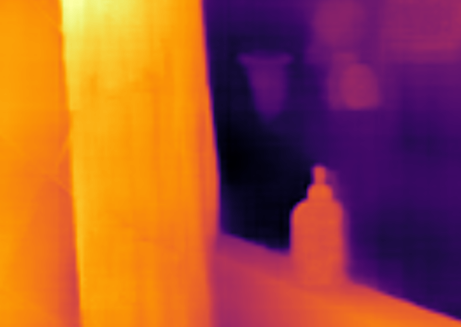}
\includegraphics[width=1.0\textwidth]{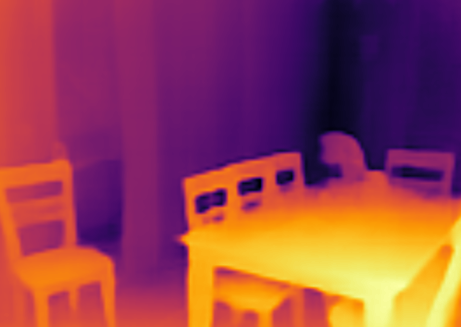}
\includegraphics[width=1.0\textwidth]{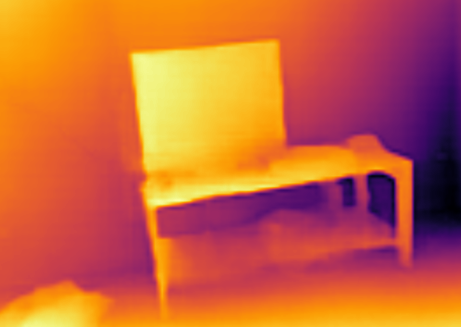}
\includegraphics[width=1.0\textwidth]{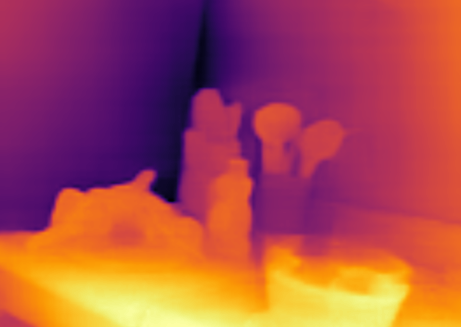}
\includegraphics[width=1.0\textwidth]{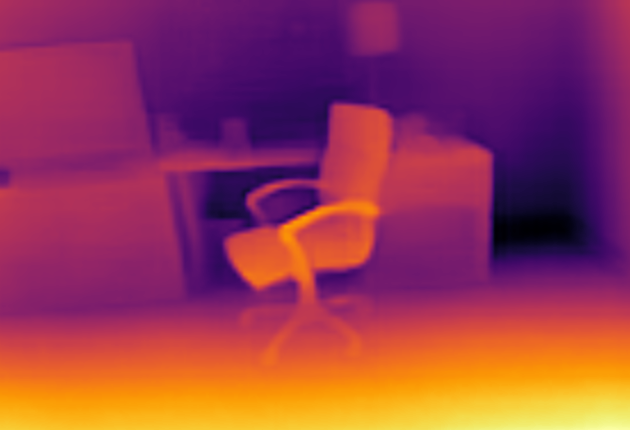}
\includegraphics[width=1.0\textwidth]{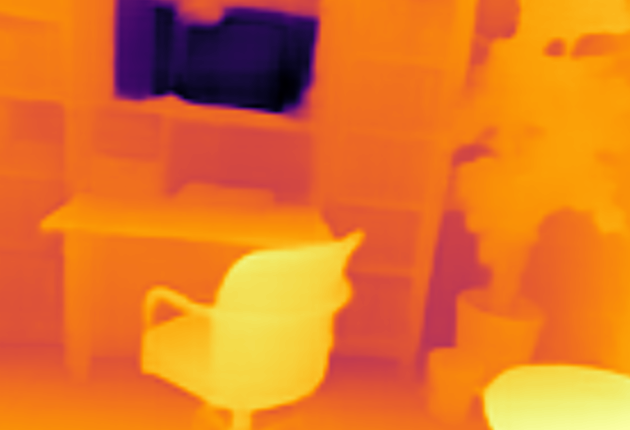}
\end{minipage}
\caption{NeWCRFs\cite{yuan2022new}}
\end{center}
\end{subfigure}
\begin{subfigure}[b]{0.2\textwidth}
\begin{center}
\begin{minipage}[b]{1.0\textwidth}
\includegraphics[width=1.0\textwidth]{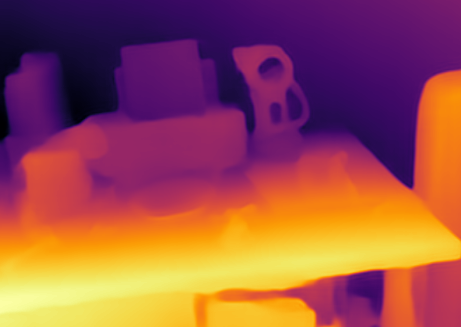}
\includegraphics[width=1.0\textwidth]{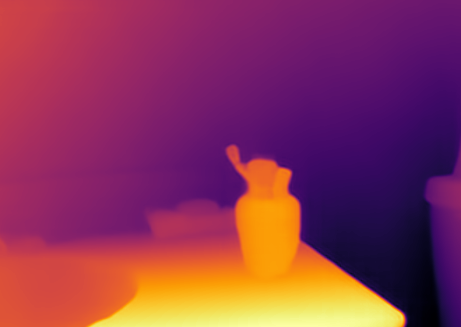}
\includegraphics[width=1.0\textwidth]{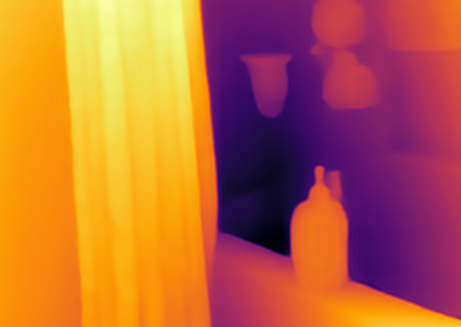}
\includegraphics[width=1.0\textwidth]{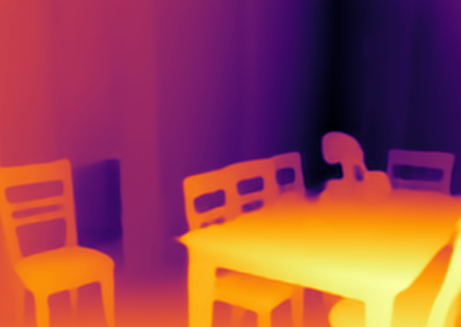}
\includegraphics[width=1.0\textwidth]{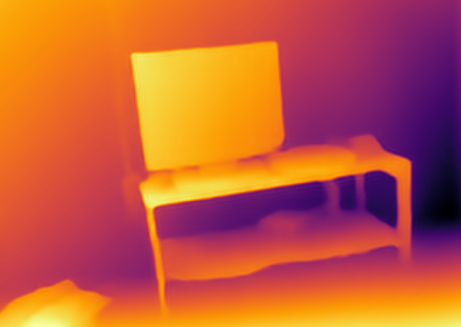}
\includegraphics[width=1.0\textwidth]{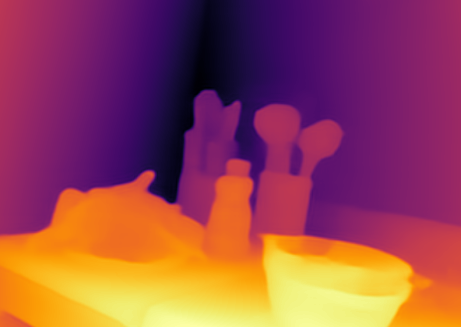}
\includegraphics[width=1.0\textwidth]{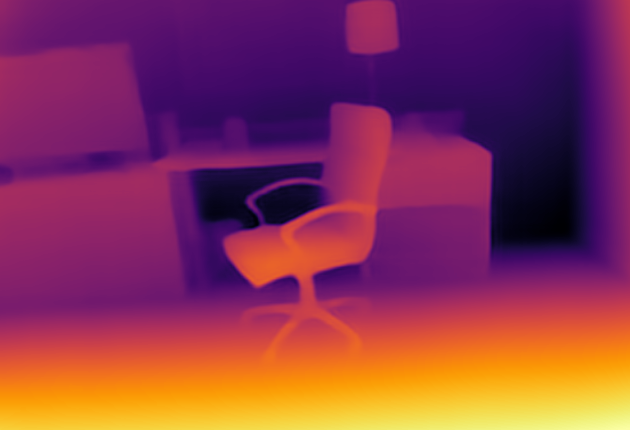}
\includegraphics[width=1.0\textwidth]{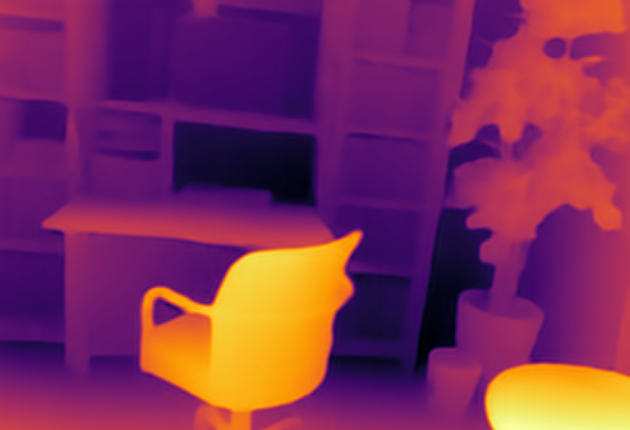}
\end{minipage}
\caption{\textbf{Ours}}
\end{center}
\end{subfigure}
\end{center}
\caption{\textbf{Qualitative Comparison on SUN RGB-D} \cite{song2015sun}. All the methods are trained on NYU Depth V2 \cite{silberman2012indoor} without fine-tuning on SUN RGB-D. Our method generalizes better on unseen scenes than (b) AdaBins \cite{bhat2021adabins} and (c) NeWCRFs \cite{yuan2022new}.}
\vspace{-4mm}
\label{fig:quality_sun_supp}
\end{figure*}

We provide more qualitative results on NYU Depth V2 \cite{silberman2012indoor}, KITTI Eigen split \cite{geiger2012we, eigen2014depth} and SUN RGB-D \cite{song2015sun} in Fig. \ref{fig:quality_nyu_supp}, Fig. \ref{fig:quality_kitti_supp}, Fig. \ref{fig:quality_sun_supp}, respectively. The depth prediction from our method contains more details about the scenes, especially in NYU Depth V2 and SUN RGB-D.

\end{document}